\documentclass[runningheads]{llncs}
\usepackage{graphicx}
\usepackage{graphicx}
\usepackage[sort,space,adjust,nocompress]{cite}

\usepackage[pagebackref=true,breaklinks=true,colorlinks,bookmarks=false]{hyperref}
\usepackage{amsmath}
\usepackage{amssymb}

\usepackage{multirow}
\usepackage{multicol}
\usepackage{diagbox}
\usepackage{comment}
\usepackage{url}
\usepackage{bbding}
\usepackage{xfrac}

\usepackage[labelformat=simple]{subcaption}

\usepackage{algorithmic}
\usepackage{algorithm}

\usepackage{caption}
\makeatletter
\usepackage{xspace}
\DeclareRobustCommand\onedot{\futurelet\@let@token\@onedot}
\def\@onedot{\ifx\@let@token.\else.\null\fi\xspace}

\makeatother

\begin{document}
\title{Shift and matching queries\\ for video semantic segmentation}

\titlerunning{Shift and matching queries for video semantic segmentation}

\author{
Tsubasa Mizuno\inst{1} \and
Toru Tamaki\inst{1}\orcidID{0000-0001-9712-7777}}
\authorrunning{T. Mizuno and T. Tamaki}
\institute{
Nagoya Institute of Technology, Nagoya, Japan\\
\email{t.mizuno.384@nitech.jp},
\email{tamaki.toru@nitech.ac.jp}
}


%
\maketitle              

\begin{abstract}
Video segmentation is a popular task, but applying image segmentation models frame-by-frame to videos does not preserve temporal consistency. In this paper, we propose a method to extend a query-based image segmentation model to video using feature shift and query matching. The method uses a query-based architecture, where decoded queries represent segmentation masks. These queries should be matched before performing the feature shift to ensure that the shifted queries represent the same mask across different frames. Experimental results on CityScapes-VPS and VSPW show significant improvements from the baselines, highlighting the method's effectiveness in enhancing segmentation quality while efficiently reusing pre-trained weights.

\keywords{
semantic segmentation
\and
vision transformer
\and
feature shift
\and
adaptation
}
\end{abstract}

\section{Introduction}

Segmentation \cite{Minaee_PAMI2022_segmentation_survey} involves predicting the labels for objects or scene elements for each pixel in an image. This is useful in many applications, such as understanding urban landscapes in automated driving technology, and research has attracted much attention.
Hence, various \emph{image} segmentation models have been proposed \cite{Minaee_PAMI2022_segmentation_survey,Kirillov_CVPR2019_panoptic_segmentation,He_ICCV2017_Mask_R-CNN,Cheng_NEURIPS2021_MaskFormer,Cheng_2022CVPR_Mask2Former,Jain_2023CVPR_OneFormer}.
However, in the case of \emph{video} segmentation, applying image segmentation models frame-by-frame does not preserve temporal consistency. Therefore, capturing the temporal relationship between frames is necessary, and many segmentation methods have been developed specifically for videos.

There are two major problems in the development of video segmentation models.
The first is the increased computational complexity and time required to process videos. Unlike image segmentation, which deals with a single image, video segmentation requires handling many frames efficiently. Therefore, methods suitable for real-time processing \cite{Park_2022_CVPR} and online processing \cite{Ying_2023_ICCV, Wu_2023_ICCV} have been proposed.
The second is the effective use of image segmentation models. Currently, many pre-trained image models are available through platforms such as Hugging Face \cite{Wolf_EMNLP2020_Transformers_Huggingface}\footnote{\url{https://huggingface.co}}. However, designing new models for large video datasets and training them is an expensive process. Therefore, many tasks have been studied by reusing weights from image models and applying them to video models \cite{Zhang_ACMMM2021_TokenShift,Rasheed_ICCV2023,Lin_ECCV2022}. Although methods for video segmentation have been proposed \cite{Ying_2023_ICCV,Li_2023_ICCV}, they do not efficiently reuse pre-trained weights.

In this study, we propose a video segmentation method that effectively uses an image segmentation model pre-trained on image datasets. The proposed model not only applies the image segmentation model to each frame, but also models temporal information using feature shift. This feature shift is widely used in action recognition \cite{Lin_2019ICCV_TSM,Zhang_ACMMM2021_TokenShift,Hashiguchi_2022_ACCVW_MSCA}, but it is the first time it has been introduced in a segmentation model. The segmentation model that we based on is a query-based method \cite{Cheng_NEURIPS2021_MaskFormer}, which represents a segmentation mask by a single trainable \emph{query}.
However, simple feature shift and query-based methods are incompatible. This is because a feature shift exchanges features of the previous and next frames, but if applied to queries, it might shift queries representing different segmentation masks. In this study, we propose a query matching before performing the feature shift and demonstrate its effectiveness.

\section{Related Work}

\subsection{Image Segmentation}

Image segmentation models using CNNs, like as U-Net \cite{Ronneberger_MICCAI2015_U-Net}, typically include an encoder that takes an input image and reduces its spatial resolution, and a decoder that increases the spatial resolution and outputs a label image of the same size as the input image.
However, after the success of DETR \cite{Carion_ECCV2020_DETR}, a query-based object detection method using Transformer \cite{Vaswani_NeurIPS2017_transformer}, methods that incorporate the idea of object query in DETR have been proposed for tasks beyond object detection, leading to many new query-based segmentation methods \cite{Cheng_NEURIPS2021_MaskFormer,Cheng_2022CVPR_Mask2Former,Jain_2023CVPR_OneFormer,Li_CVPR2023_MaskDINO}.
In this study, we adopt MaskFormer \cite{Cheng_NEURIPS2021_MaskFormer} as the backbone and extend it to video using feature shift.

\subsection{Video Segmentation}

Various methods have been proposed for modeling temporal information in video segmentation. These include aggregating information from previous frames \cite{Jain_2019_CVPR}, using multiresolution information by changing the frame sampling stride \cite{Lao_2023_CVPR}, using pseudolabels for frames, and employing attention to past frames \cite{Hu_2020_CVPR,10.1145/3474085.3475409}.

However, all of these models are specific to video segmentation and no study has efficiently reused image segmentation models as in this study. Previous studies that extended query-based methods to video include Tube-Link \cite{Li_2023_ICCV} and TarViS \cite{Athar_2023_CVPR} for universal video segmentation. These require matching between multiple queries across frames, making the mechanism very complex. In contrast, this study aims for an extension method that is easy to implement and analyze, by matching queries and applying temporal feature shift.

\begin{figure}[t]
    \centering

    \includegraphics[width=\linewidth]{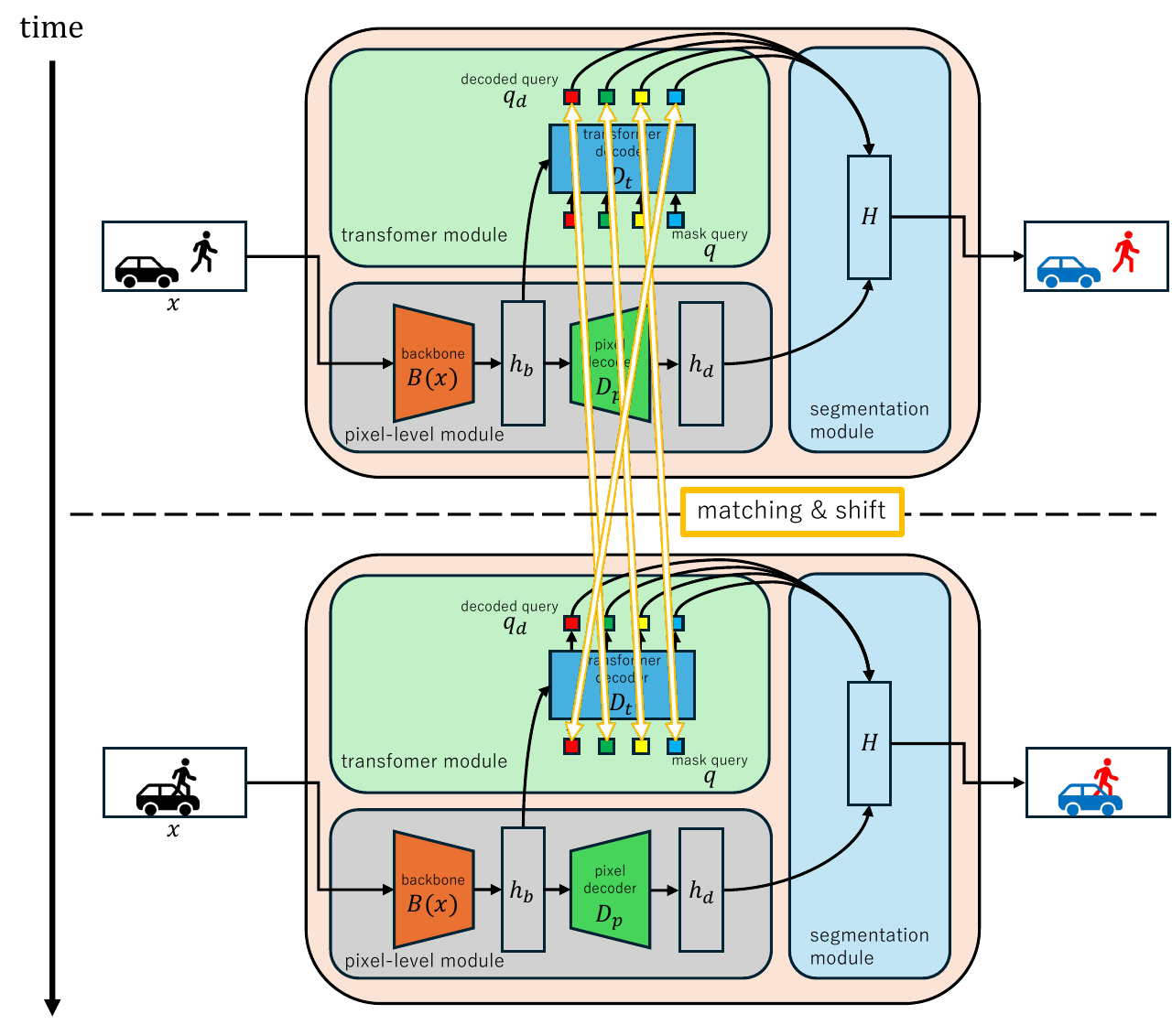}

    \caption{
        The architecture of our proposed method includes feature shift and query matching. It utilize an image segmentation model (the orange plate) frame-by-frame, which has a backbone for feature extraction, a pixel decoder, a transformer decoder for processing mask queries, and a segmentation module for prediction. The proposed query matching ensures temporal consistency even when feature shift is applied to decoded queries of different frames.
    }
    \label{fig:model overview}

\end{figure}

\section{Method}

In this section, we first describe the architecture common to query-based segmentation models and then explain how we propose to extend it to videos by using feature shift and query matching.

\subsection{Segmentation architecture}
\label{sec:segmentation architecture}

Semantic segmentation models for images often consist of a backbone $B(x)$ that receives an input image $x$, a pixel decoder $D_p(h_b)$ that transforms the backbone features $h_b = B(x)$ back to the input image size, a transformer decoder $D_t(h_b, q)$ that transforms \emph{mask queries} $q$ based on $h_b$, and a header $H(h_d, q_d)$ that predicts segmentation masks and classes using the \emph{decoded queries} $q_d = D_t(h_d, q)$ and the pixel decoder output $h_d$.

When applied to video, each frame $v(t), t=1,\ldots,T$ of an input video $v$ is fed $B$ (Fig.\ref{fig:model overview}).
However, this naive frame-by-frame approach does not preserve temporal coherency.

\subsection{Feature shift}

In this study, the feature shift is used to exchange temporal information in a video, by shifting the features of specific channels of a model in the temporal direction \cite{Lin_2019ICCV_TSM}.

Suppose $z_{in} \in \mathbb{R}^{T\times D}$ is an input feature of dimension $D$, which is a stack of frame features of a video clip of $T$ frames.
In the following, $z_{in, t, d}$ denotes the element at $(t, d)$ in $z_{in}$.
It is fed to shifting modules \cite{Hashiguchi_2022_ACCVW_MSCA} to produces output $z_{out}\in \mathbb{R}^{T\times D}$, a temporally shifted version of the input as follows;
\begin{align}
    z_{\mathrm{out}, t, d}
        &= 
    \begin{cases}
        z_{\mathrm{in}, t-1, d}, & \text{if } 1 < t \le T, 1 \le d \le D_f \\
        z_{\mathrm{in}, t+1, d}, & \text{if } 1 \le t < T, D - D_b \le d \le D \\
        z_{\mathrm{in}, t, d},   & \text{if } \forall t, D_f+1 \le d \le D - D_b-1. \\
    \end{cases}
    \label{eq:feature shift}
\end{align}
This means that first $D_f$ channels are shifted forward to the next frame $t + 1$,
the last $D_b$ channels are shifted backward to the previous frame $t - 1$, and
the remaining channels are left untouched.
Apparently, this feature shift only works when the channel indexes have the same meaning. However, queries in query-based methods may refer to different things or stuff (i.e., segmentation masks) for each frame. Therefore, a naive shift does not maintain feature compatibility between frames.

In this experiment, we have choices for the features that are shifted in the backbone $B$, pixel decoder $D_p$, transformer decoder $D_t$, or queries $q$ and $q_d$. Among these, shifting the decoded mask queries $q_d$ is reasonable because it directly affects the quality of the output segmentation mask,
and our preliminary experiments have shown a small impact on performance when we apply the feature shift to other modules.

\subsection{Query matching}

Since each decoded mask query corresponds to a segmentation mask, simply shifting the feature of the queries with the same index across frames might mix up features of queries that actually represent different segmentation masks in different frames. To address this, we propose a method to match queries across frames that are likely to correspond to the same segmentation mask (see Fig.\ref{fig:model overview}).

First, we compute cosine similarities for each pair of $N$ queries in adjacent frames. Next, we match the queries by solving a bipartite matching problem using the Hungarian algorithm to find the optimal permutation $\hat{\sigma}$ as follows:
\begin{equation}
    \hat{\sigma} = \underset{\sigma \in \mathrm{\Omega}} {\operatorname{argmin}}
    \sum_{i} \mathrm{sim} (q_{i}^{t}, q_{\hat{\sigma}(i)}^{t+1}),
\end{equation}
where $\mathrm{sim}()$ is cosine similarity
between the $i$-th query $q_{i}^{t}$ at time $t$ and the query $q_{\hat{\sigma}(i)}^{t+1}$ of index $\hat{\sigma}(i)$ at time $t + 1$,
and $\mathrm{\Omega}$ is a set of permutations.
This query matching is performed on all adjacent frames, and then feature shift is applied to the matched queries.

\section{Experiments}

\subsection{Datasets}

\noindent\textbf{Cityscapes-VPS}
\cite{Kim_2020_CVPR_VideoPanopticSegmentation} is a dataset for Video Panoptic Segmentation (VPS) built upon Cityscapes \cite{Cordts_2016_CVPR_Cityscapes}, which is a well-known dataset for image panoptic segmentation
but only has 5,000 finely annotated frames from 5,000 videos. In contrast, Cityscapes-VPS sampled every five frames from 500 Cityscapes videos, making it suitable for our purpose because the temporal feature shift is suitable for the denser frame sampling.
The resulting dataset contains 2,400 finely annotated frames for training and 500 for validation,
with 19 categories at a resolution of 1,024$\times$2,048.
There are six annotated frames in each video, with approximately 0.29 seconds between each frame,
since the frame rate of the videos is 17 fps.
Although instance IDs are also associated, we do not use them because this paper focuses on video semantic segmentation.

\noindent\textbf{VSPW}
(Video Scene Parsing in the Wild) \cite{Miao_2021_CVPR} is a dataset that
includes full annotations for 3,337 videos, each approximately 3 to 10 seconds (5 seconds on average), recorded at 15 fps with a frame interval of approximately 0.067 seconds. Pixel-level annotations cover 124 categories for 239,934 frames, with resolutions ranging from 720P to 4K.
Because the length varies by video, in this experiment, we used only videos with a minimum length of 32 frames and excluded shorter videos.
Consequently, 161,984 frames were used for training and 18,944 frames for validation.

\subsection{Model}

In this experiment, we use MaskFormer \cite{Cheng_NEURIPS2021_MaskFormer}, a simple query-based architecture for images. It consists of a backbone that encodes the input image, a pixel decoder, a transformer decoder that transforms mask queries, and a segmentation head, as mentioned in Section \ref{sec:segmentation architecture}.
The pixel-level module, which includes the backbone and the pixel decoder, extracts visual embeddings used to generate binary mask predictions. The transformer decoder computes segment-level embeddings from mask queries and generates decoded queries. The segmentation head generates predictions from the visual embeddings and the decoded queries.
We applied the feature shift with or without query matching for the decoded queries. The dimension of a query is $D=256$, and we specify the amount of feature shift as a fraction \cite{Lin_2019ICCV_TSM} ranging from $\sfrac{1}{128}$ (two channels are shifted; one channel forward and one channel backward) to $\sfrac{1}{4}$ (64 channels are shifted) of the dimension $D$.

\subsubsection{Training setup.}
In this experiment, we used MaskFormer \cite{Cheng_NEURIPS2021_MaskFormer} pre-trained on ADE20K \cite{Zhou_Springer2019_ADE20K} as the image model and applied the proposed method to the video datasets.
The experimental settings for Cityscapes-VPS \cite{Kim_2020_CVPR_VideoPanopticSegmentation} are a crop size of $256 \times 512$, 50 epochs, a learning rate of 1e-3, and a batch size of 6 frames (i.e., one video).
For VSPW \cite{Miao_2021_CVPR}, the crop size is $288 \times 512$, 10 epochs, a learning rate of 5e-4, and a batch size of 16 frames.
The evaluation metrics used are 
mean Intersection over Union (mIoU) \cite{Everingham_2015_IJCV_mIoU} and 
pixel accuracy \cite{Long_2015_CVPR}.

\subsection{Quantitative results}

\begin{table}[t] 
    \centering 

    \caption{
        Performance comparison on CityScapes-VPS for different shift channels and query-matching configurations.
        Performance differences from the baseline (no shifts) is shown in parentheses.
    }
    \label{tab:performance comparison cityscapes-vps}

    \begin{tabular}{cc|c|c|c}
        \multicolumn{2}{c}{shift} \\
        fraction & channels & matching & mIoU & pix. acc. \\ \hline
        0                                 &0&           & 55.66           & 92.91           \\ \hline
        \multirow{2}{*}{$\sfrac{1}{128}$} &\multirow{2}{*}{2}&           & 53.40 $(-2.26)$ & 92.28 $(-0.63)$ \\ 
                                          & & \checkmark& 53.26 $(-2.40)$ & 92.71 $(-0.20)$ \\ \hline
        \multirow{2}{*}{$\sfrac{1}{ 64}$} &\multirow{2}{*}{4}&           & 50.69 $(-4.97)$ & 91.35 $(-1.56)$ \\       
                                          & & \checkmark& 56.61 $(+0.95)$ & 92.68 $(-0.23)$ \\ \hline
        \multirow{2}{*}{$\sfrac{1}{ 32}$} &\multirow{2}{*}{8}&           & 51.49 $(-4.17)$ & 91.86 $(-1.05)$ \\       
                                          & & \checkmark& \textbf{57.68} $(+2.02)$ & \textbf{93.23} $(+0.32)$ \\ \hline
        \multirow{2}{*}{$\sfrac{1}{ 16}$} &\multirow{2}{*}{16}&           & 54.22 $(-1.44)$ & 92.03 $(-0.88)$ \\       
                                          & & \checkmark& 56.23 $(+0.63)$ & 92.57 $(-0.34)$ \\ \hline
        \multirow{2}{*}{$\sfrac{1}{  8}$} &\multirow{2}{*}{32}&           & 54.10 $(-1.56)$ & 92.21 $(-0.70)$ \\       
                                          & & \checkmark& 56.88 $(+1.22)$ & 92.71 $(-0.20)$ \\ \hline
        \multirow{2}{*}{$\sfrac{1}{  4}$} &\multirow{2}{*}{64}&           & 55.92 $(+0.26)$ & 92.42 $(-0.49)$ \\       
                                          & & \checkmark& 53.73 $(-1.93)$ & 92.56 $(-0.35)$ \\ \hline
    \end{tabular}

\end{table}

\begin{table*}[t]
    \centering

    \caption{Performance comparison on VSPW for different shift channels and query-matching configurations.} 
    \label{tab:performance comparison vpsw}

    \begin{tabular}{cc|c|c|c}
        \multicolumn{2}{c}{shift}\\
        fraction & channels & matching & mIoU & pix. acc. \\ \hline
        0                                &0&          &56.85          &87.66         \\ \hline
        \multirow{2}{*}{$\sfrac{1}{128}$}&\multirow{2}{*}{2}&          &54.27 $(-2.57)$&87.66 $(\pm0.00)$\\
                                         & &\checkmark&58.22 $(+2.72)$&88.35 $(+0.98)$\\ \hline
        \multirow{2}{*}{$\sfrac{1}{ 64}$}&\multirow{2}{*}{4}&          &57.26 $(+1.21)$&87.99 $(+0.62)$\\
                                         & &\checkmark&57.20 $(+1.70)$&88.97 $(+1.60)$\\ \hline
        \multirow{2}{*}{$\sfrac{1}{ 32}$}&\multirow{2}{*}{8}&          &57.66 $(+0.81)$&88.79 $(+1.13)$\\
                                         & &\checkmark&57.36 $(+1.86)$&87.77 $(+0.40)$\\ \hline
        \multirow{2}{*}{$\sfrac{1}{ 16}$}&\multirow{2}{*}{16}&          & 58.44 $(+1.59)$&87.96 $(+0.30)$\\
                                         & &\checkmark& \textbf{58.68} $(+3.18)$&88.37 $(+1.00)$\\ \hline
        \multirow{2}{*}{$\sfrac{1}{  8}$}&\multirow{2}{*}{32}&          &57.59 $(+0.74)$&88.17 $(+0.80)$\\        
                                         & &\checkmark&58.29 $(+2.79)$& \textbf{89.08} $(+1.71)$\\ \hline 
        \multirow{2}{*}{$\sfrac{1}{  4}$}&\multirow{2}{*}{64}&          &57.82 $(+2.32)$&88.18 $(+0.81)$\\        
                                         & &\checkmark&55.52 $(-1.33)$&87.99 $(+0.33)$\\ \hline
    \end{tabular}

\end{table*}

Table \ref{tab:performance comparison cityscapes-vps} shows the performance on CityScapes-VPS for different configurations. The top row shows the baseline configuration (no shifts, no query matching) as a reference, with differences from the baseline shown in parentheses in the following rows.
Without query matching, the mIoU generally decreases as the amount of shift increases from 1/128 to 1/32. Pixel accuracy also shows a slight decline. However, for fractions lower than 1/32 (or more channels are shifted), both indicators rise to the same level as the baseline.
Introducing the query matching with a shift between 1/64 to 1/8 consistently outperforms the baseline and their non-matching counterparts in both mIoU and pixel accuracy. In particular, the 1/32 shift with query matching outperforms the baseline by about 2\%, compared to a 6\% decrease without query matching. 
This indicates that the introduction of feature shift and query matching is effective and can significantly improve performance when the amount of feature shift is appropriately selected for the dataset, since making the shift fraction very small or very large decreases performance in both cases.

The performance on VSPW is shown in Table~\ref{tab:performance comparison vpsw}, where the proposed method demonstrated improvements compared to Table~\ref{tab:performance comparison cityscapes-vps}.
One notable observation is that the performance on VSPW improves even without query matching. However, using query matching further enhances the performance, with mIoU consistently better for a shift between 1/128 and 1/8.
The significant improvement on VSPW may be attributed to its nature as a denser video dataset. While CityScapes-VPS videos have a frame interval of 0.29 seconds, VSPW videos have a much shorter interval of 0.067 seconds. This results in very little difference in appearance between frames, making the feature shift highly effective.

\subsection{Qualitative results}

\begin{figure*}[t]
    \centering
    \def\resultimagewidth{0.15}
    
    \subcaptionbox{
        original frames
    }{
        \includegraphics[width=\resultimagewidth\linewidth]{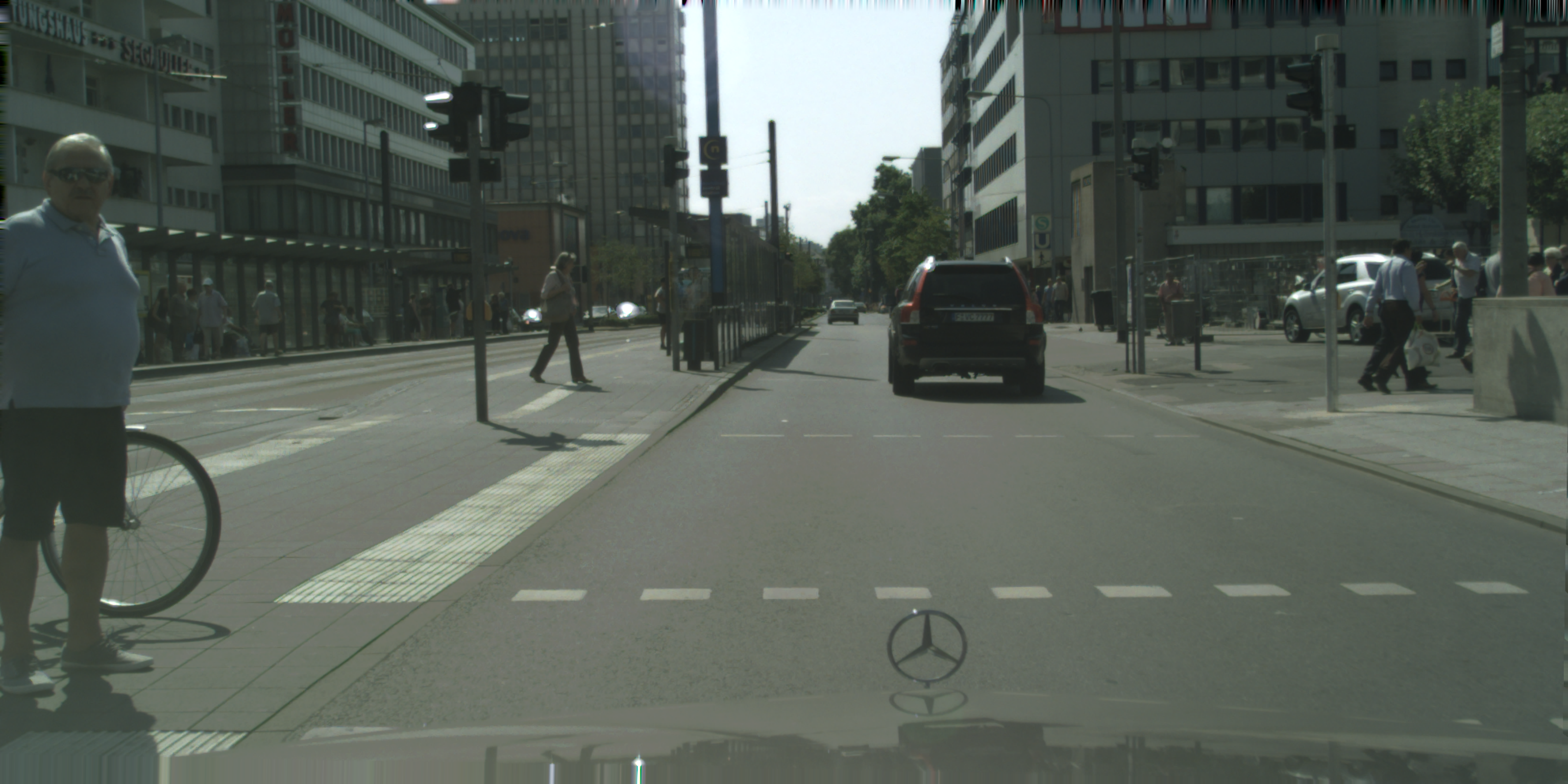}
        \includegraphics[width=\resultimagewidth\linewidth]{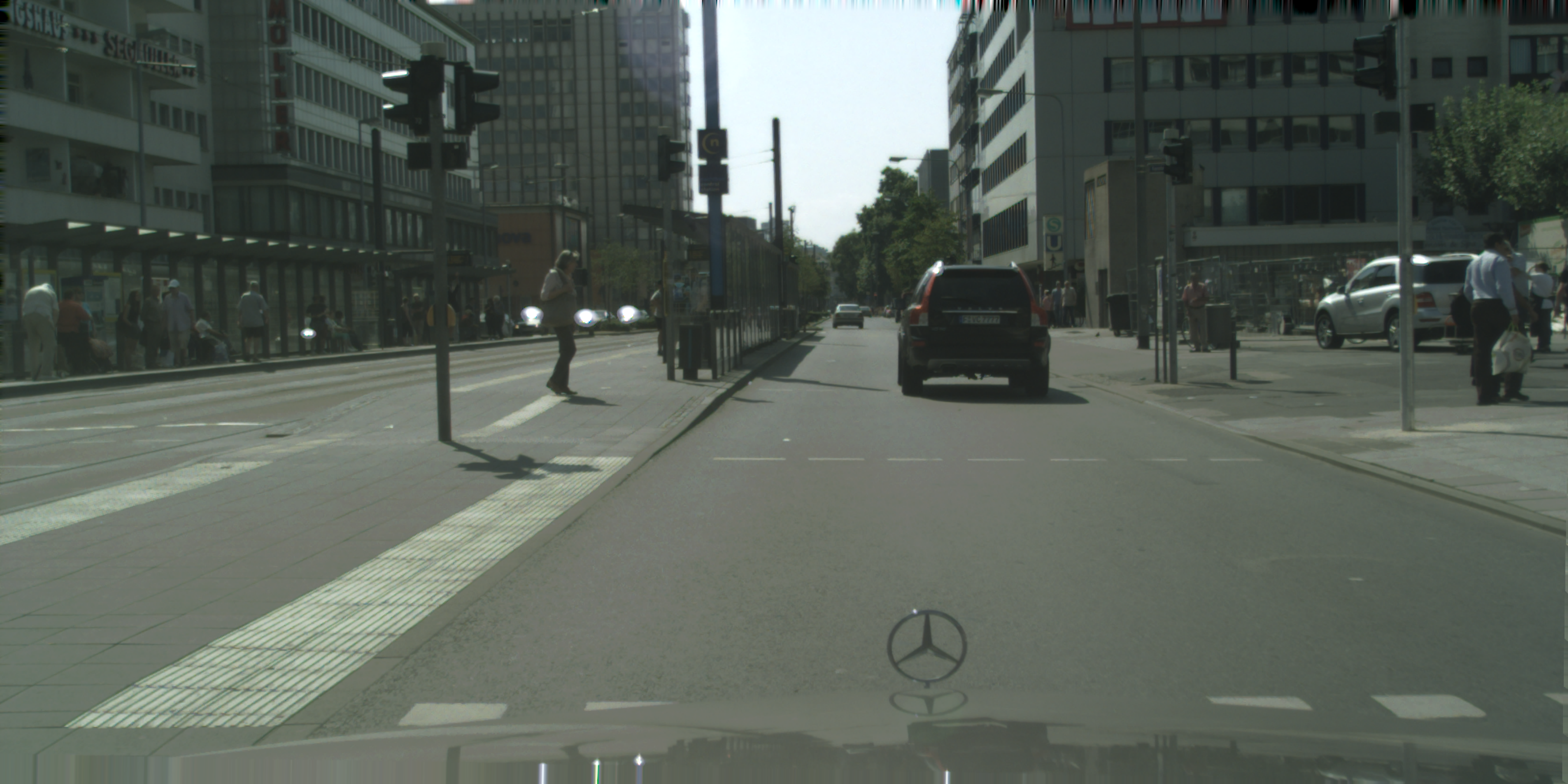}
        \includegraphics[width=\resultimagewidth\linewidth]{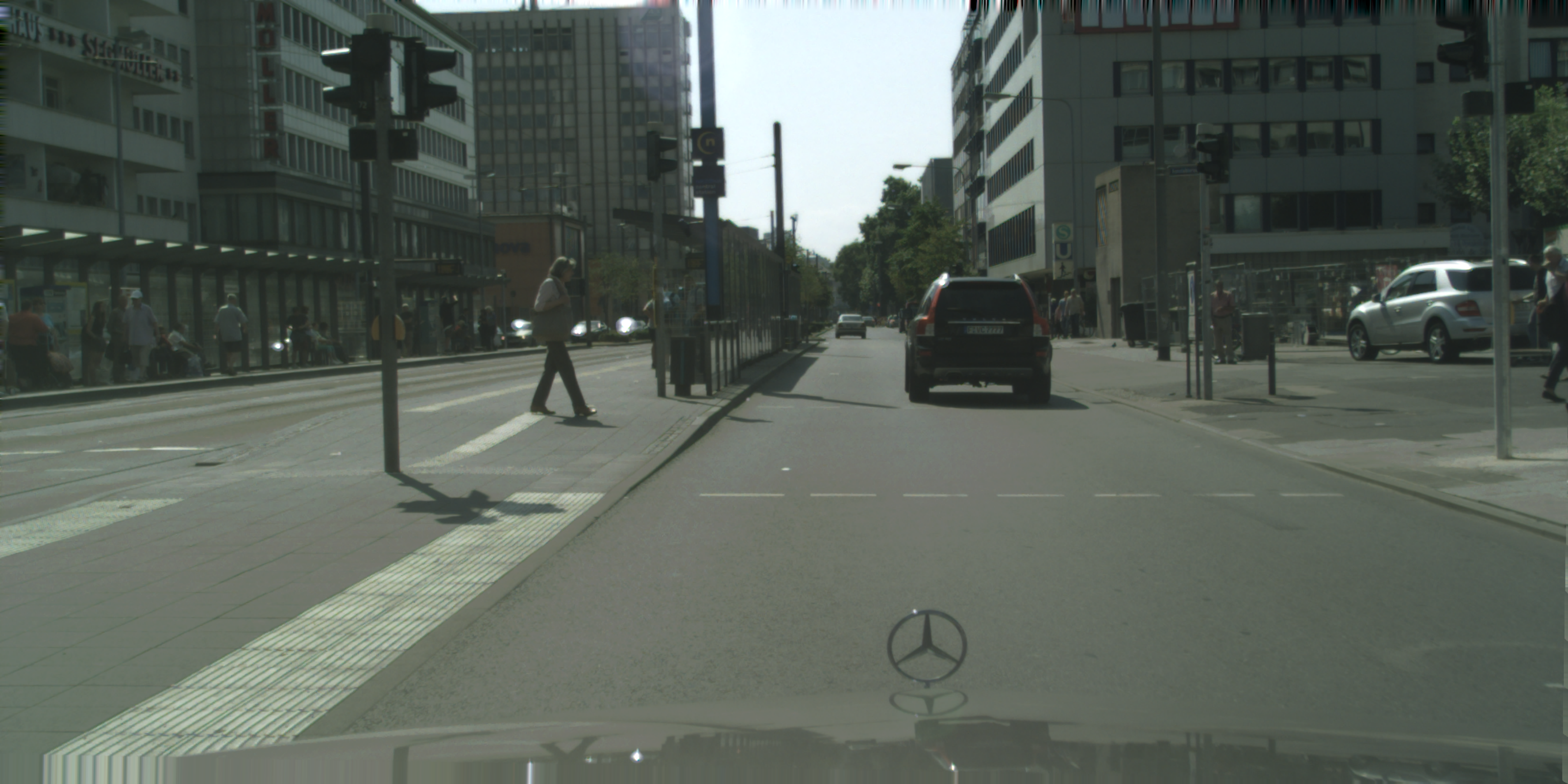}
        \includegraphics[width=\resultimagewidth\linewidth]{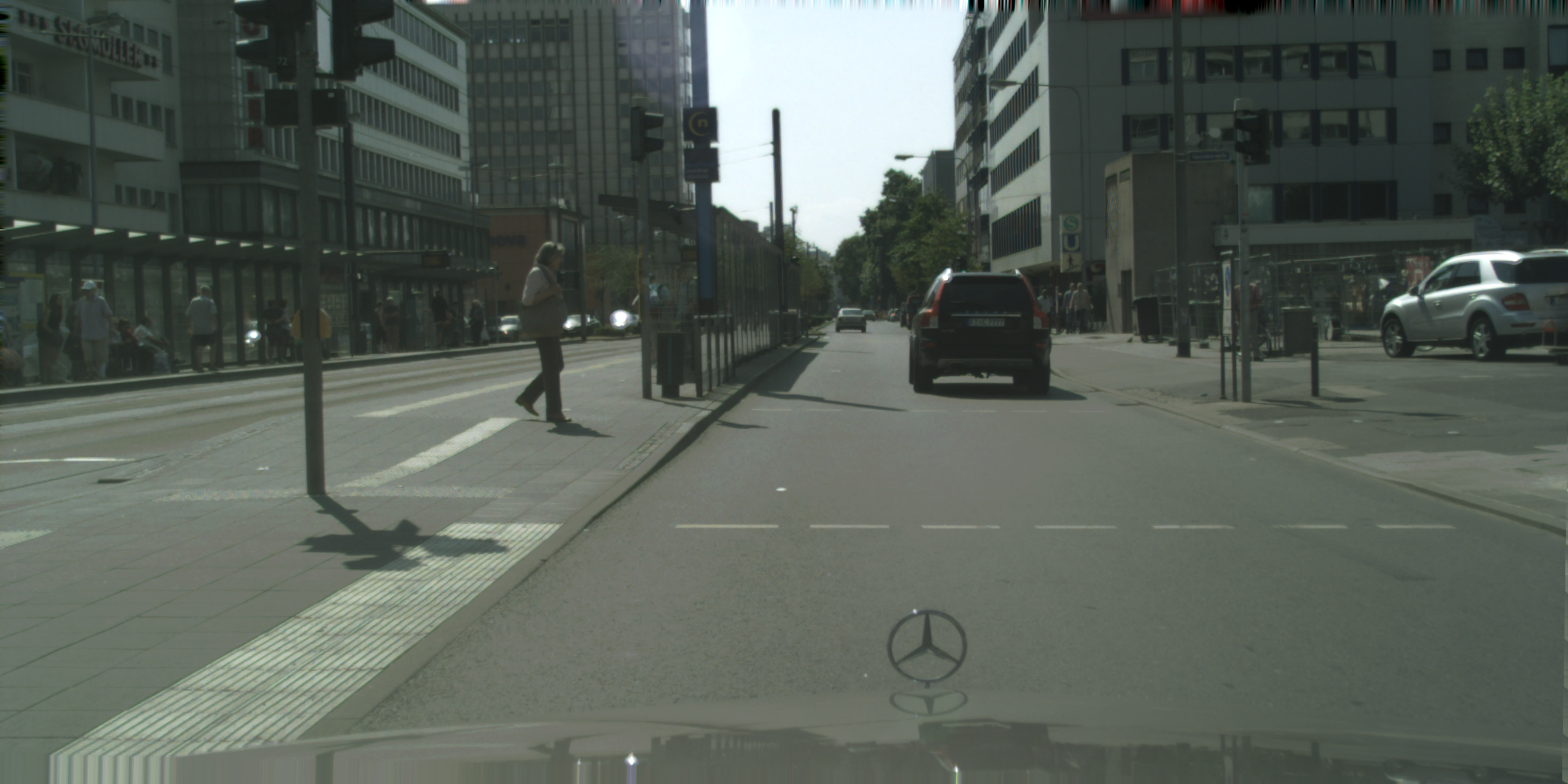}
        \includegraphics[width=\resultimagewidth\linewidth]{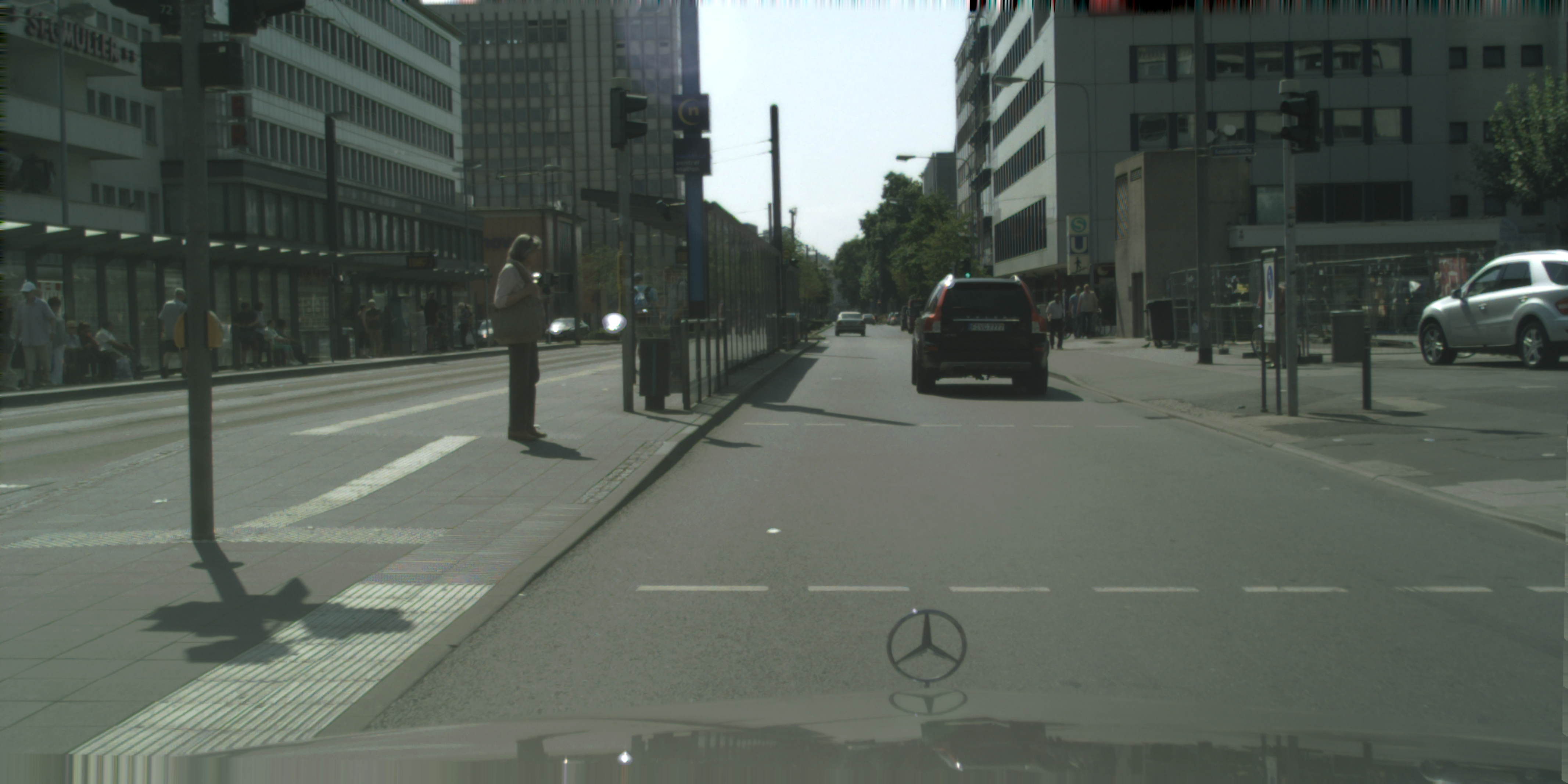}
        \includegraphics[width=\resultimagewidth\linewidth]{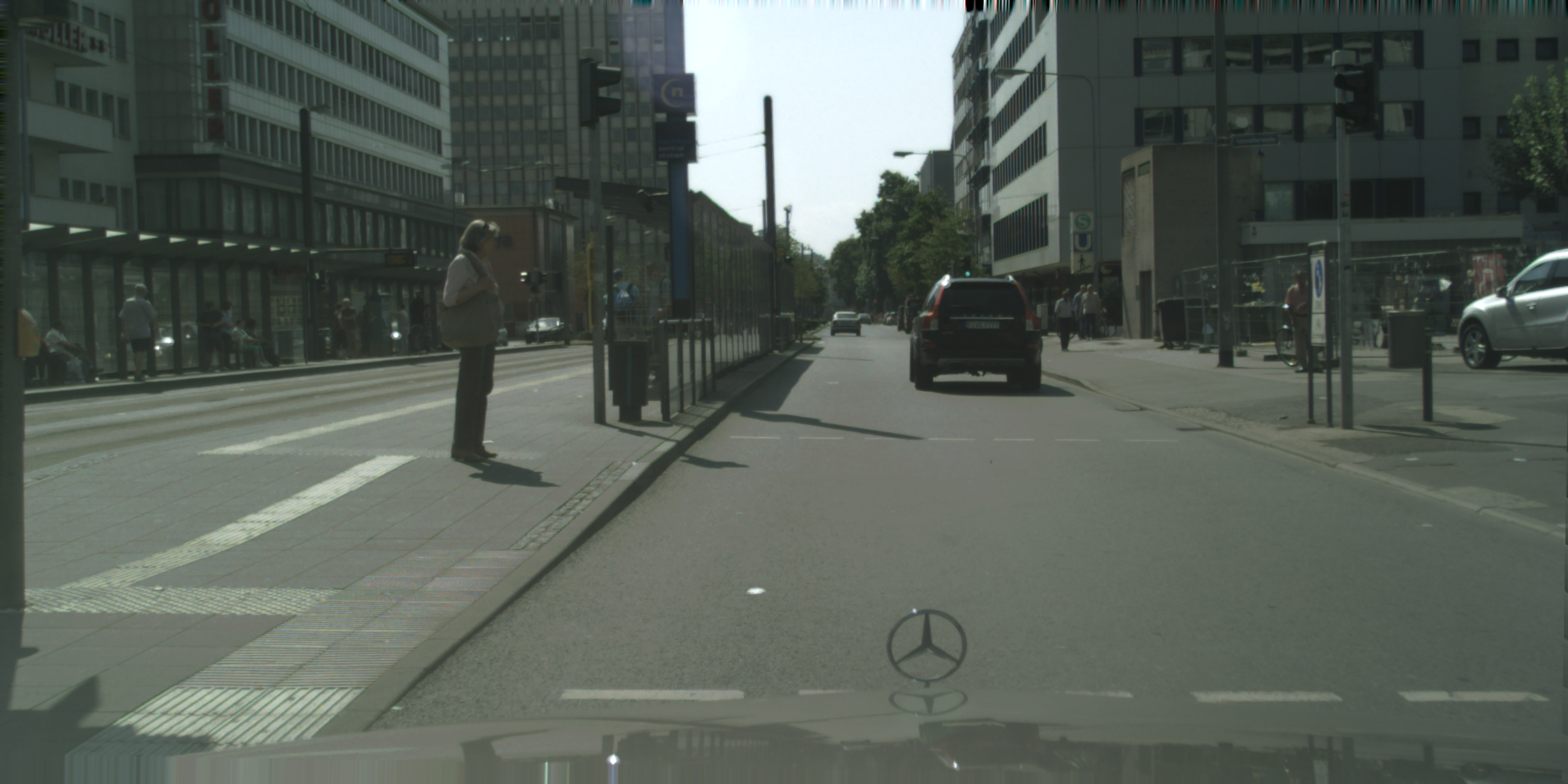}
    }
    
    \subcaptionbox{
        ground truth
    }{
        \includegraphics[width=\resultimagewidth\linewidth]{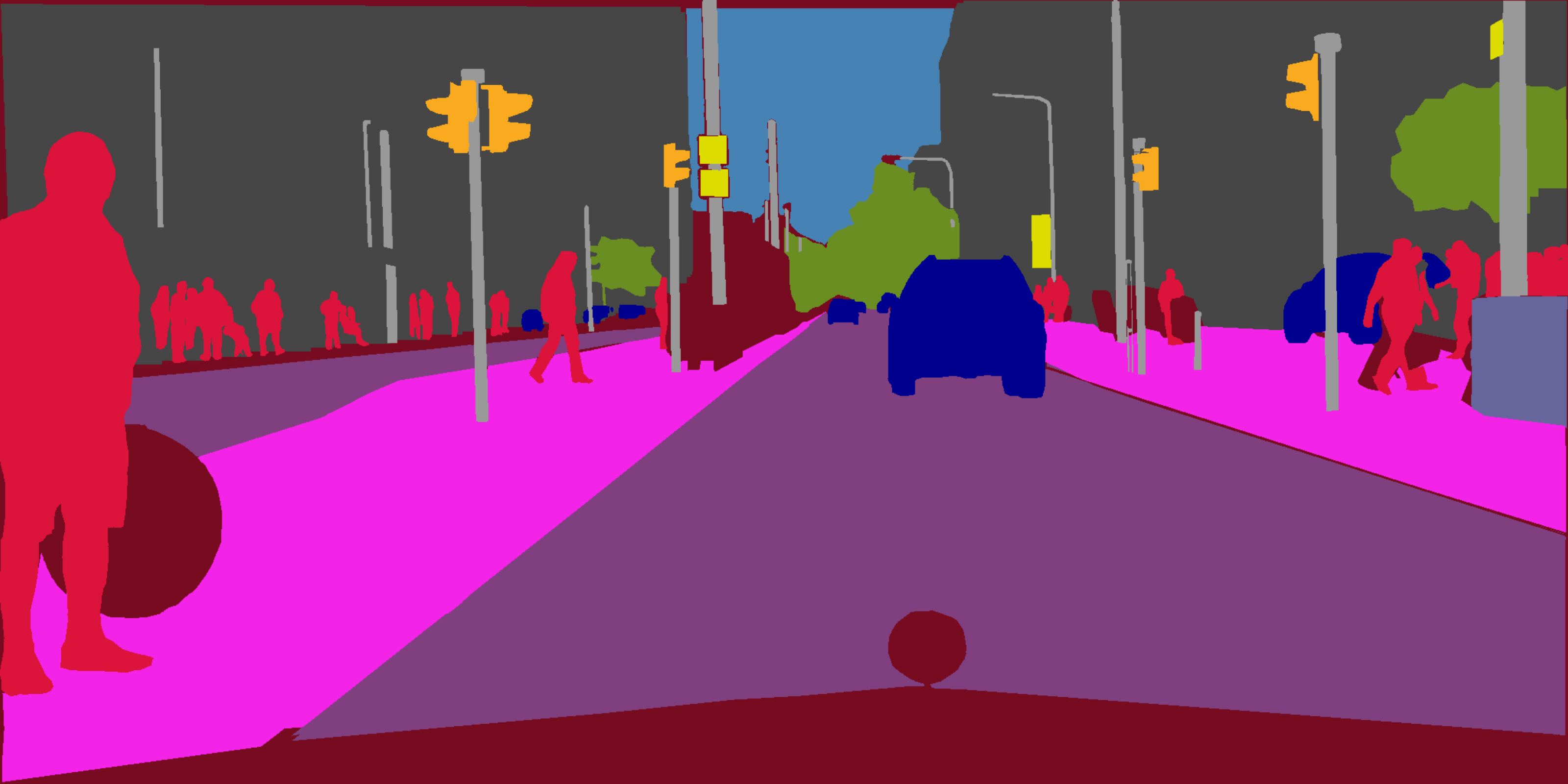}
        \includegraphics[width=\resultimagewidth\linewidth]{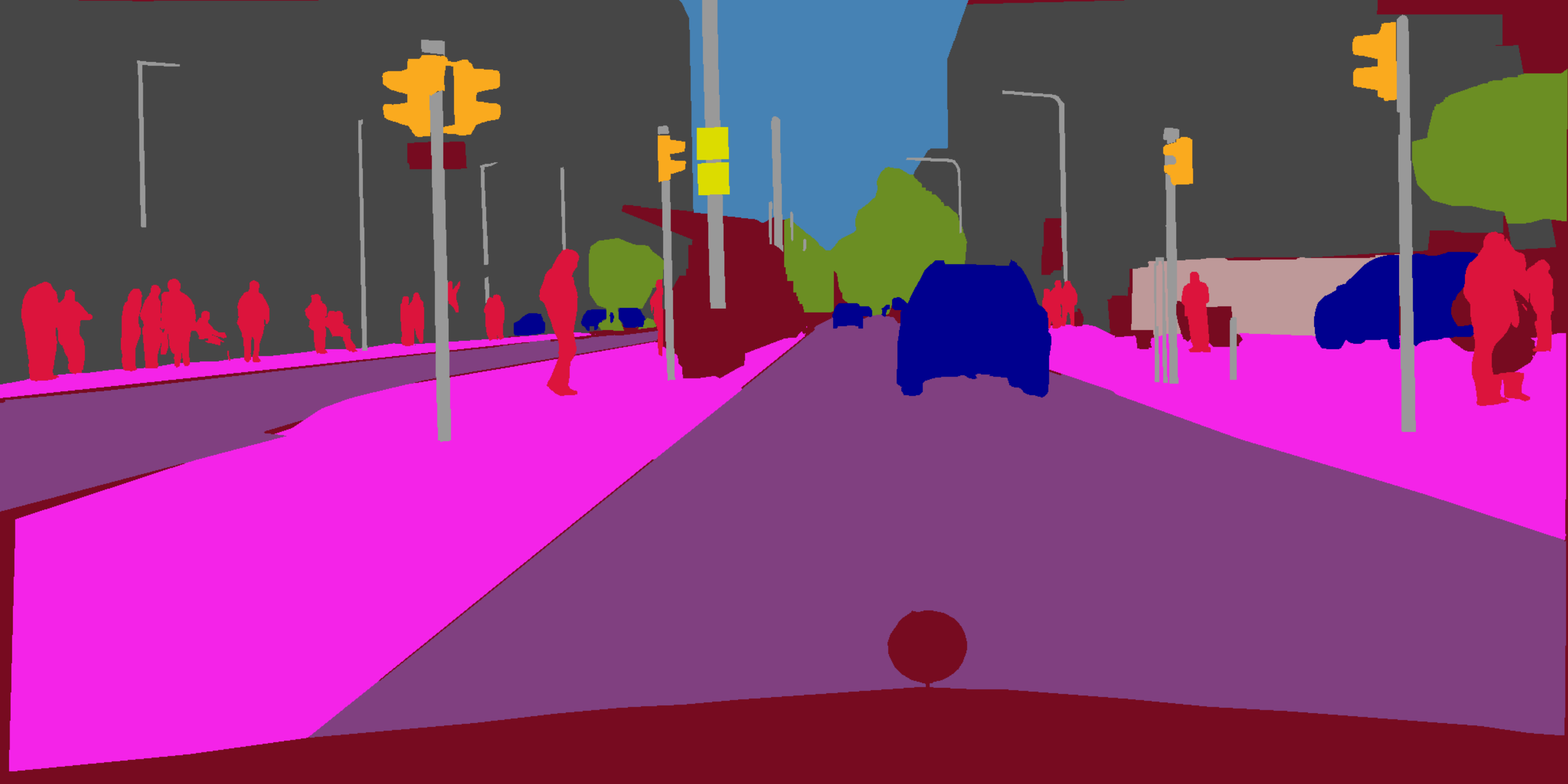}
        \includegraphics[width=\resultimagewidth\linewidth]{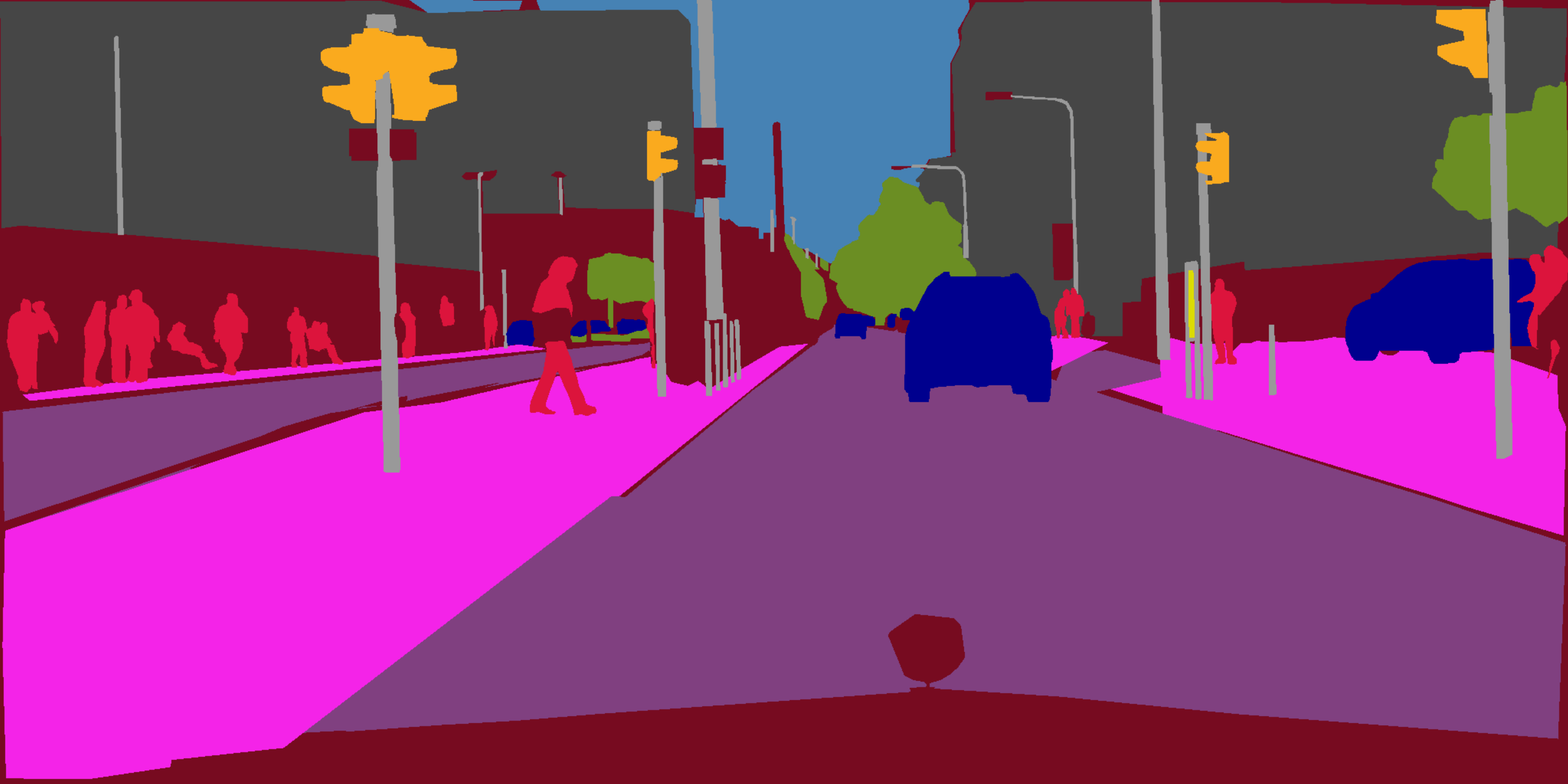}
        \includegraphics[width=\resultimagewidth\linewidth]{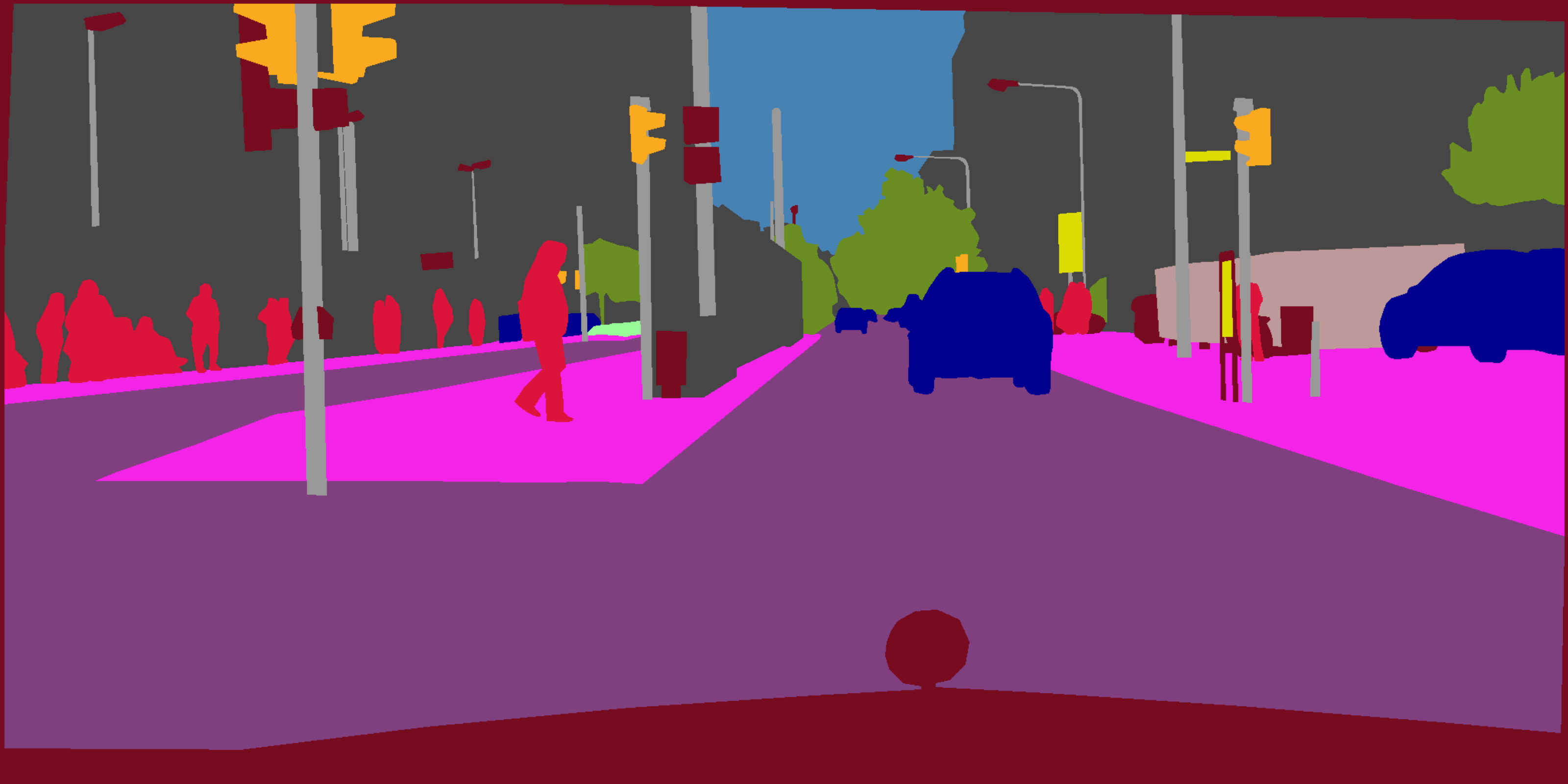}
        \includegraphics[width=\resultimagewidth\linewidth]{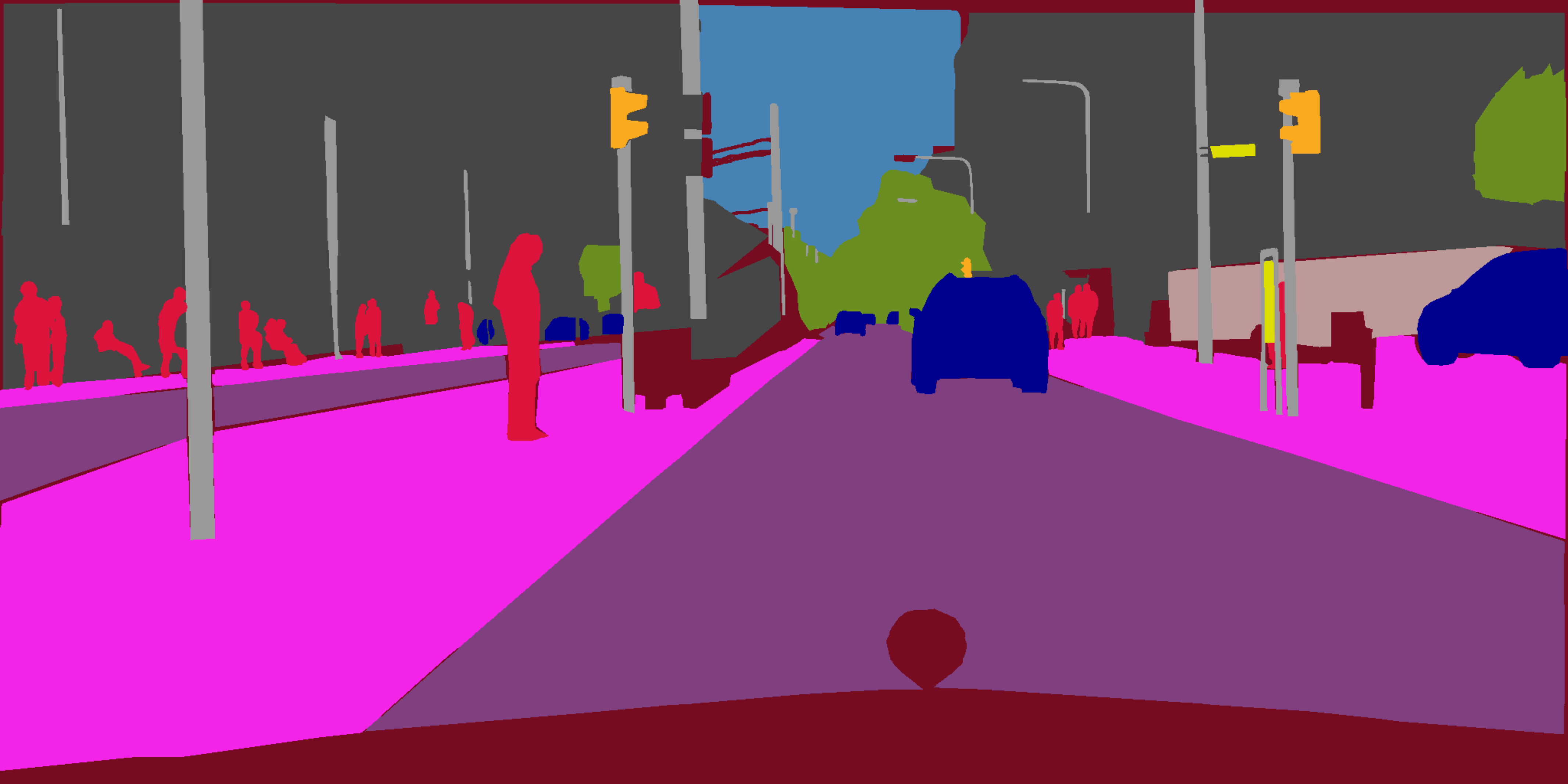}
        \includegraphics[width=\resultimagewidth\linewidth]{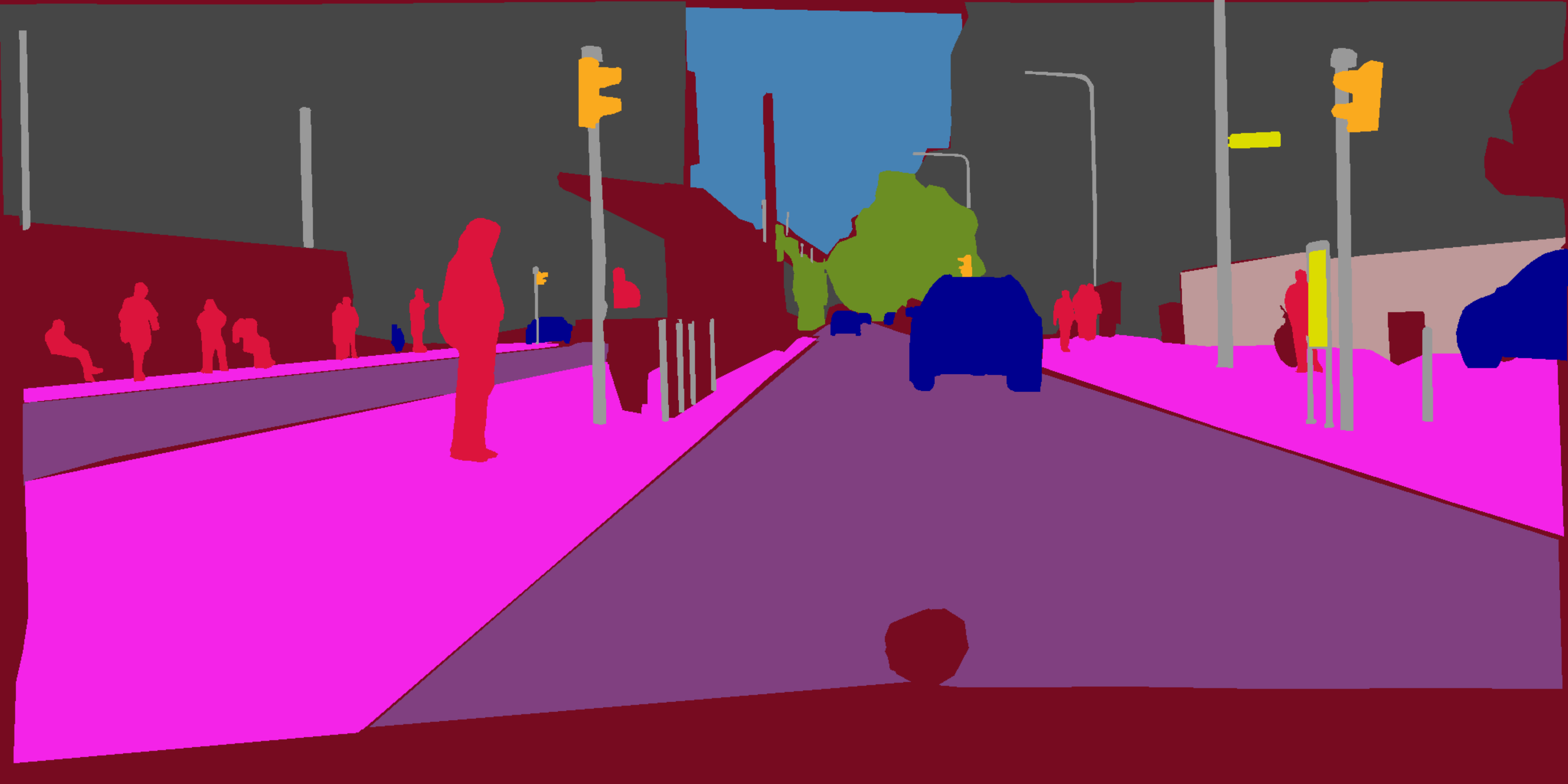}
    }
    
    \subcaptionbox{
        no shift
    }{
        \includegraphics[width=\resultimagewidth\linewidth]{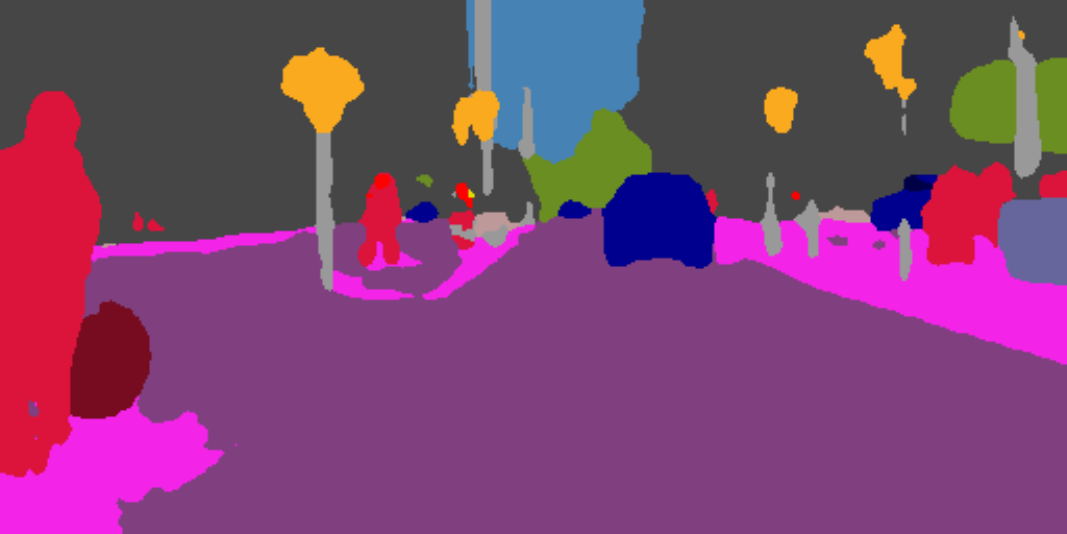}
        \includegraphics[width=\resultimagewidth\linewidth]{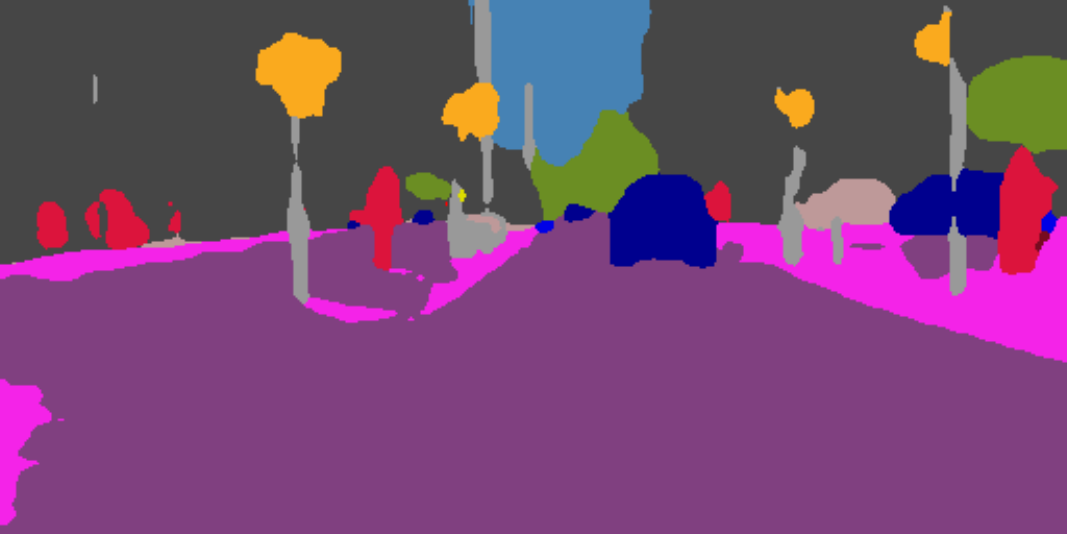}
        \includegraphics[width=\resultimagewidth\linewidth]{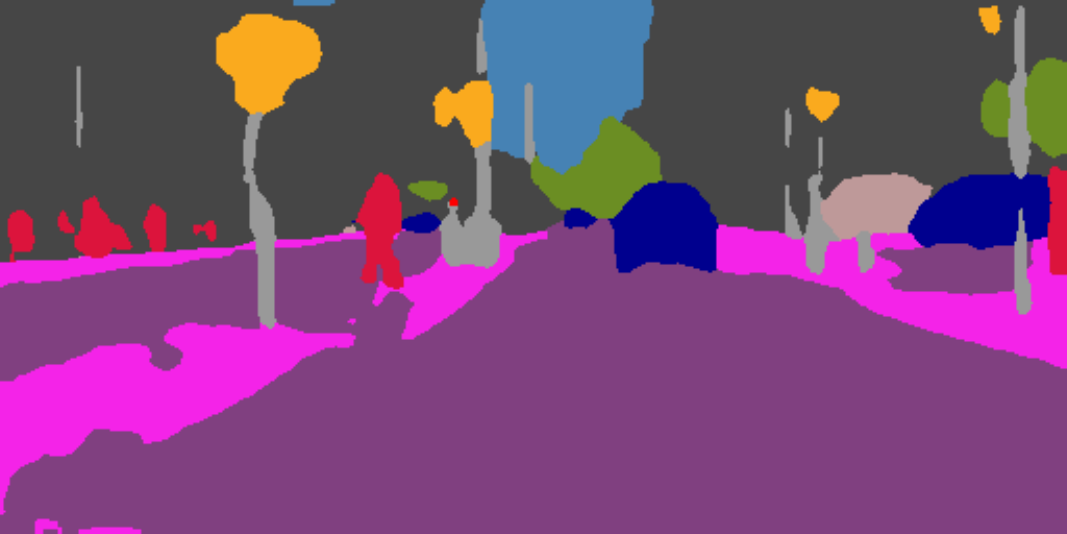}
        \includegraphics[width=\resultimagewidth\linewidth]{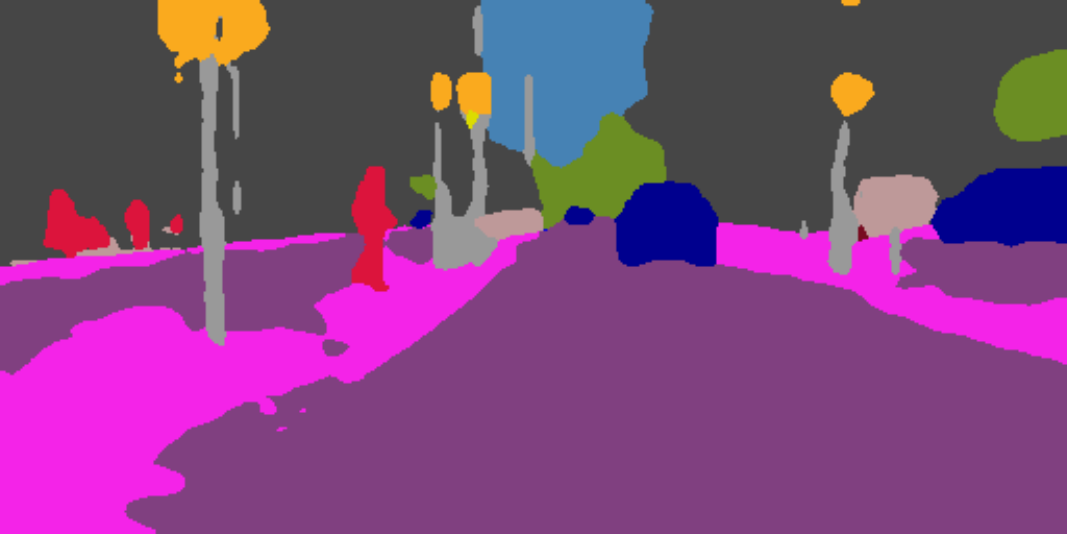}
        \includegraphics[width=\resultimagewidth\linewidth]{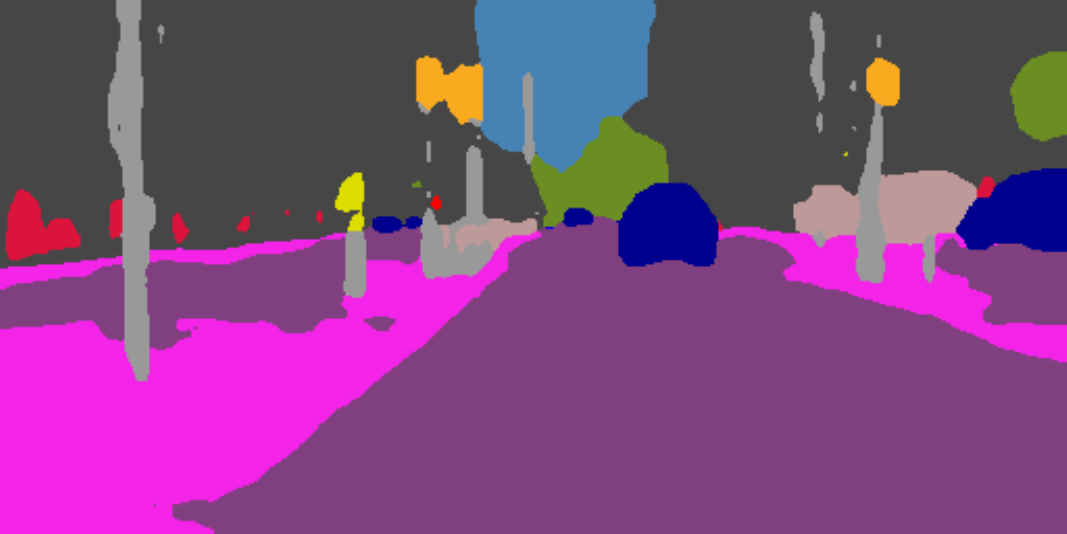}
        \includegraphics[width=\resultimagewidth\linewidth]{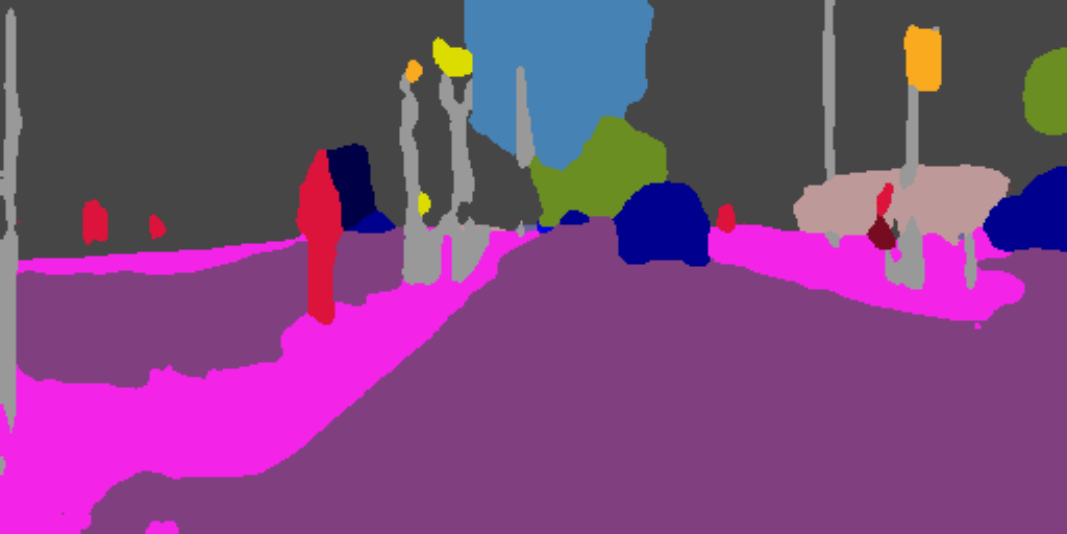}
    }

    \subcaptionbox{
        1/32 with matching
        \label{fig:cityscape 1/32 with matching}
    }{
        \includegraphics[width=\resultimagewidth\linewidth]{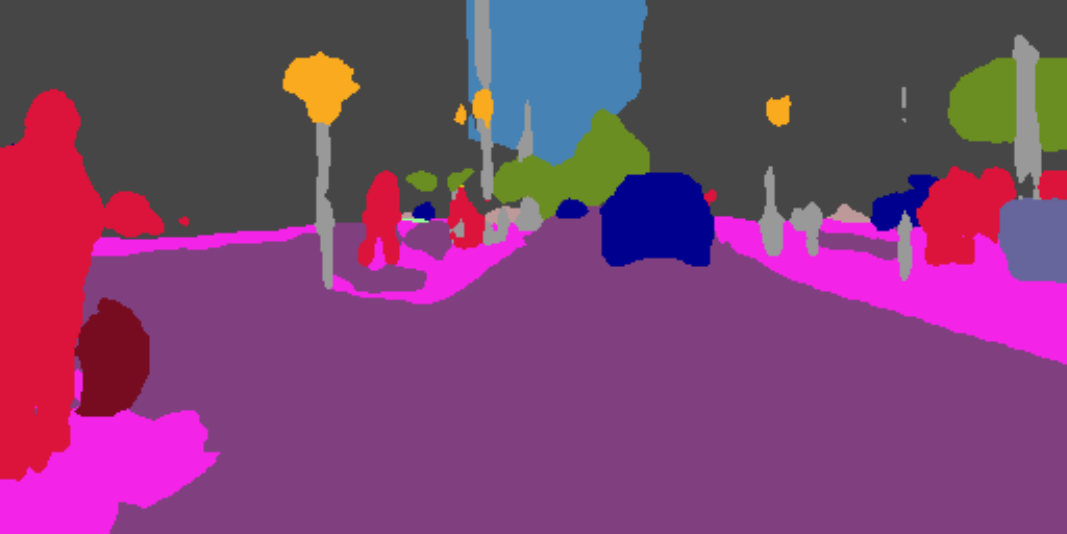}
        \includegraphics[width=\resultimagewidth\linewidth]{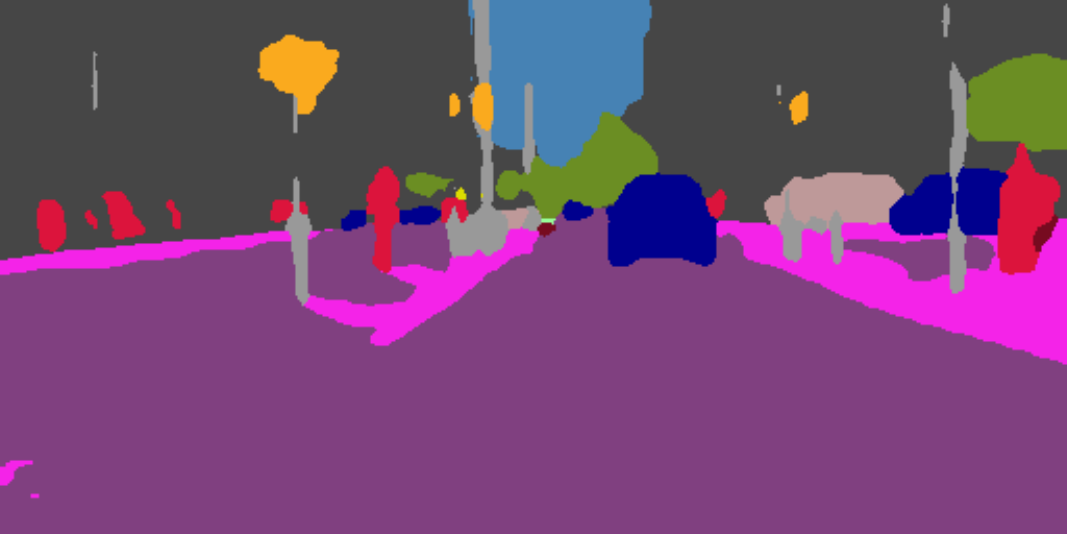}
        \includegraphics[width=\resultimagewidth\linewidth]{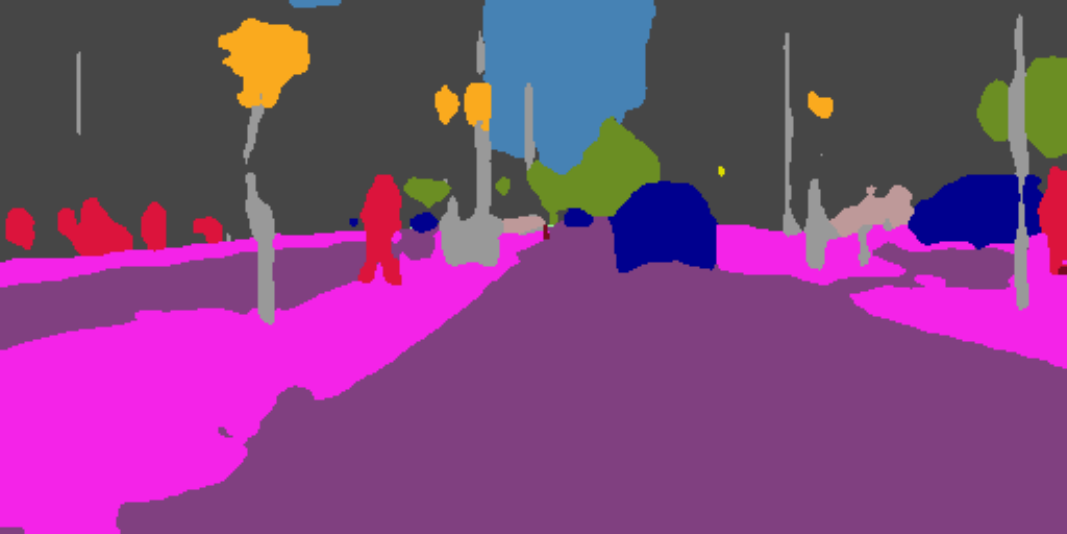}
        \includegraphics[width=\resultimagewidth\linewidth]{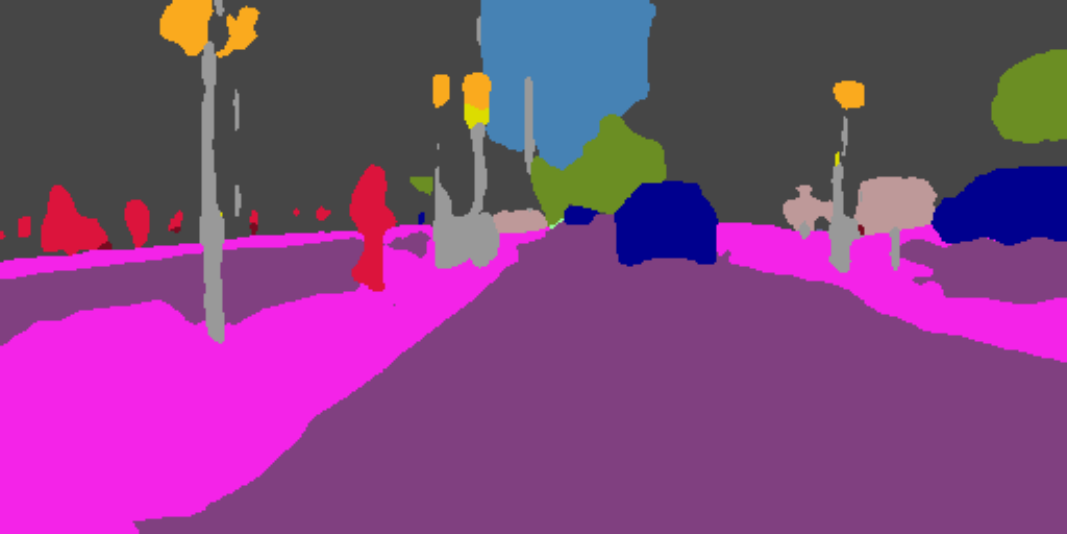}
        \includegraphics[width=\resultimagewidth\linewidth]{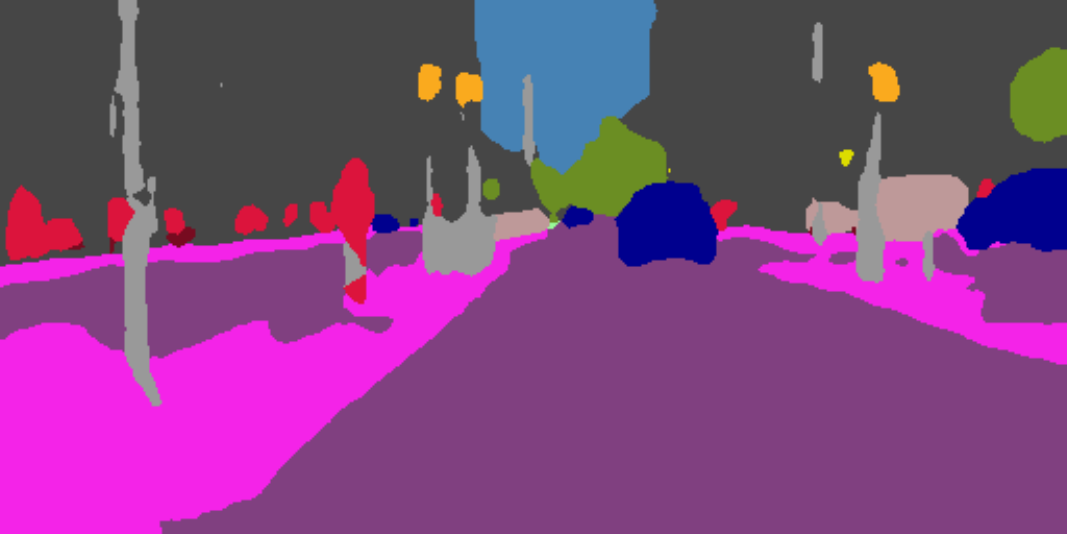}
        \includegraphics[width=\resultimagewidth\linewidth]{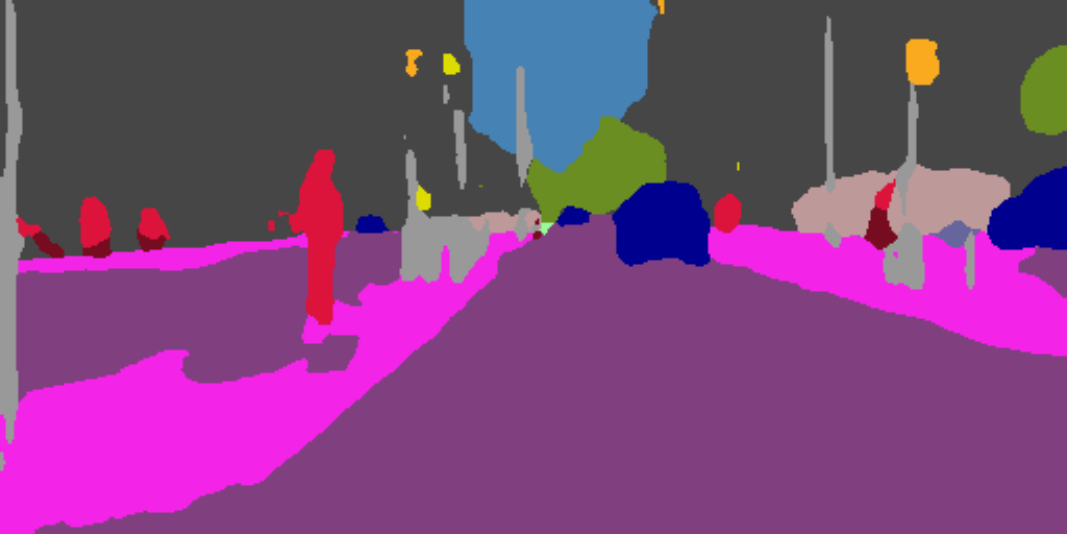}
    }
    
    \subcaptionbox{
        1/32 without matching
        \label{fig:cityscape 1/32 without matching}
    }{
        \includegraphics[width=\resultimagewidth\linewidth]{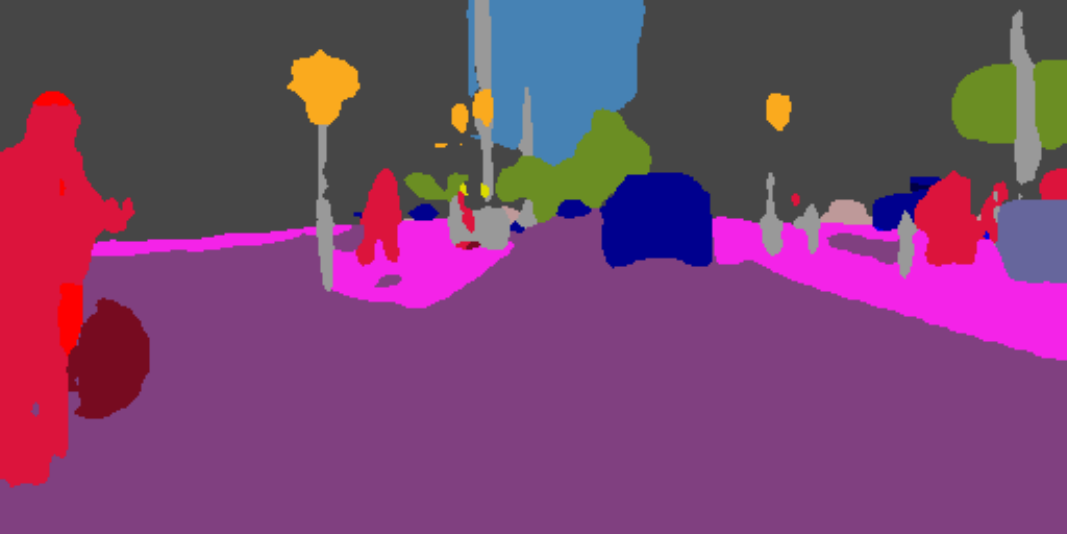}
        \includegraphics[width=\resultimagewidth\linewidth]{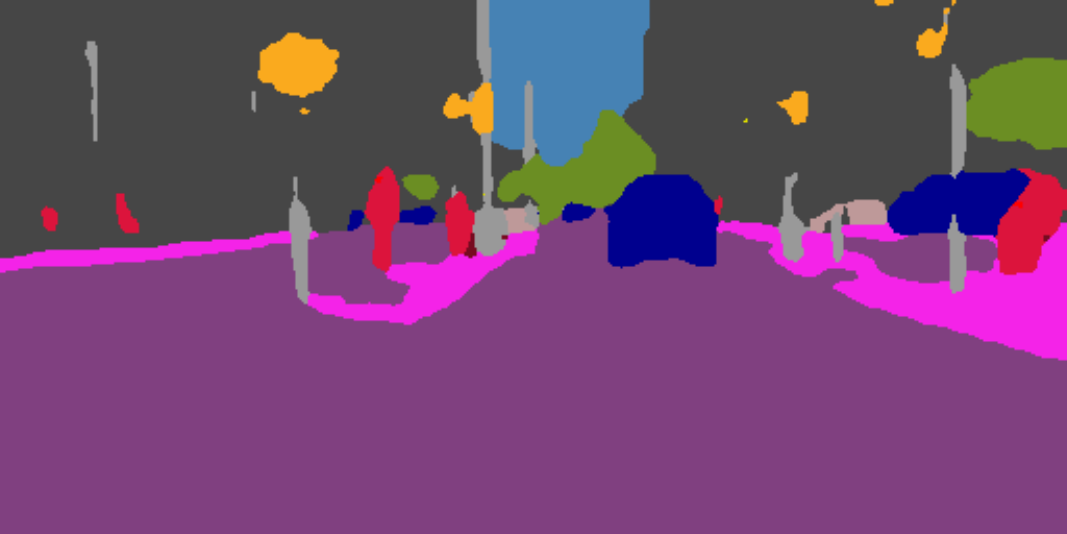}
        \includegraphics[width=\resultimagewidth\linewidth]{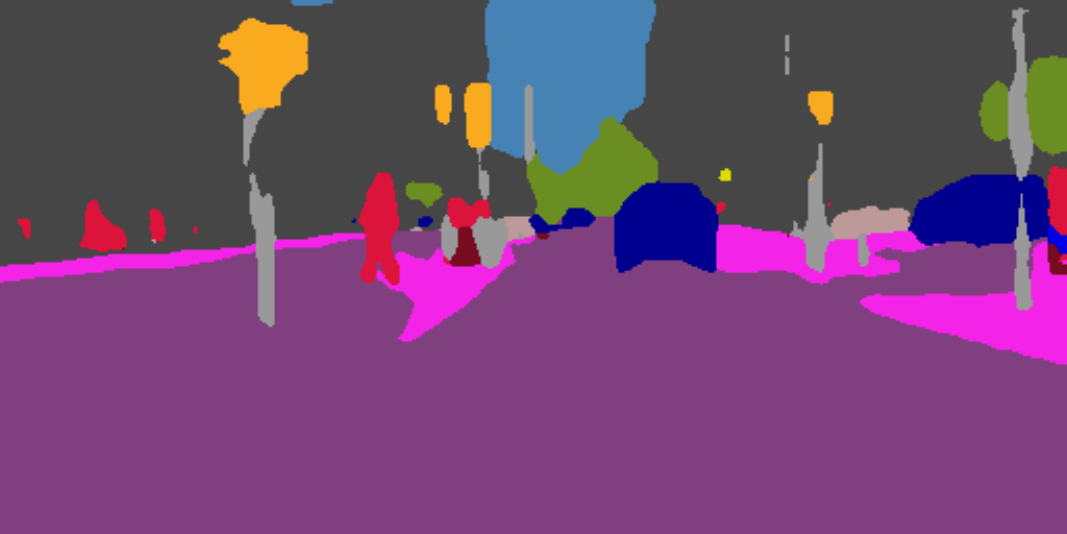}
        \includegraphics[width=\resultimagewidth\linewidth]{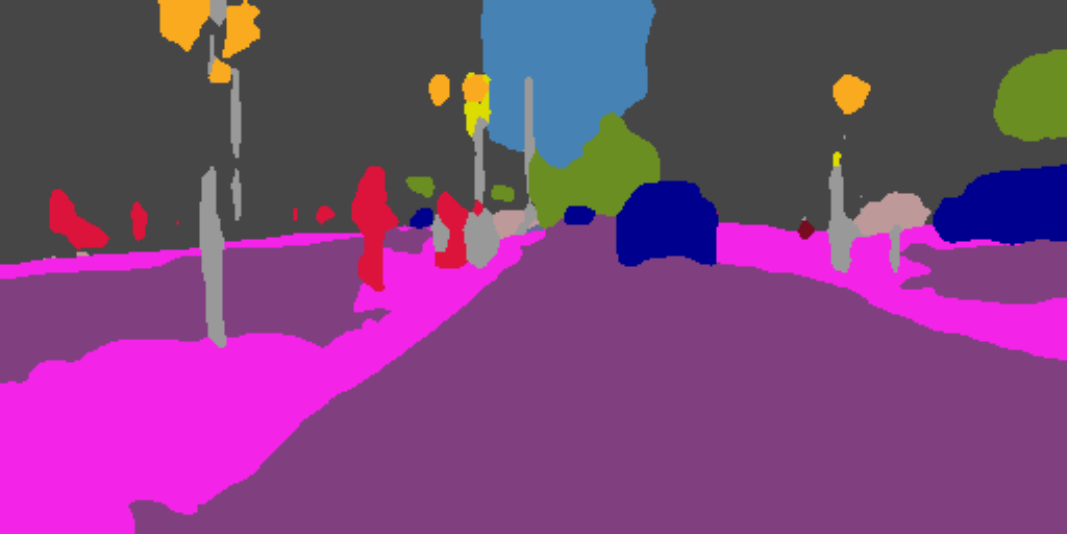}
        \includegraphics[width=\resultimagewidth\linewidth]{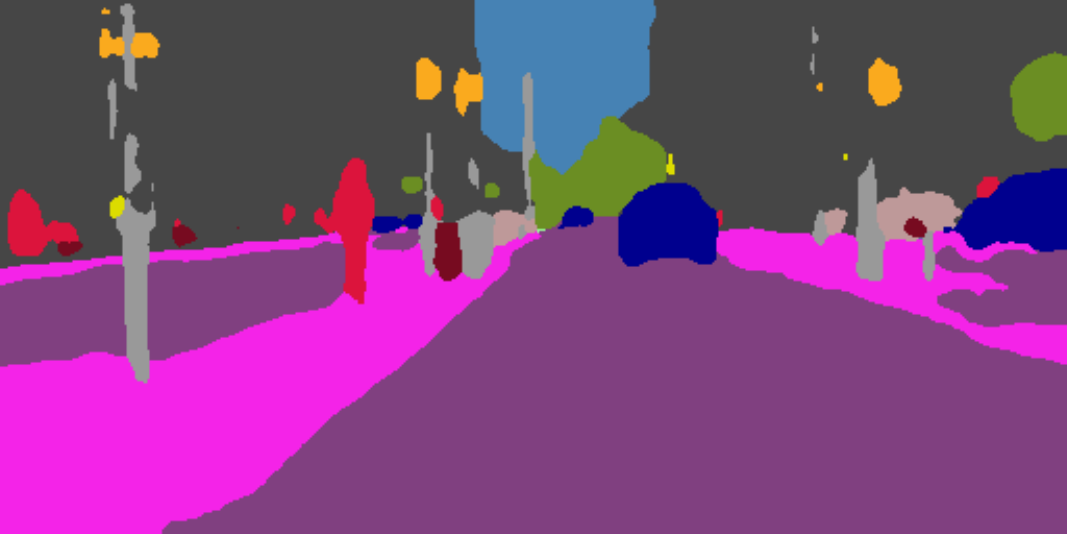}
        \includegraphics[width=\resultimagewidth\linewidth]{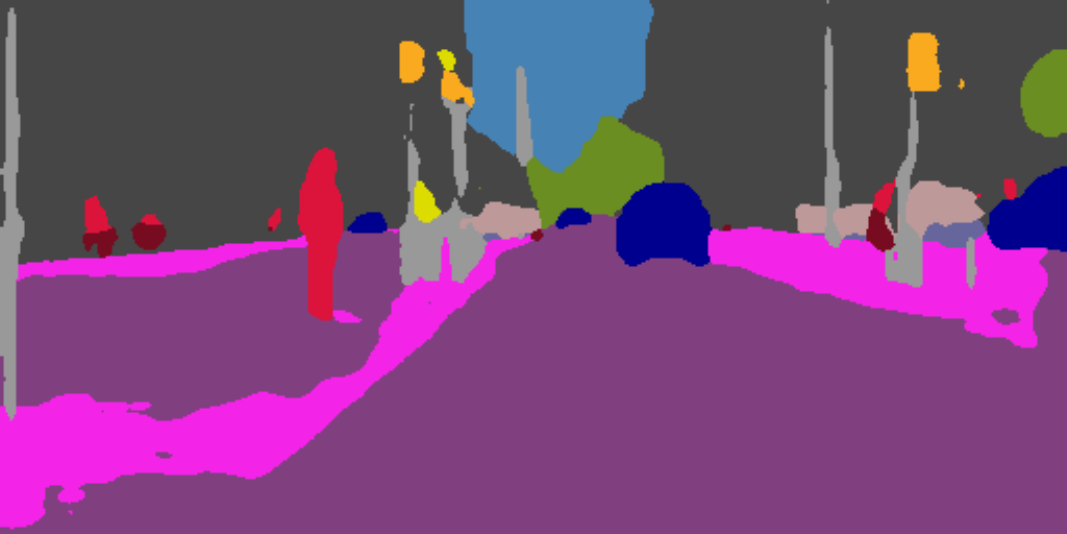}
    }
    
    \subcaptionbox{
        1/16 with matching
        \label{fig:cityscape 1/16 with matching}
    }{
        \includegraphics[width=\resultimagewidth\linewidth]{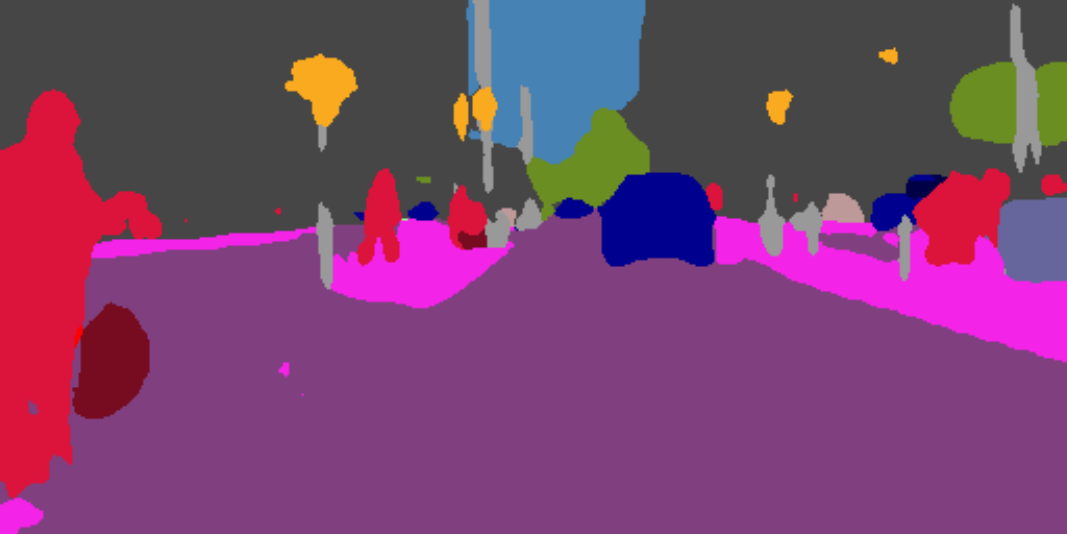}
        \includegraphics[width=\resultimagewidth\linewidth]{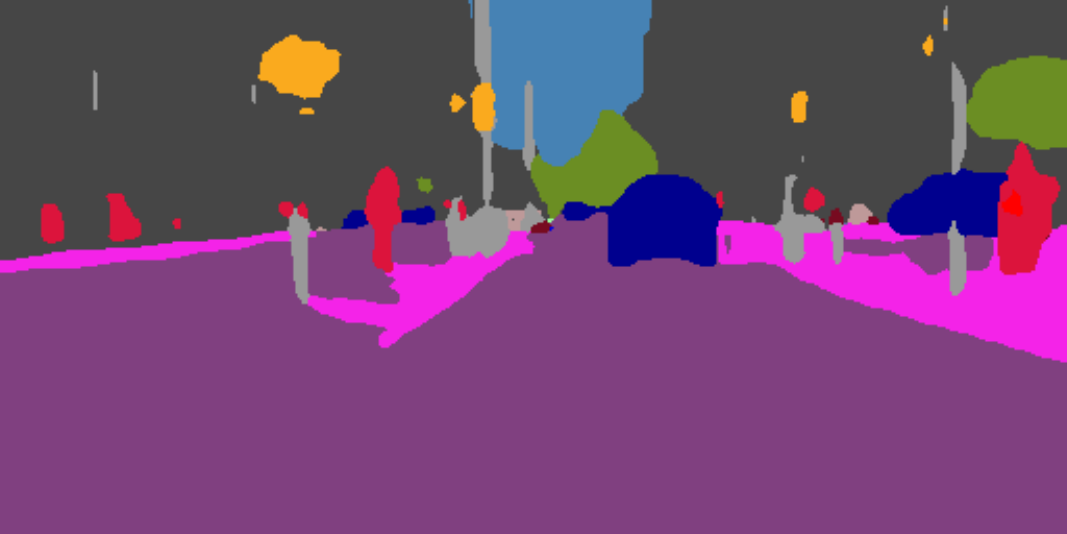}
        \includegraphics[width=\resultimagewidth\linewidth]{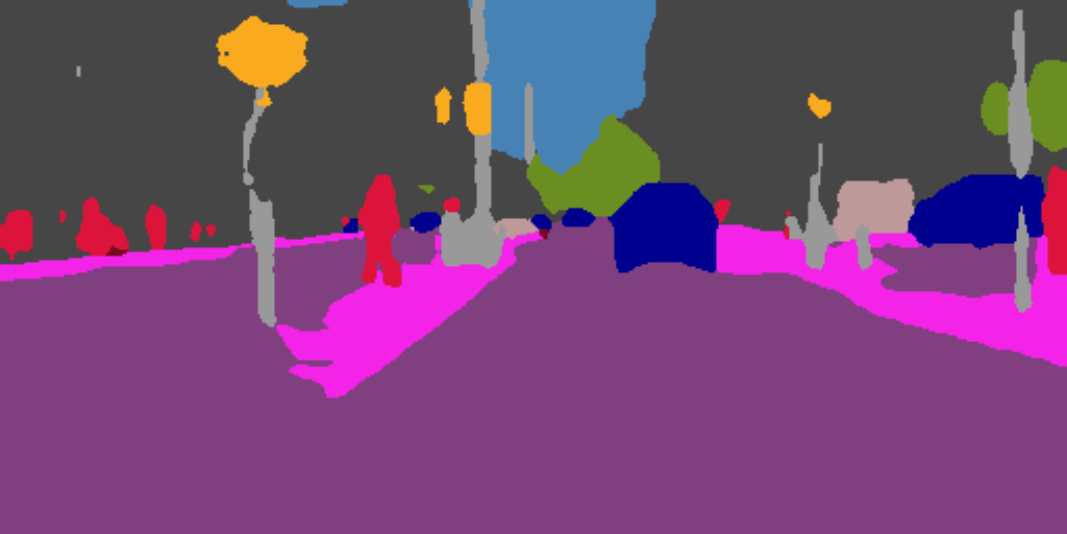}
        \includegraphics[width=\resultimagewidth\linewidth]{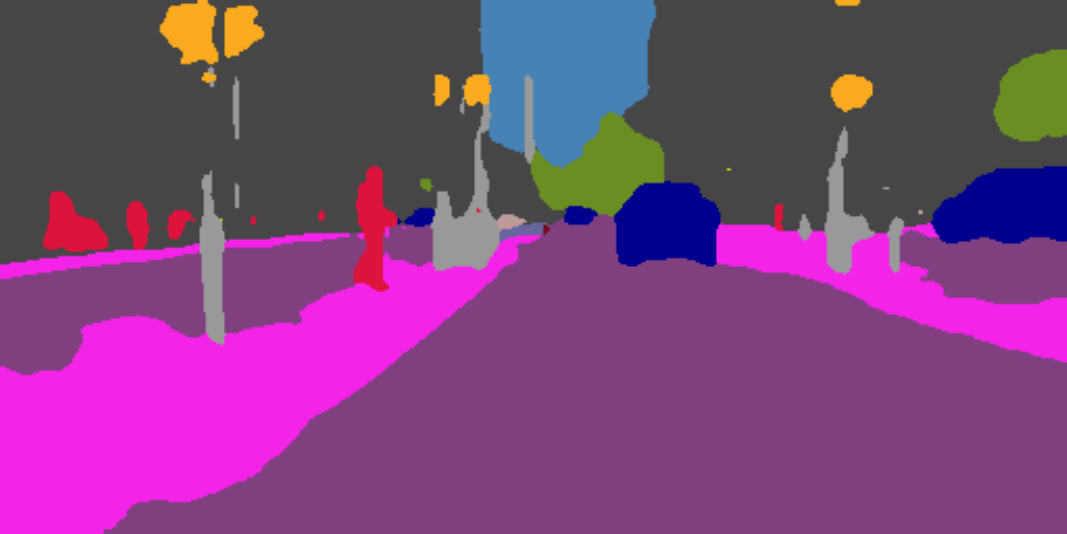}
        \includegraphics[width=\resultimagewidth\linewidth]{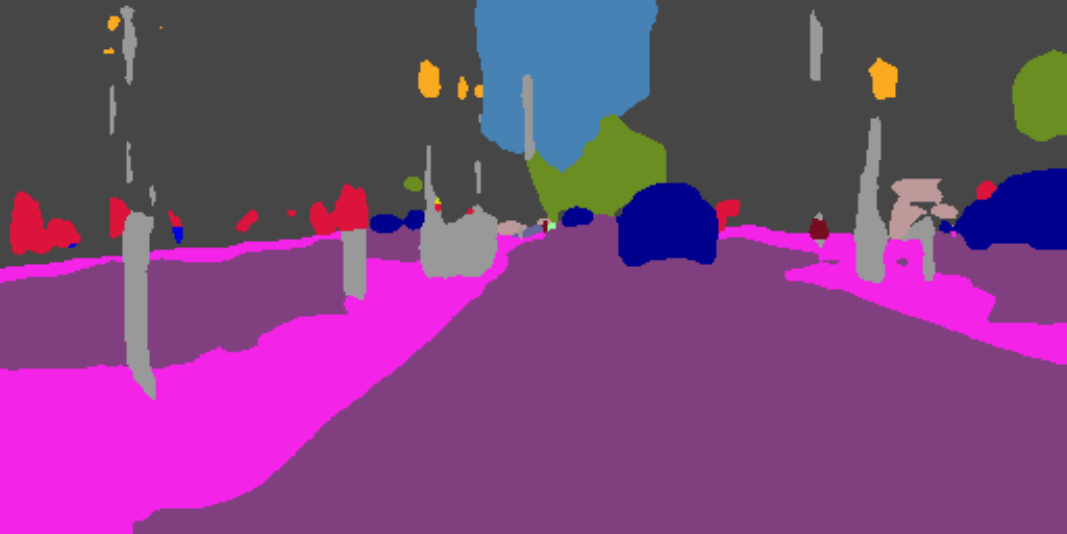}
        \includegraphics[width=\resultimagewidth\linewidth]{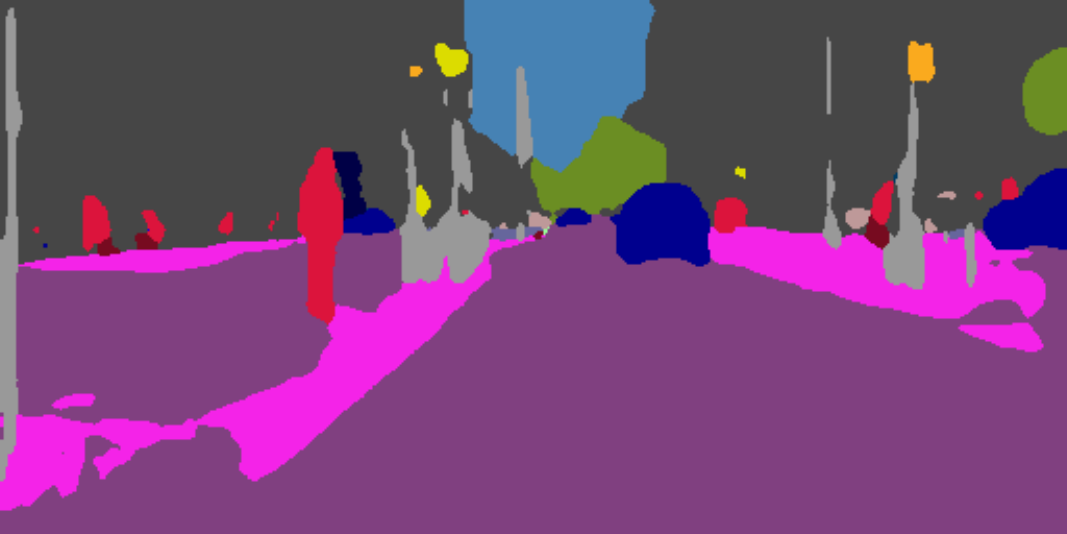}
    }

    \subcaptionbox{
        1/16 without matching
        \label{fig:cityscape 1/16 without matching}
    }{
        \includegraphics[width=\resultimagewidth\linewidth]{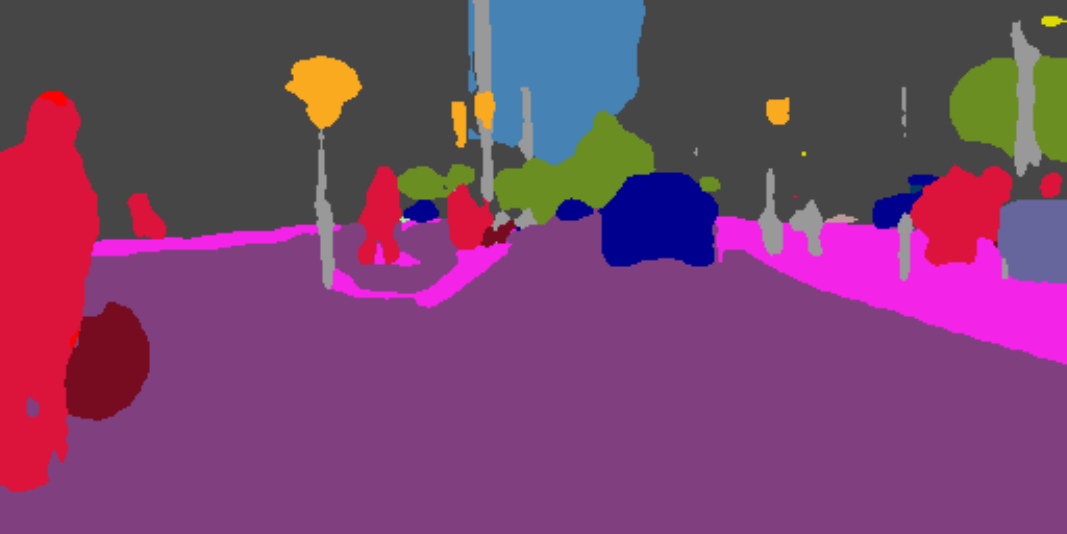}
        \includegraphics[width=\resultimagewidth\linewidth]{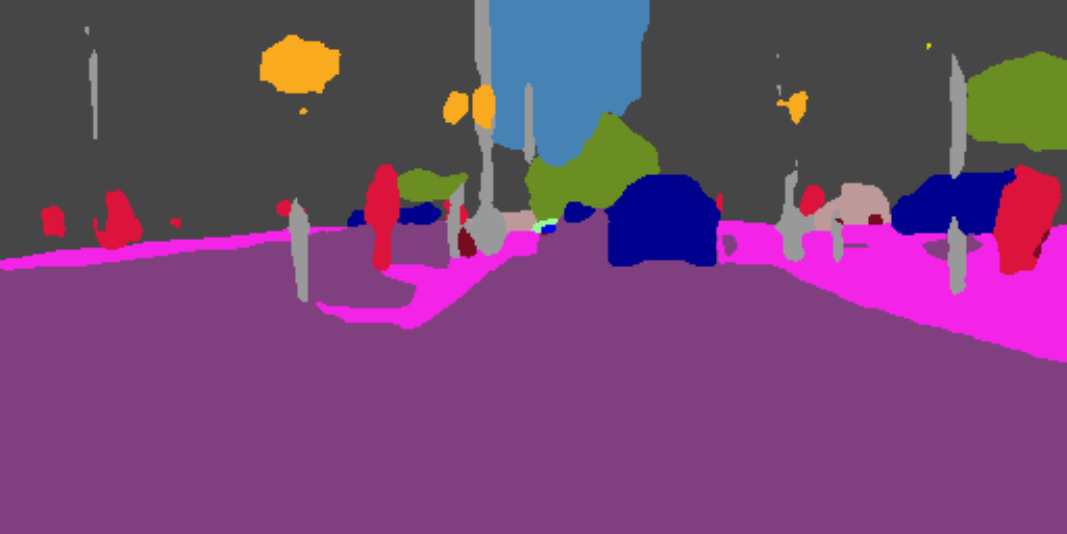}
        \includegraphics[width=\resultimagewidth\linewidth]{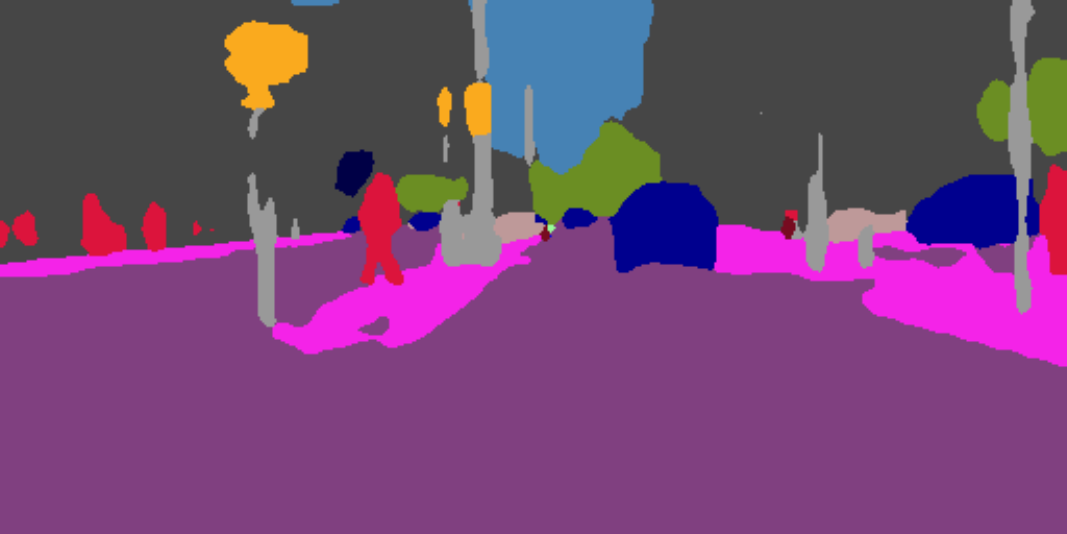}
        \includegraphics[width=\resultimagewidth\linewidth]{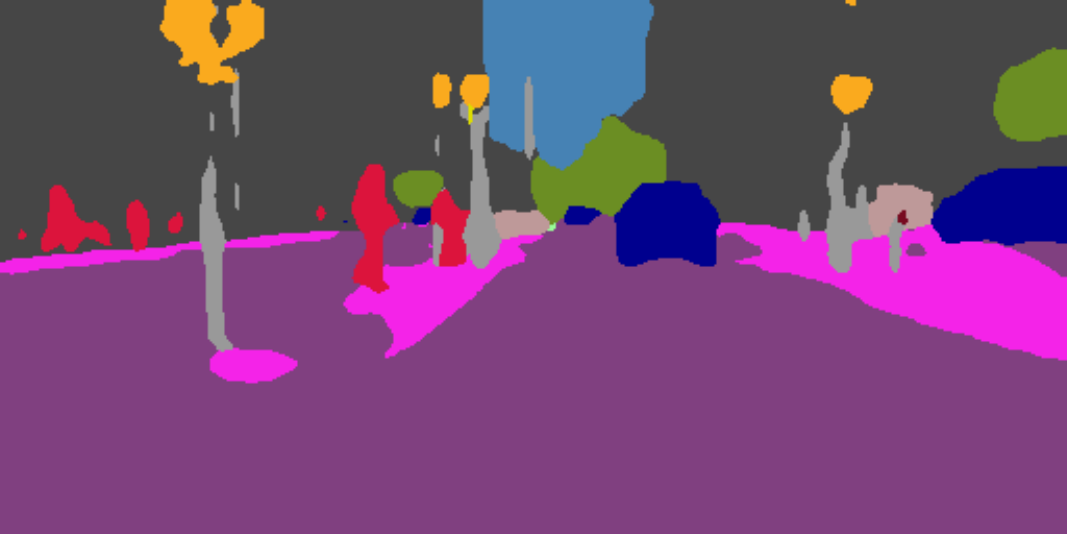}
        \includegraphics[width=\resultimagewidth\linewidth]{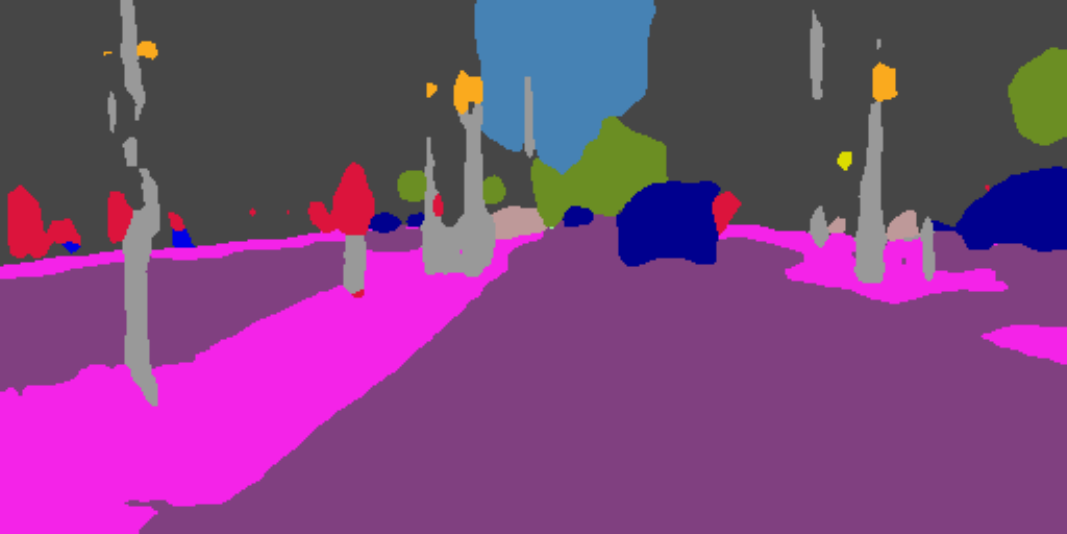}
        \includegraphics[width=\resultimagewidth\linewidth]{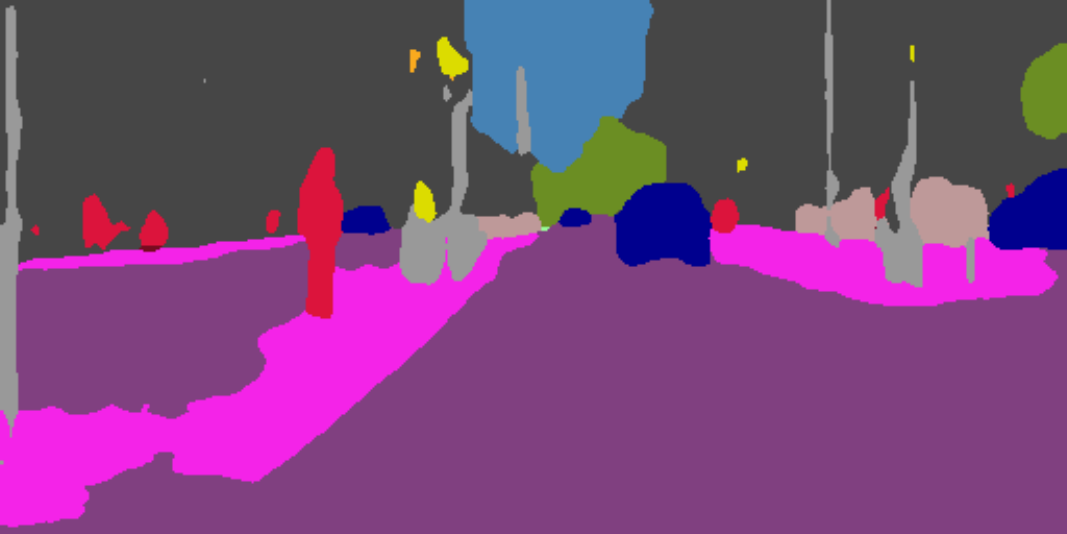}
    }

    \caption{
        Segmentation results for CityScapes-VPS.
        First two rows show
        (a) original frames, and 
        (b) ground truth segmentation labels.
        The following rows show results of
        (c) the baseline (no shifts, no query matching),
        (d) 1/32 shift with and (e) without matching,
        (f) 1/16 shift with and (g) without matching.
    }
    \label{segmentation_result_1}

\end{figure*}

Figure~\ref{segmentation_result_1} shows examples of segmentation results for CityScapes-VPS with and without query matching for a shift of 1/32 and 1/16.
As seen in Table~\ref{tab:performance comparison cityscapes-vps}, applying the proposed query matching improves performance, particularly in the ``sidewalk'' category. Without query matching, a large part of the sidewalk is missing in the fourth frame of Fig.~\ref{fig:cityscape 1/16 without matching} and the third frame of Fig.~\ref{fig:cityscape 1/32 without matching}. These missing parts are improved with query matching, as shown in Fig.~\ref{fig:cityscape 1/16 with matching} and Fig.~\ref{fig:cityscape 1/32 with matching}.
In the first frame of the shift of 1/16, the small ``person'' region (in red), visible far ahead (or on the right side in the image) of the car (blue) in the center, was misidentified as a ``tree'' (green) without matching (Fig.~\ref{fig:cityscape 1/16 without matching}). This was improved with matching, as shown in Fig.~\ref{fig:cityscape 1/16 with matching}, which demonstrates the effect of the feature shift between the same objects across the frames by the proposed query matching.

\begin{figure*}[t]
    \centering
    \def\resultimagewidth{0.15}
    
    \subcaptionbox{
        original frames
    }{
        \includegraphics[width=\resultimagewidth\linewidth]{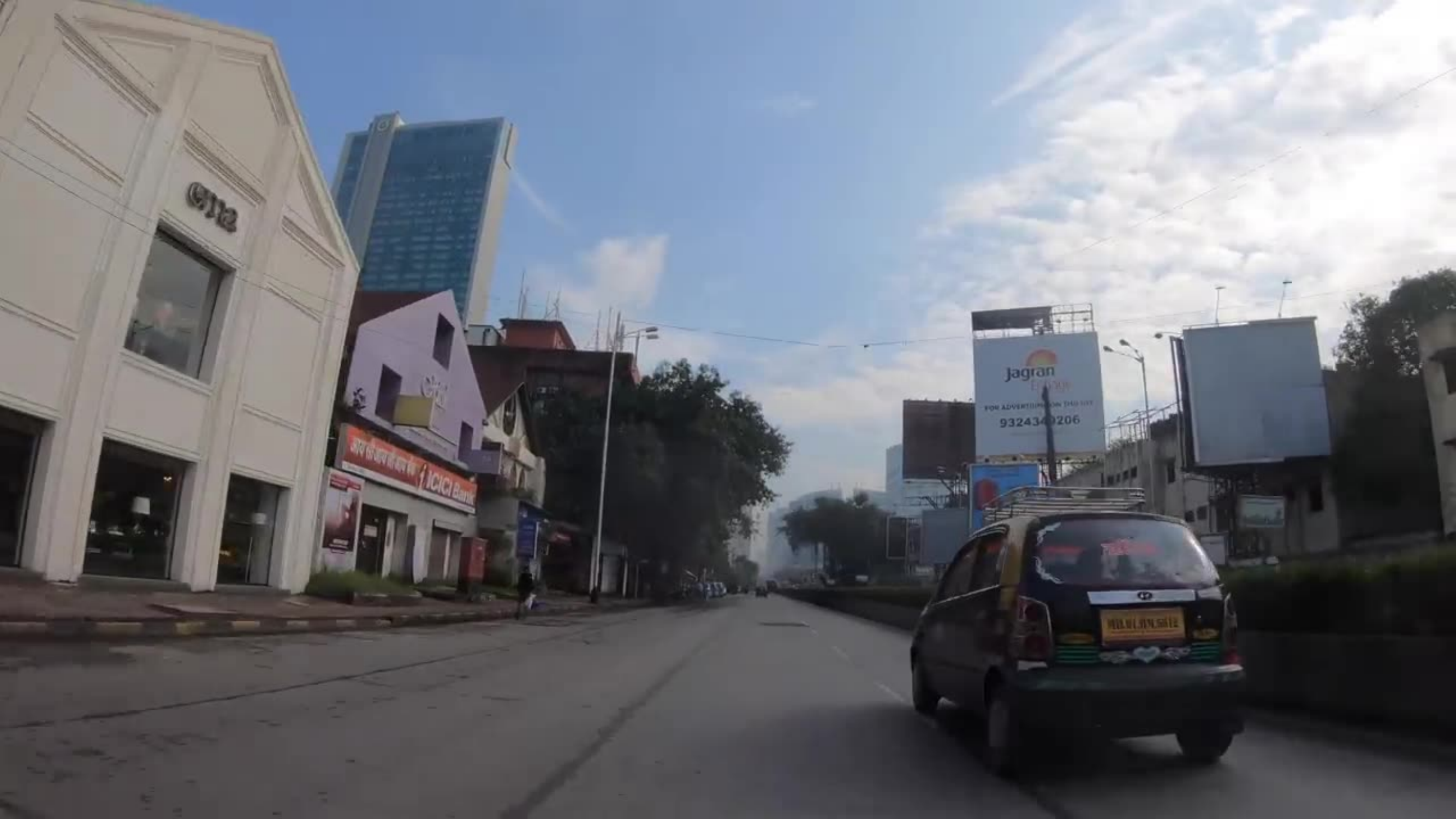}
        \includegraphics[width=\resultimagewidth\linewidth]{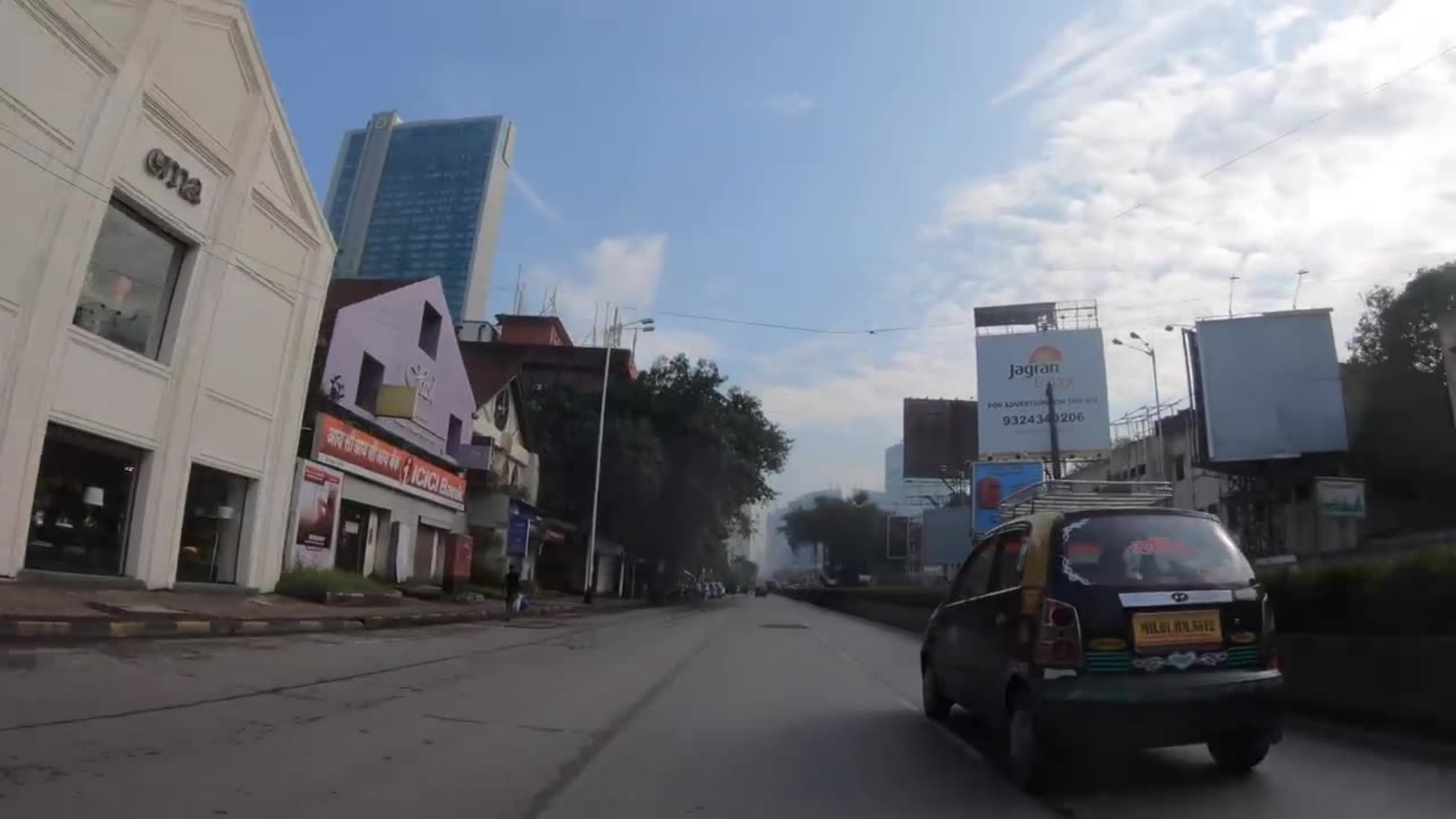}
        \includegraphics[width=\resultimagewidth\linewidth]{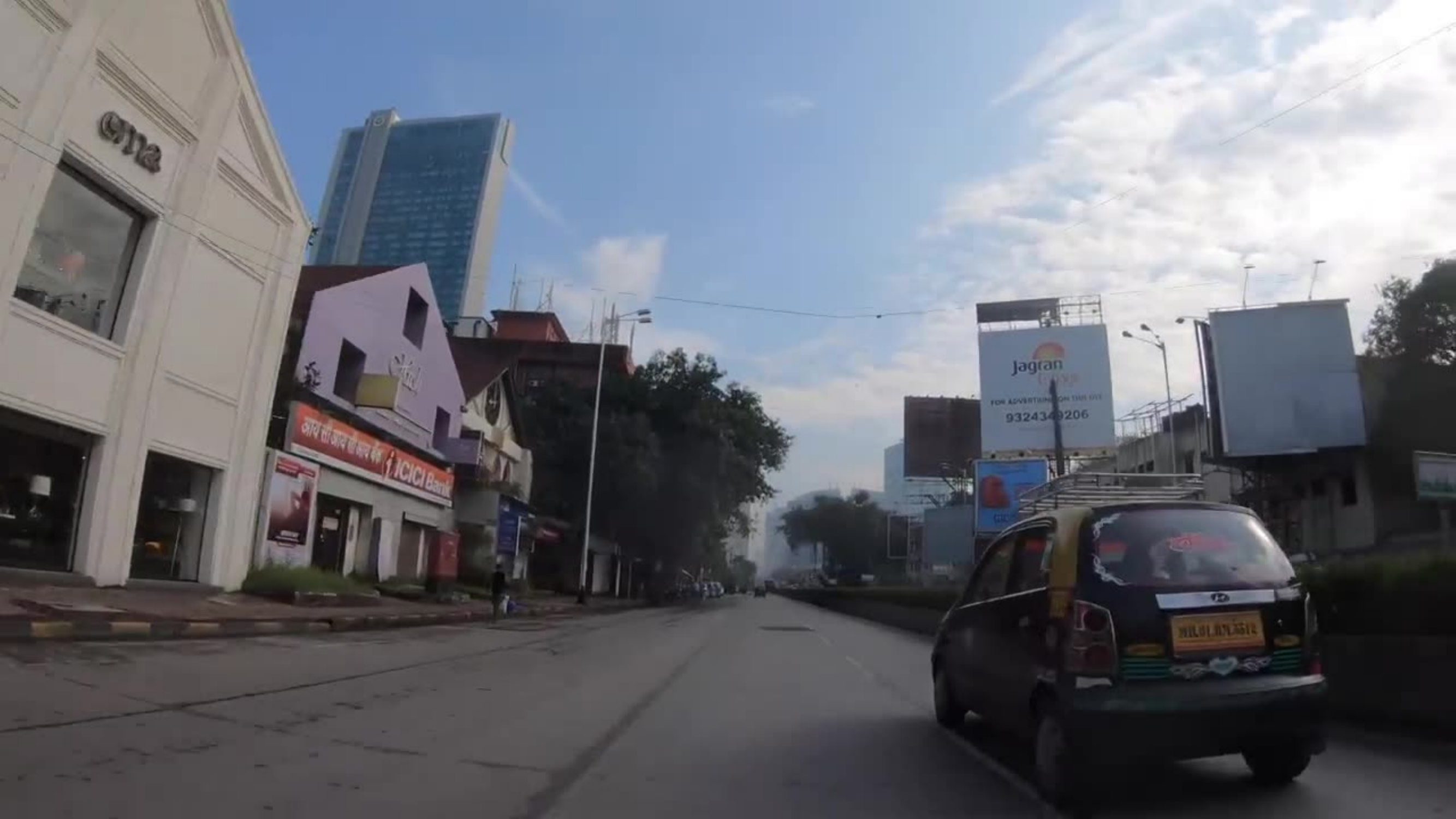}
        \includegraphics[width=\resultimagewidth\linewidth]{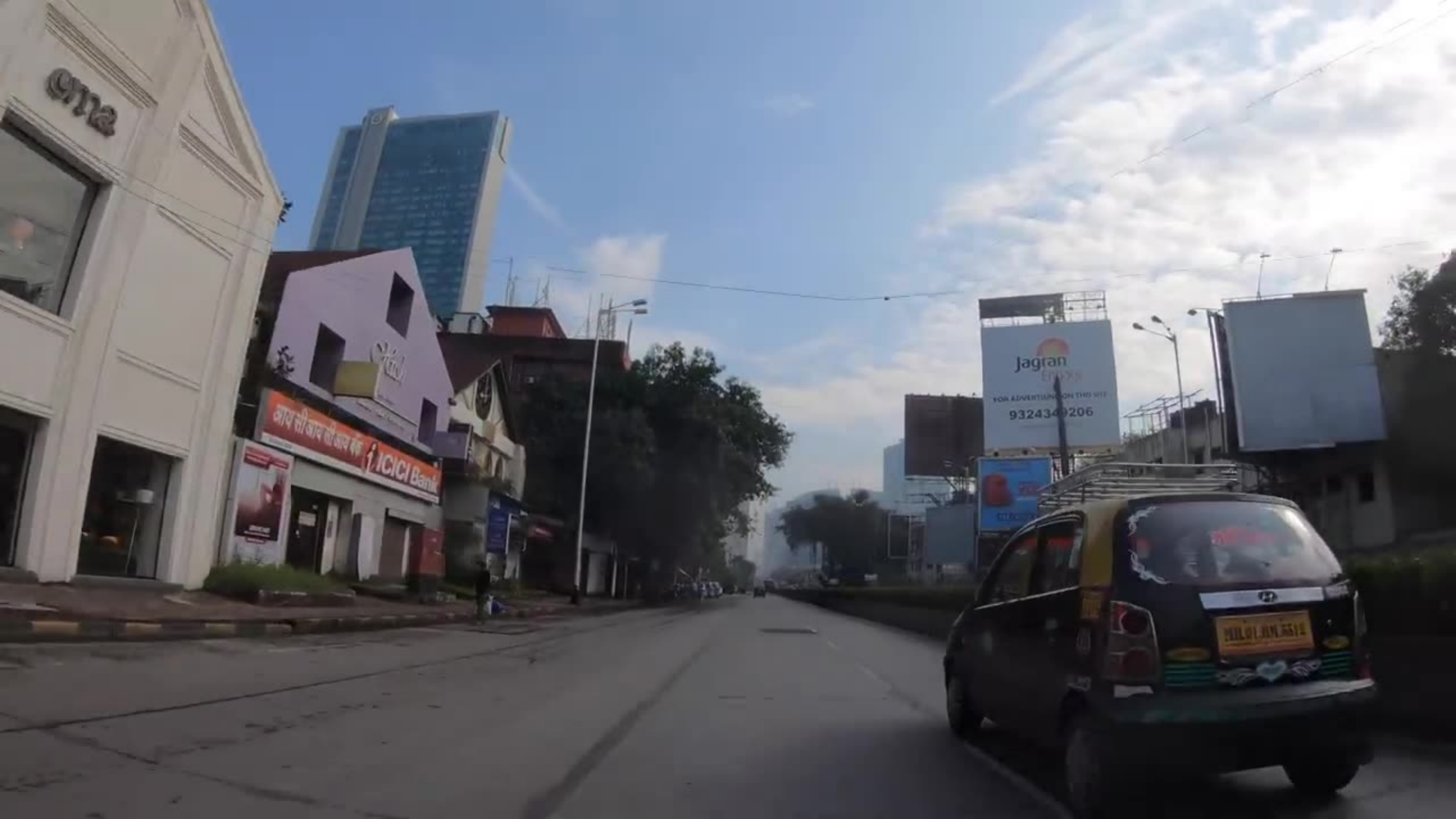}
        \includegraphics[width=\resultimagewidth\linewidth]{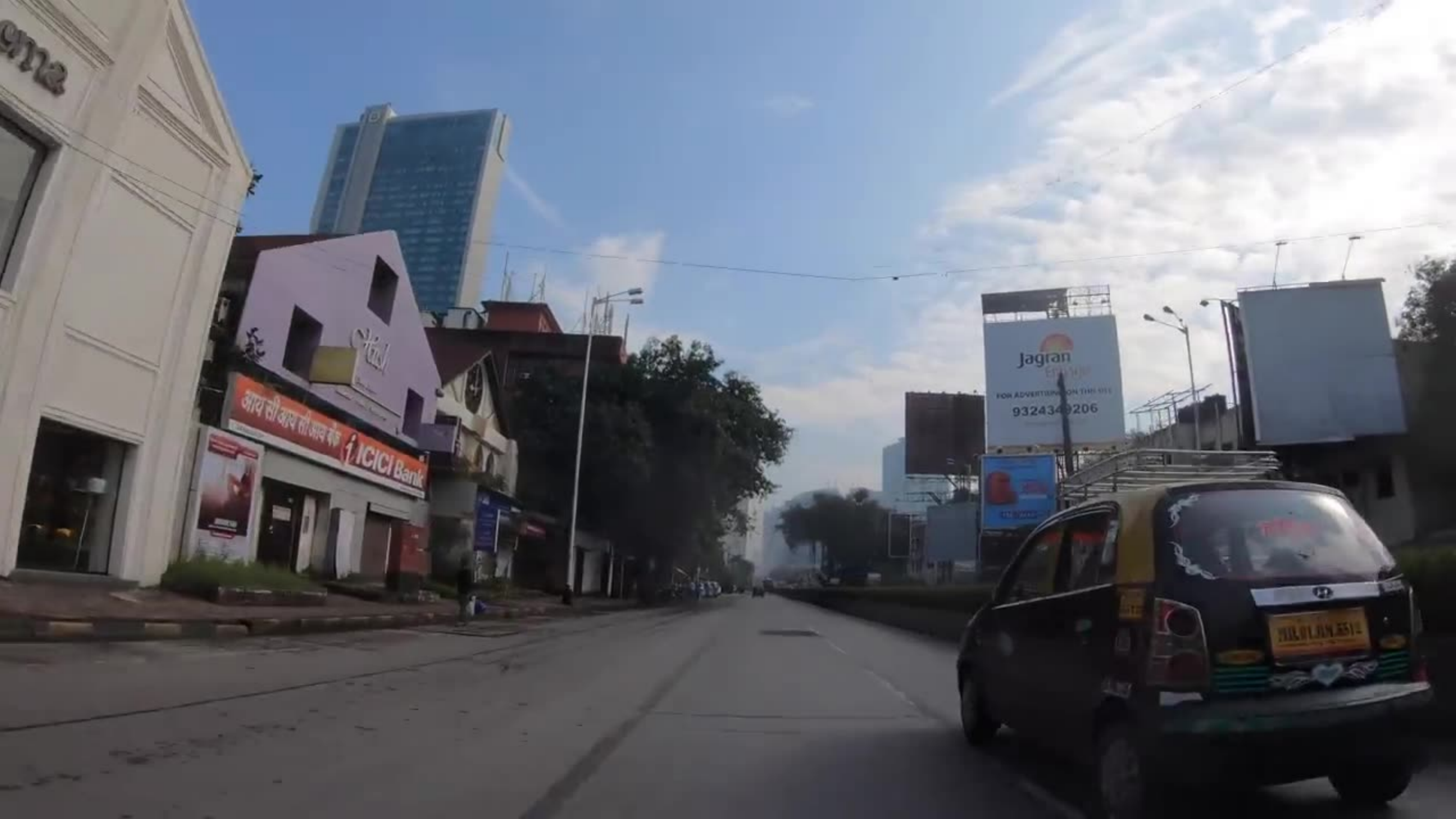}
        \includegraphics[width=\resultimagewidth\linewidth]{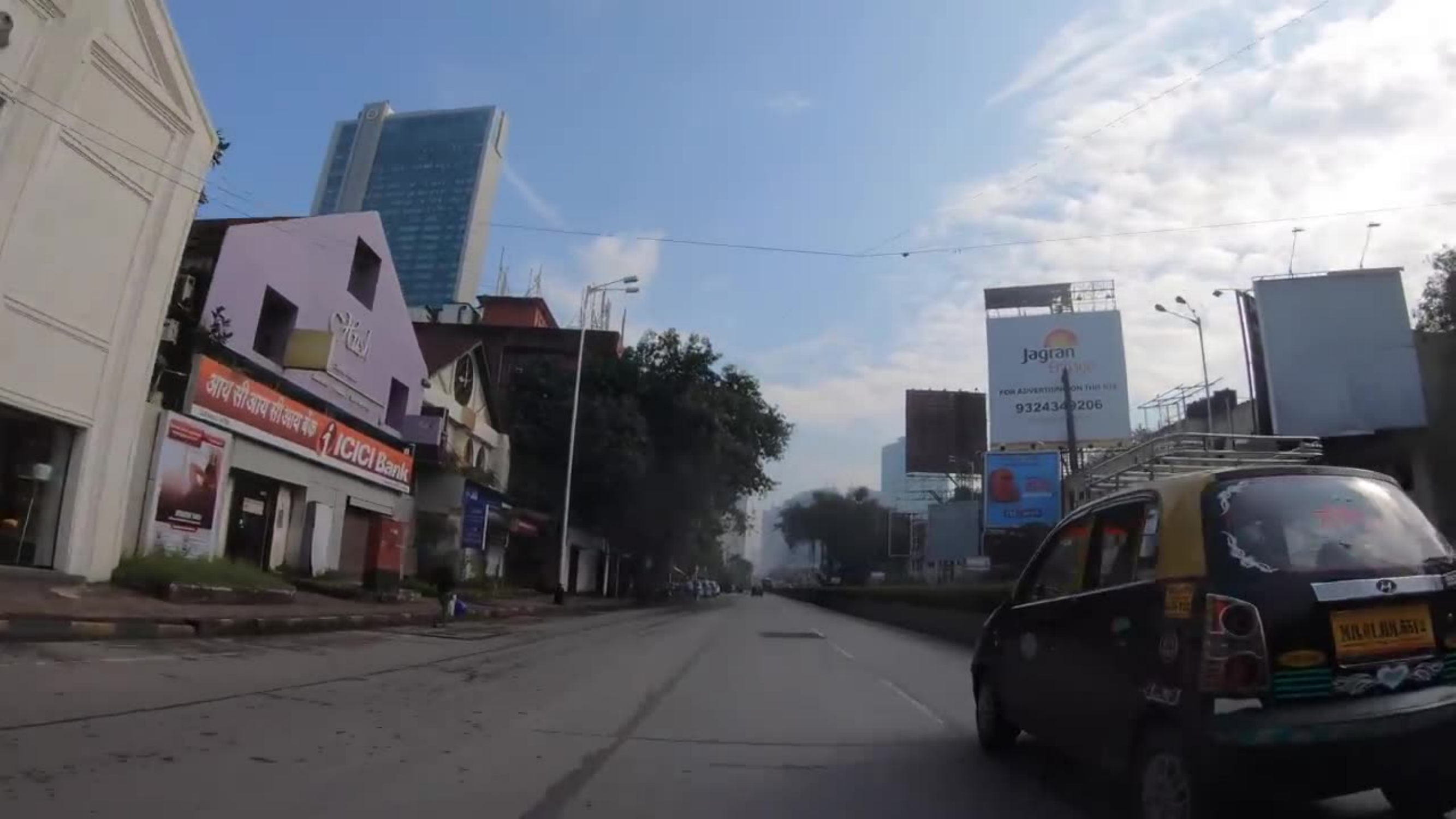}
    }
    
    \subcaptionbox{
        ground truth
    }{
        \includegraphics[width=\resultimagewidth\linewidth]{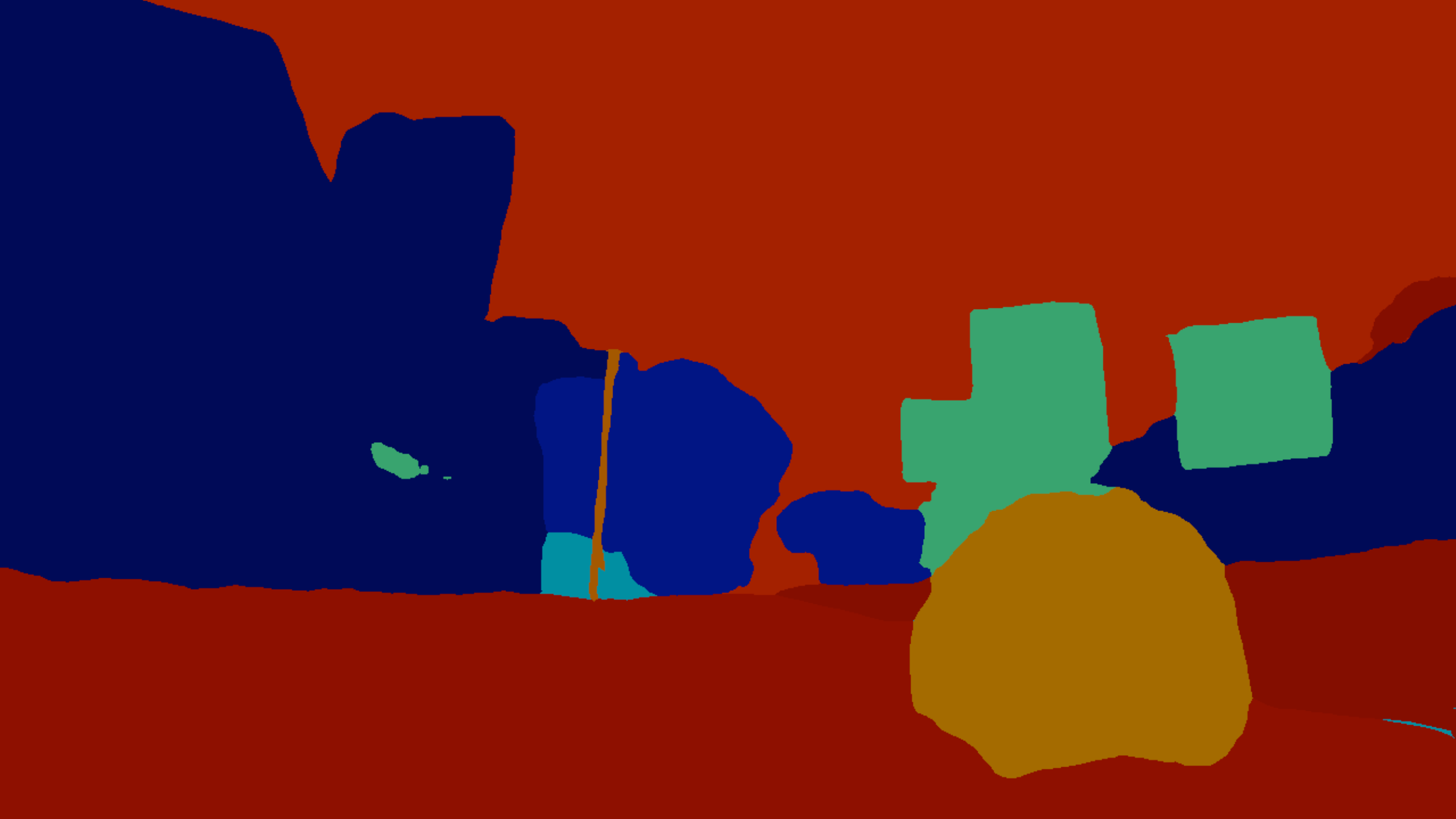}
        \includegraphics[width=\resultimagewidth\linewidth]{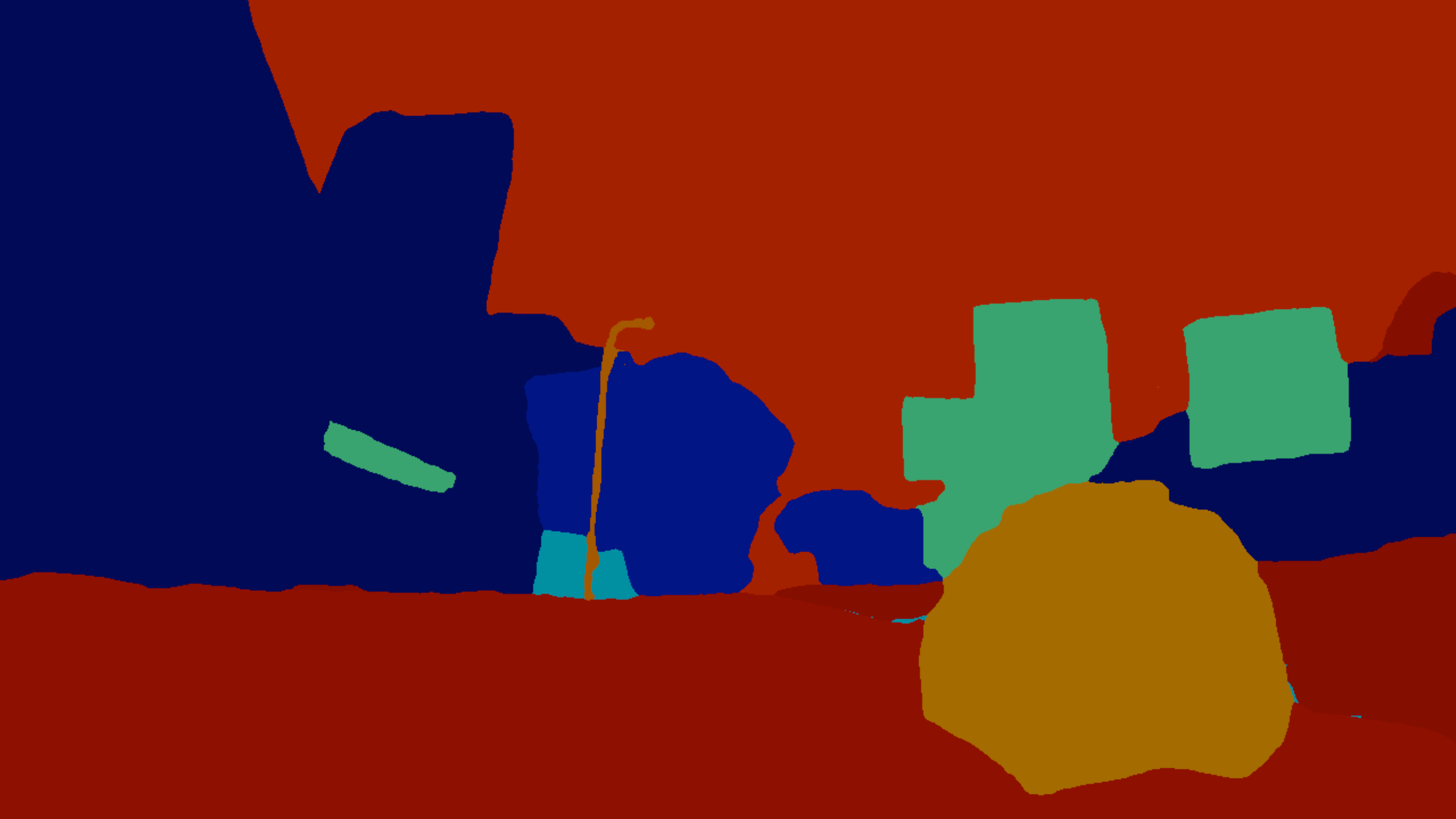}
        \includegraphics[width=\resultimagewidth\linewidth]{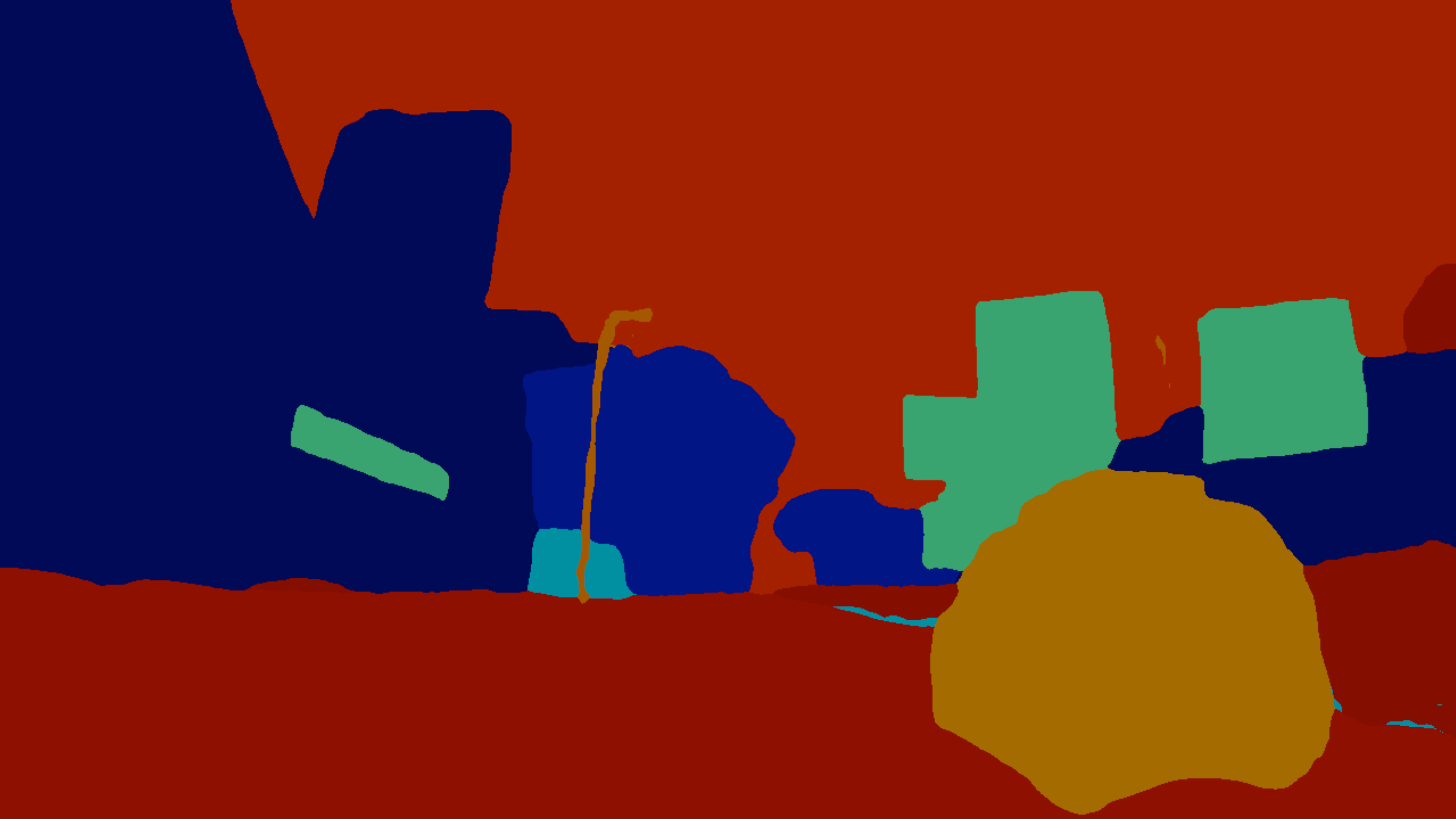}
        \includegraphics[width=\resultimagewidth\linewidth]{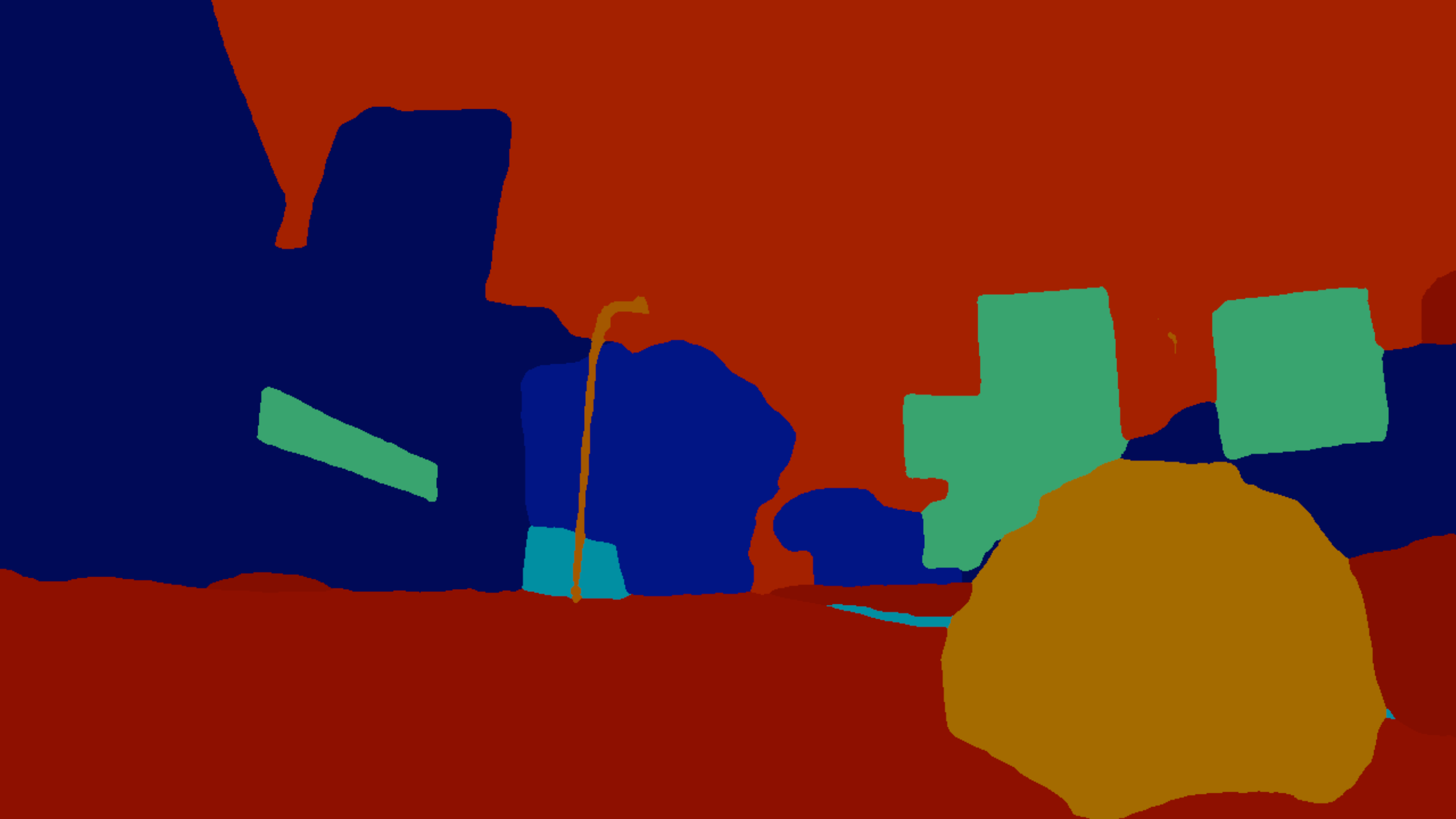}
        \includegraphics[width=\resultimagewidth\linewidth]{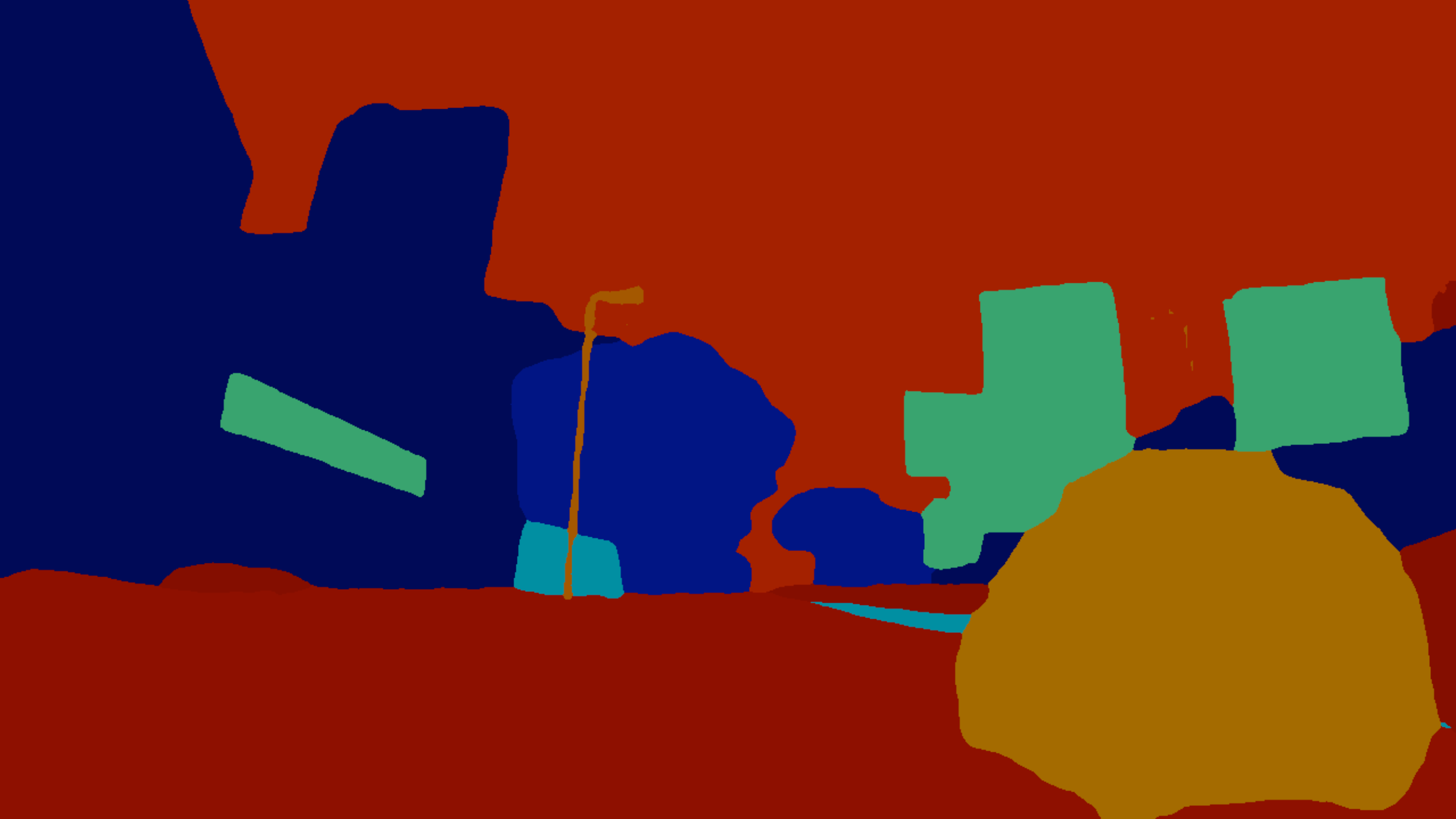}
        \includegraphics[width=\resultimagewidth\linewidth]{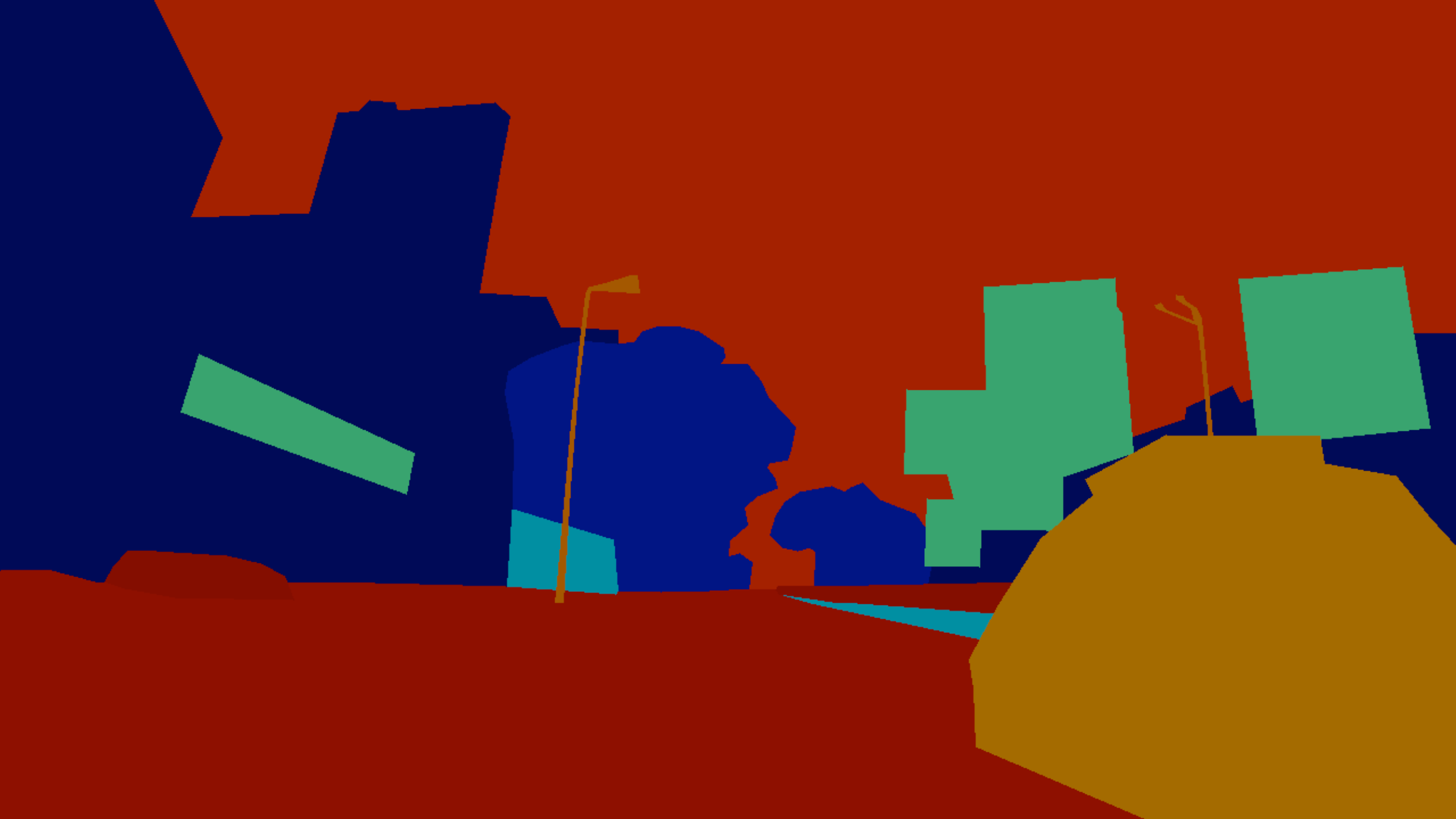}
    }
    
    \subcaptionbox{
        no shift
    }{
        \includegraphics[width=\resultimagewidth\linewidth]{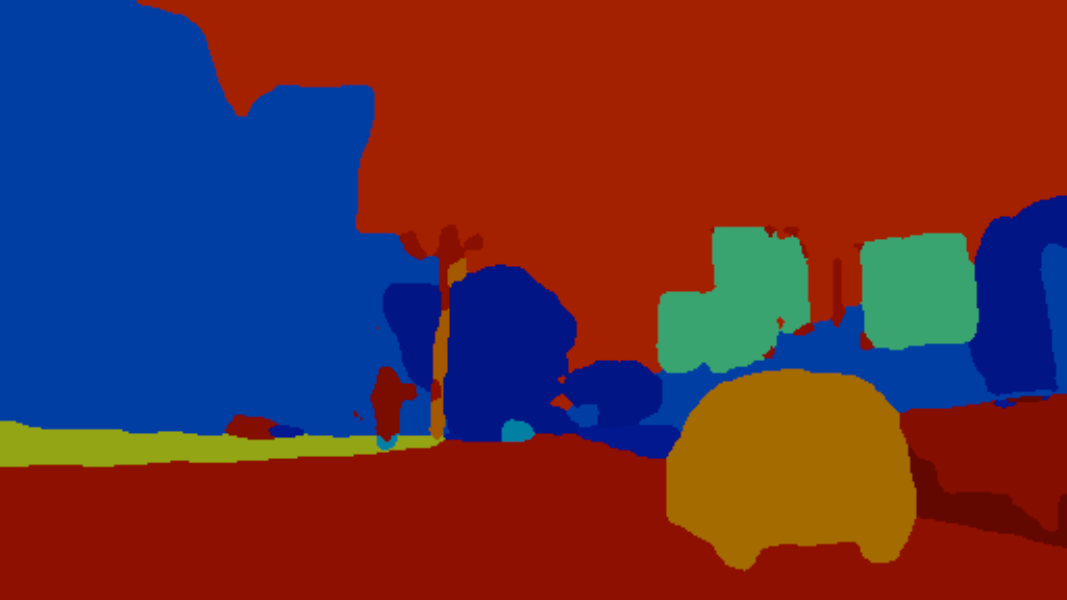}
        \includegraphics[width=\resultimagewidth\linewidth]{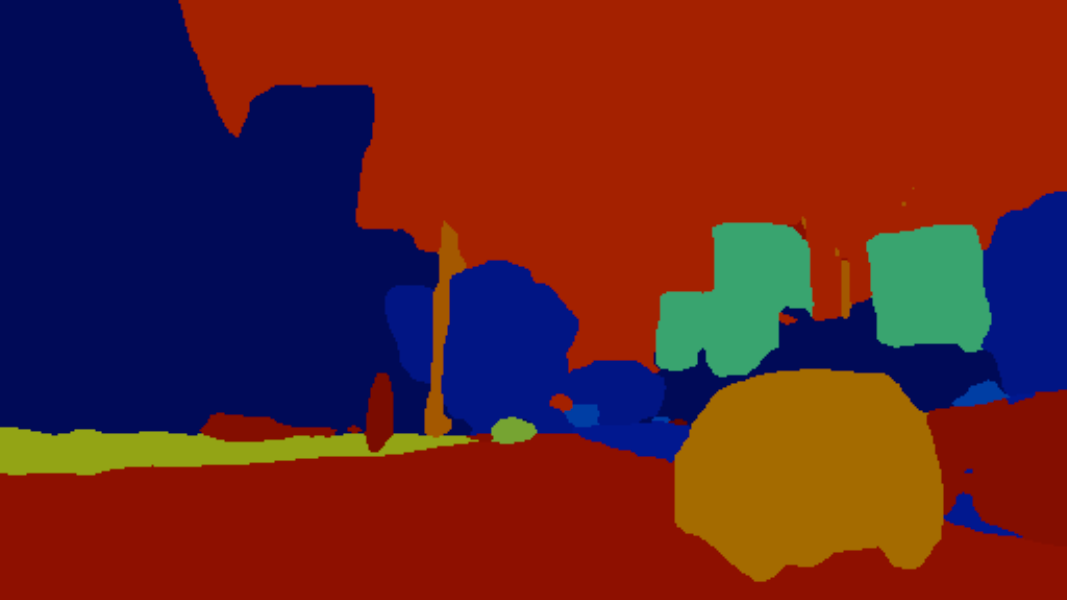}
        \includegraphics[width=\resultimagewidth\linewidth]{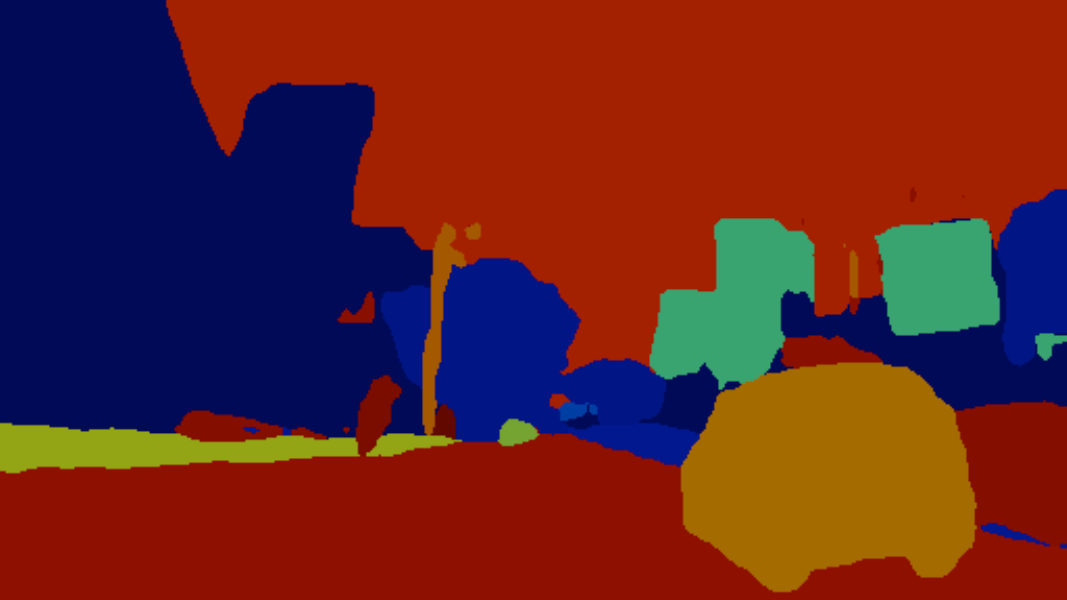}
        \includegraphics[width=\resultimagewidth\linewidth]{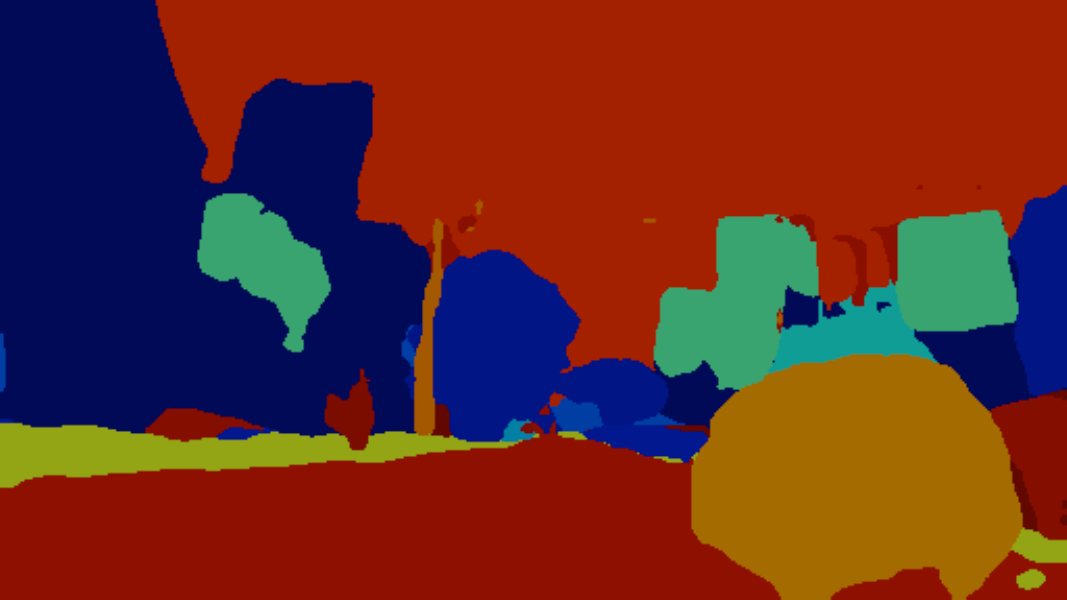}
        \includegraphics[width=\resultimagewidth\linewidth]{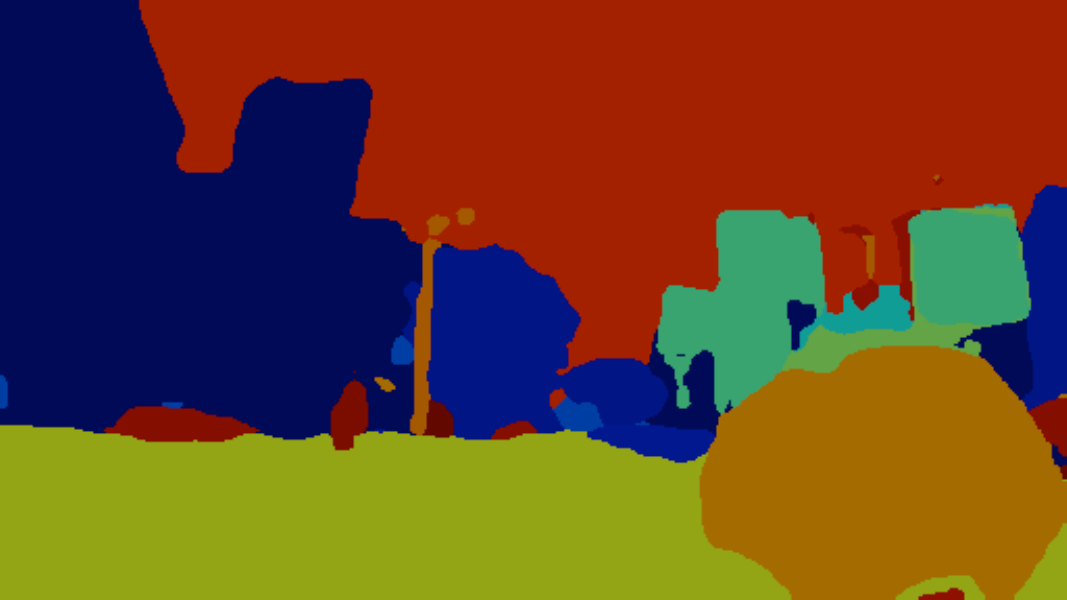}
        \includegraphics[width=\resultimagewidth\linewidth]{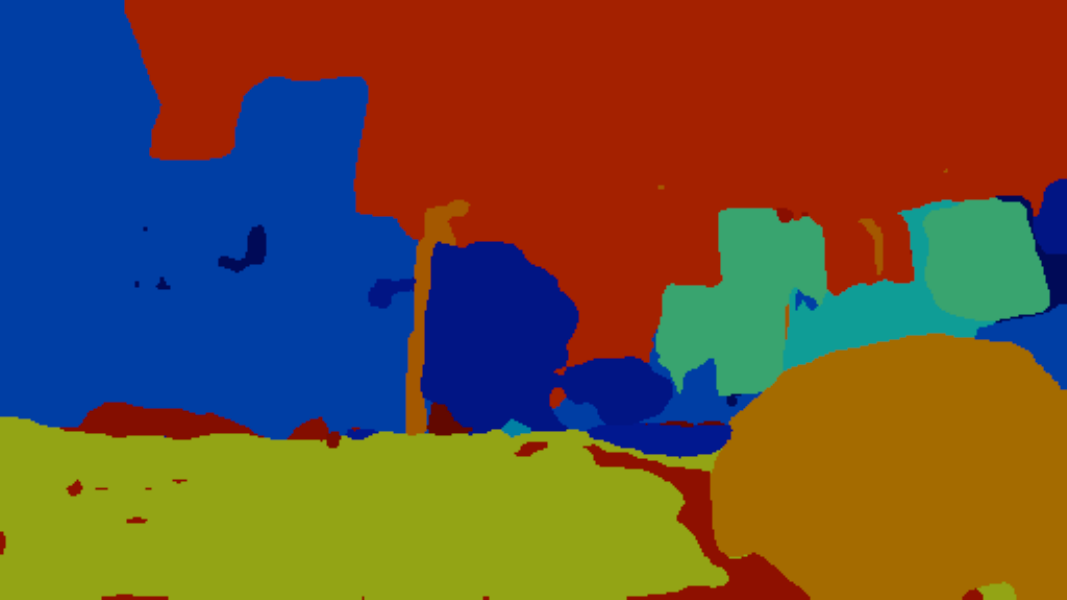}
    }
    
    \subcaptionbox{
        1/16 with matching
        \label{fig:vspw 1/16 with matching}
    }{
        \includegraphics[width=\resultimagewidth\linewidth]{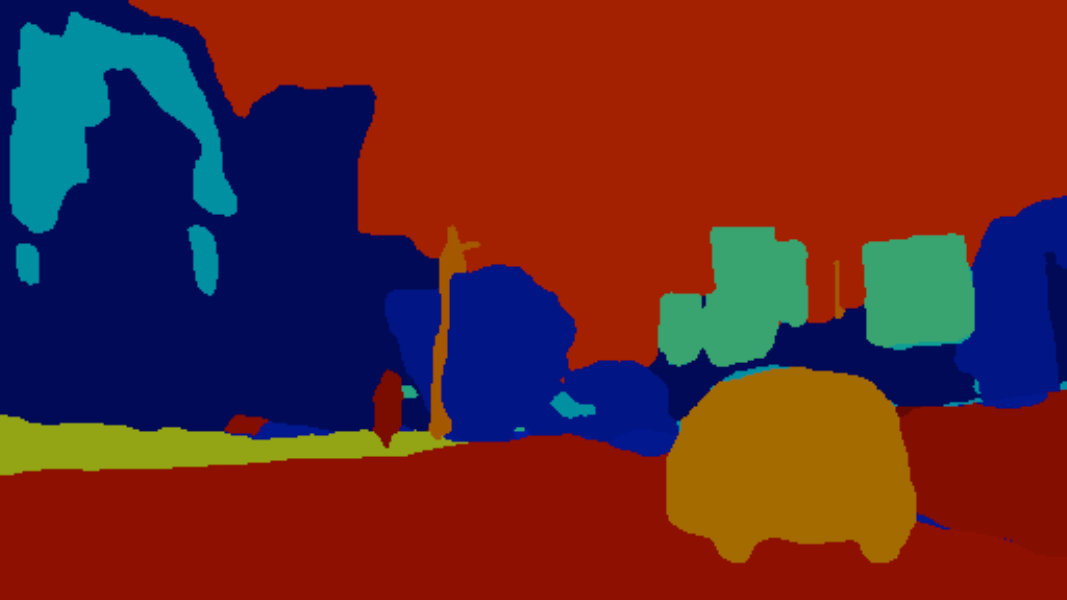}
        \includegraphics[width=\resultimagewidth\linewidth]{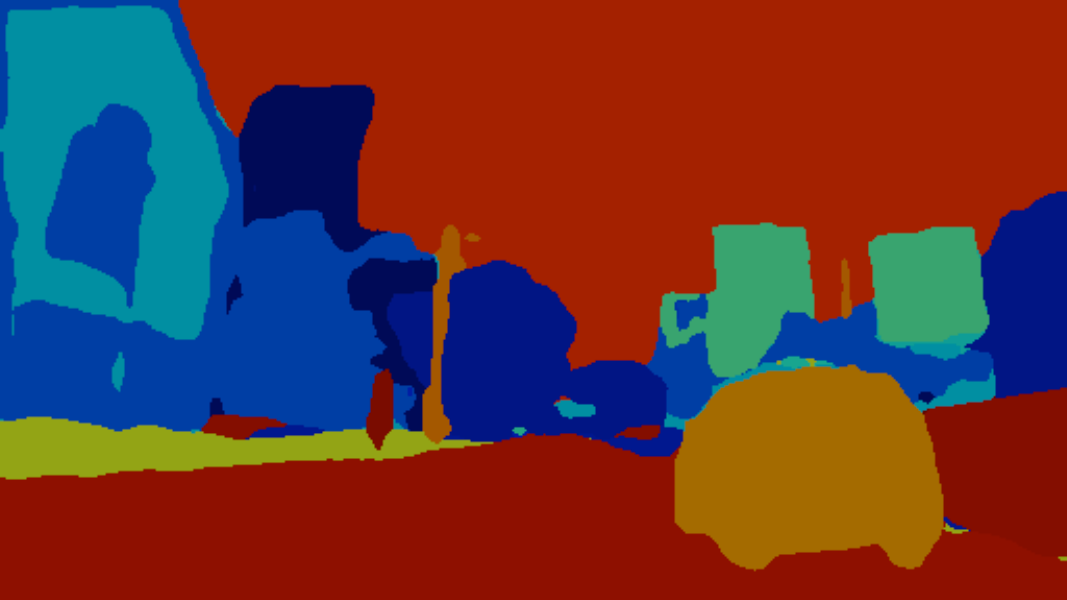}
        \includegraphics[width=\resultimagewidth\linewidth]{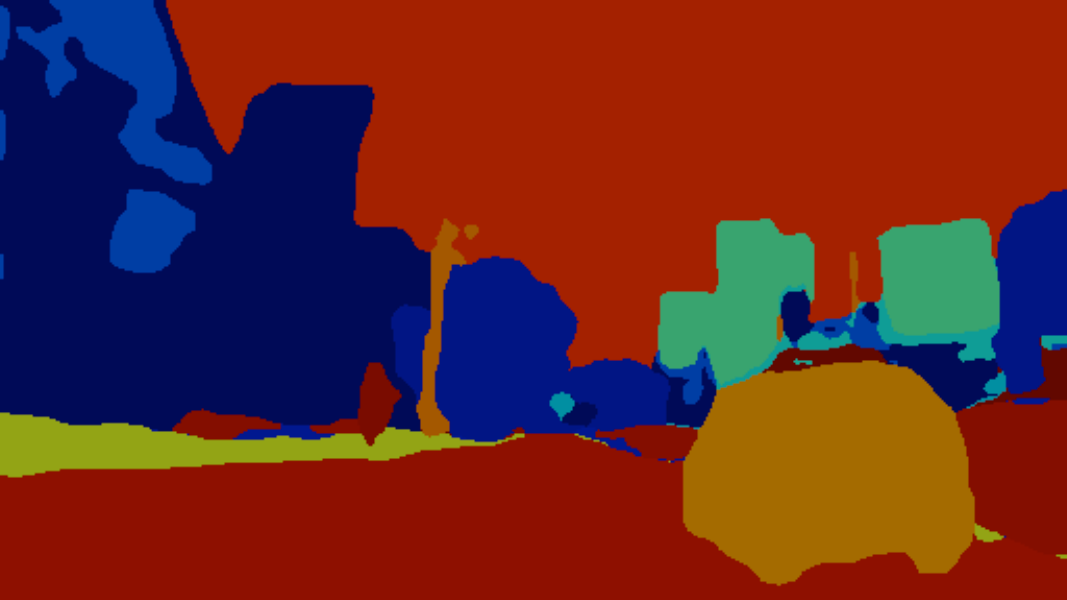}
        \includegraphics[width=\resultimagewidth\linewidth]{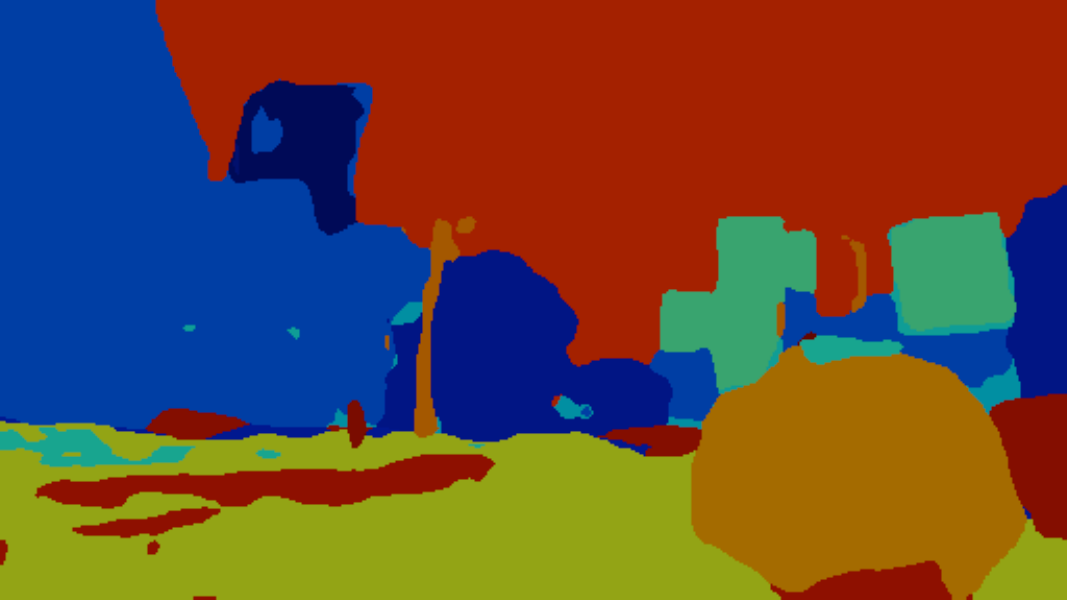}
        \includegraphics[width=\resultimagewidth\linewidth]{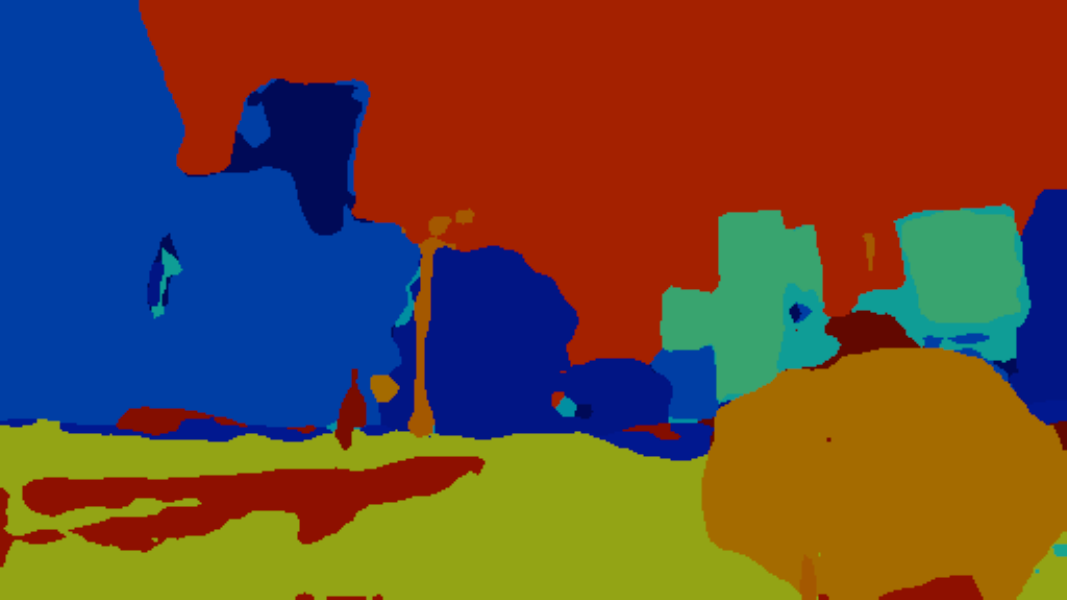}
        \includegraphics[width=\resultimagewidth\linewidth]{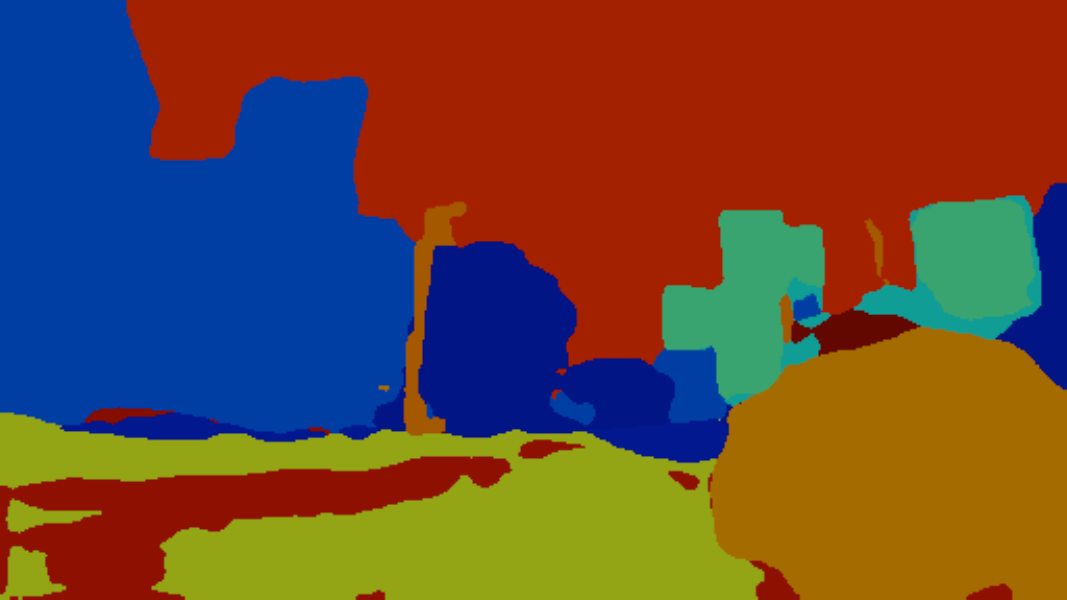}
    }

    \subcaptionbox{
        1/16 without matching
        \label{fig:vspw 1/16 without matching}
    }{
        \includegraphics[width=\resultimagewidth\linewidth]{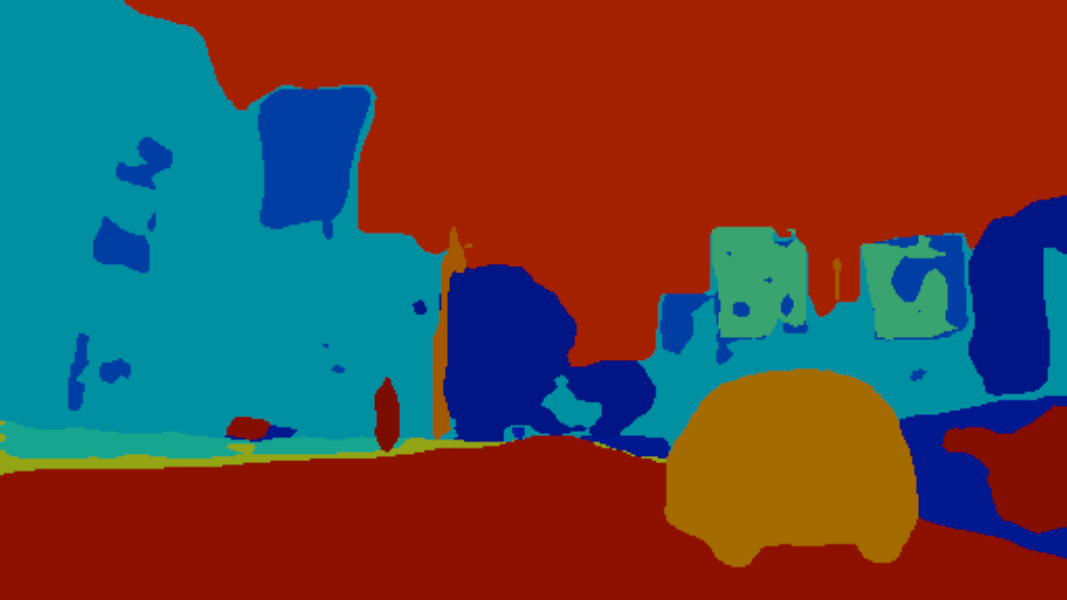}
        \includegraphics[width=\resultimagewidth\linewidth]{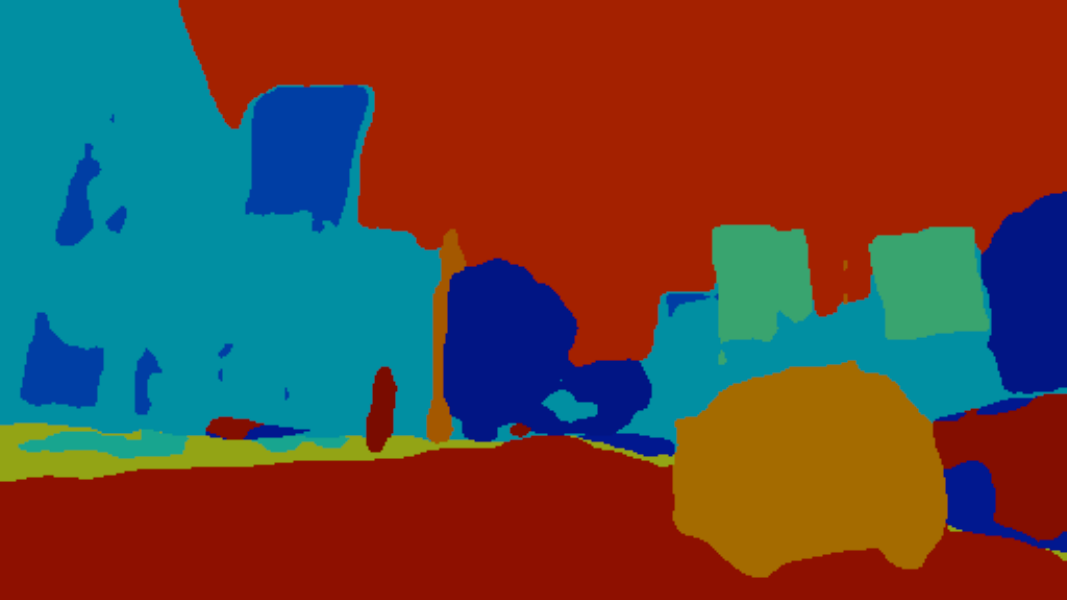}
        \includegraphics[width=\resultimagewidth\linewidth]{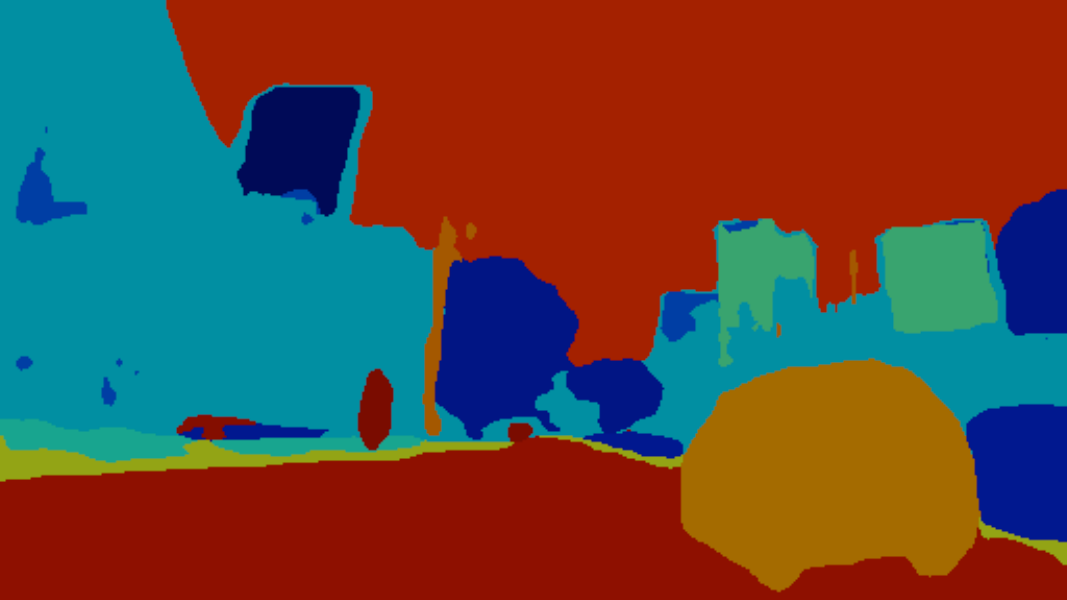}
        \includegraphics[width=\resultimagewidth\linewidth]{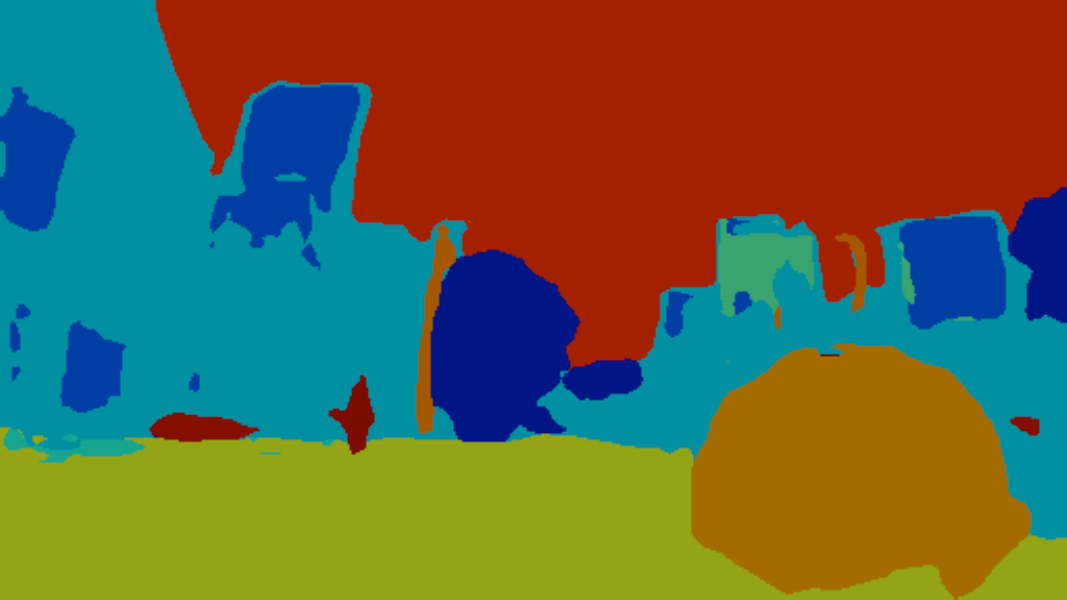}
        \includegraphics[width=\resultimagewidth\linewidth]{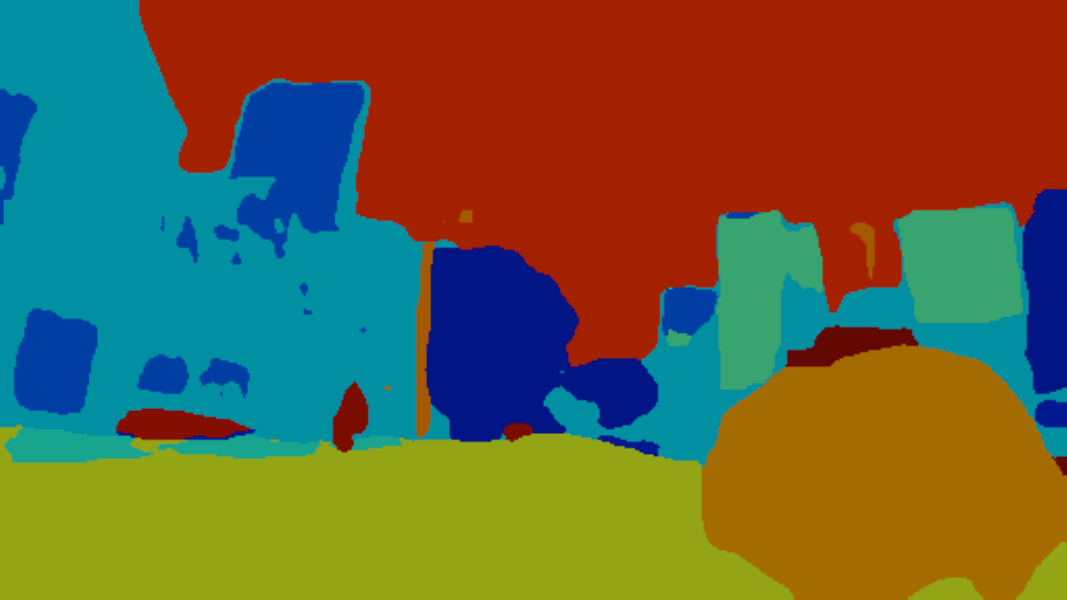}
        \includegraphics[width=\resultimagewidth\linewidth]{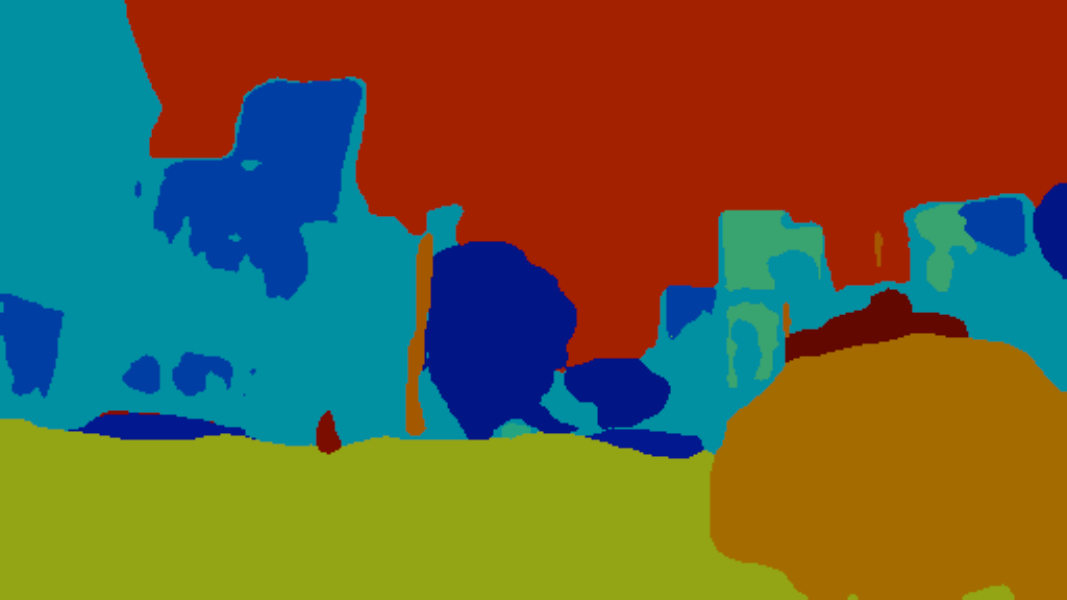}
    }
    
    \subcaptionbox{
        1/8 with matching
        \label{fig:vspw 1/8 with matching}
    }{
        \includegraphics[width=\resultimagewidth\linewidth]{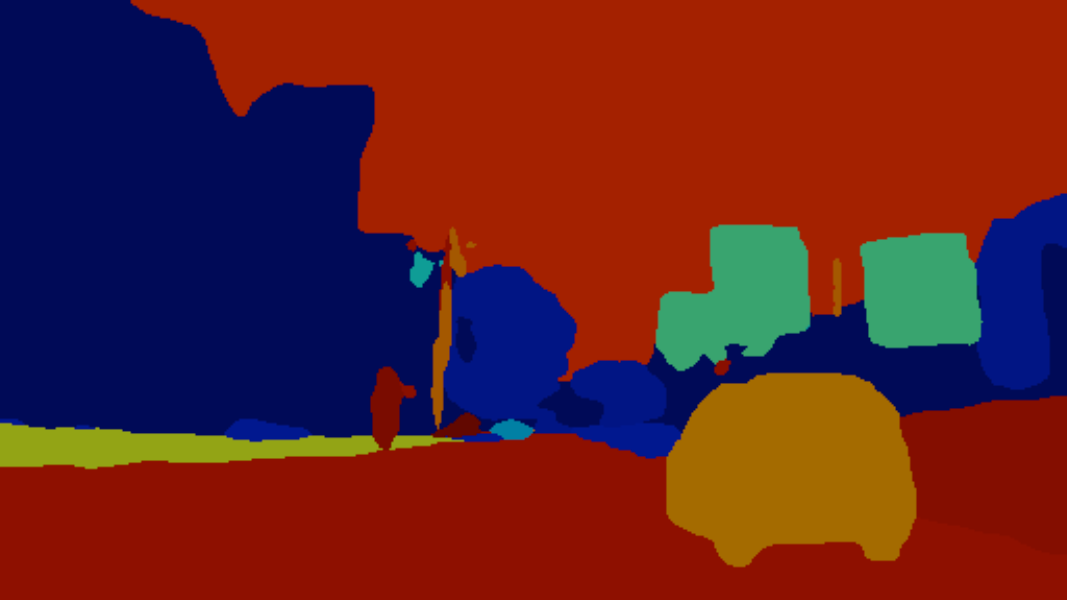}
        \includegraphics[width=\resultimagewidth\linewidth]{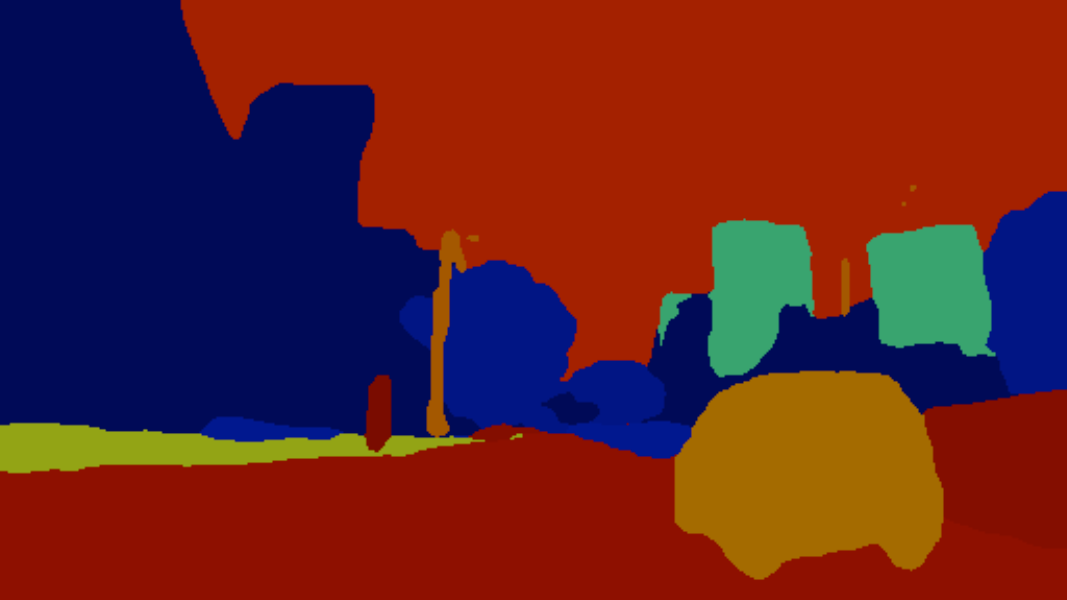}
        \includegraphics[width=\resultimagewidth\linewidth]{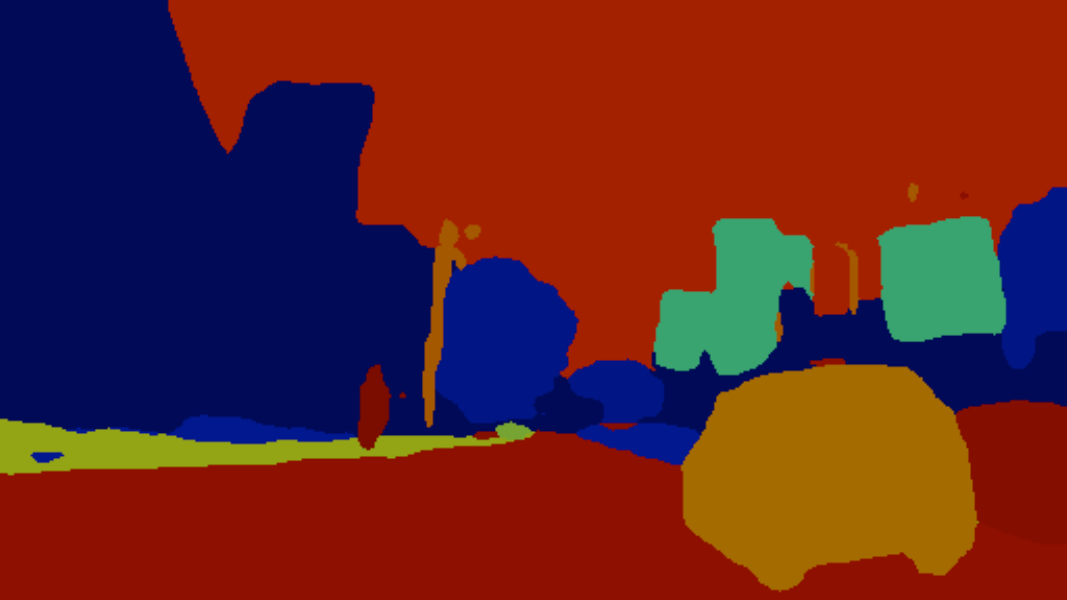}
        \includegraphics[width=\resultimagewidth\linewidth]{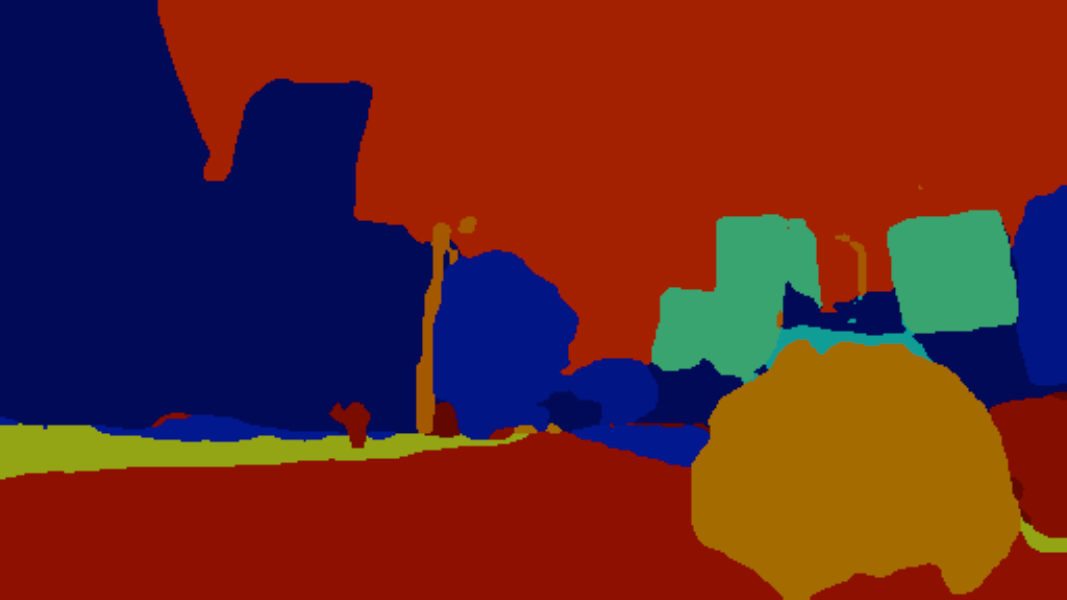}
        \includegraphics[width=\resultimagewidth\linewidth]{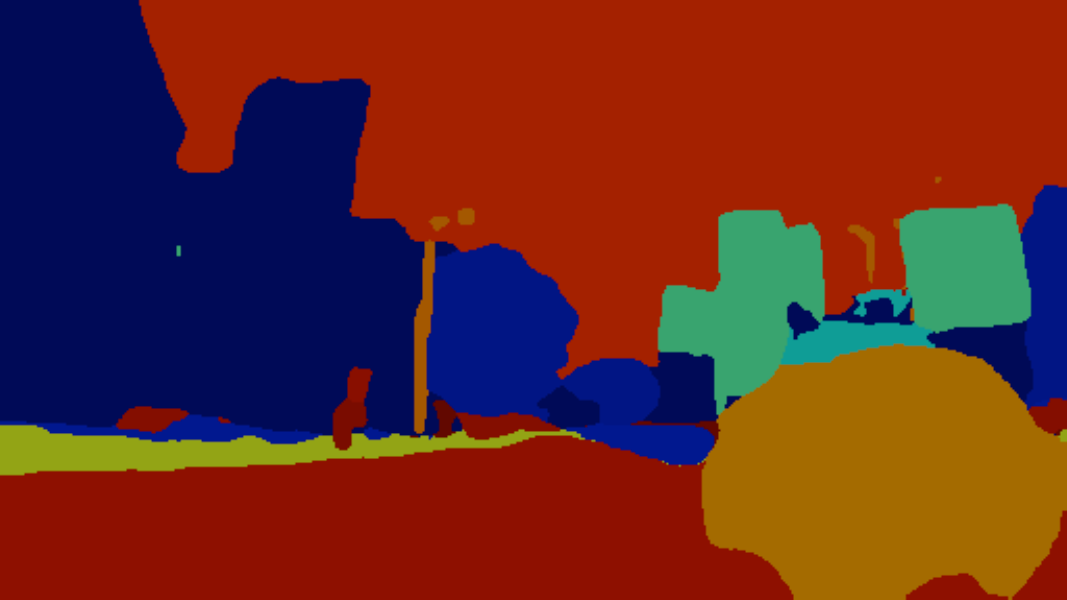}
        \includegraphics[width=\resultimagewidth\linewidth]{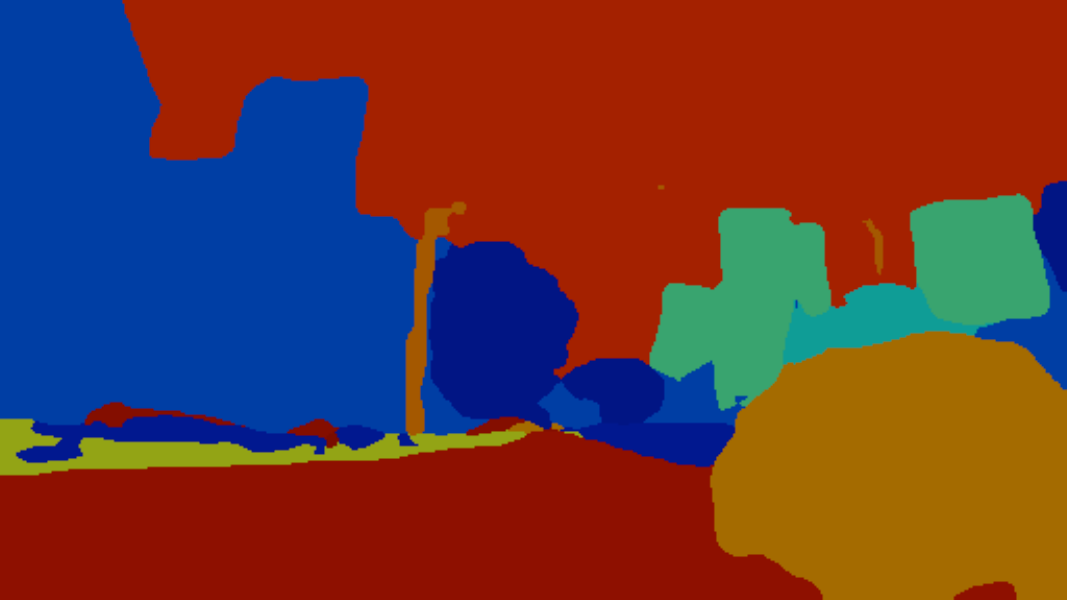}
    }

    \subcaptionbox{
        1/8 without matching
        \label{fig:vspw 1/8 without matching}
    }{
        \includegraphics[width=\resultimagewidth\linewidth]{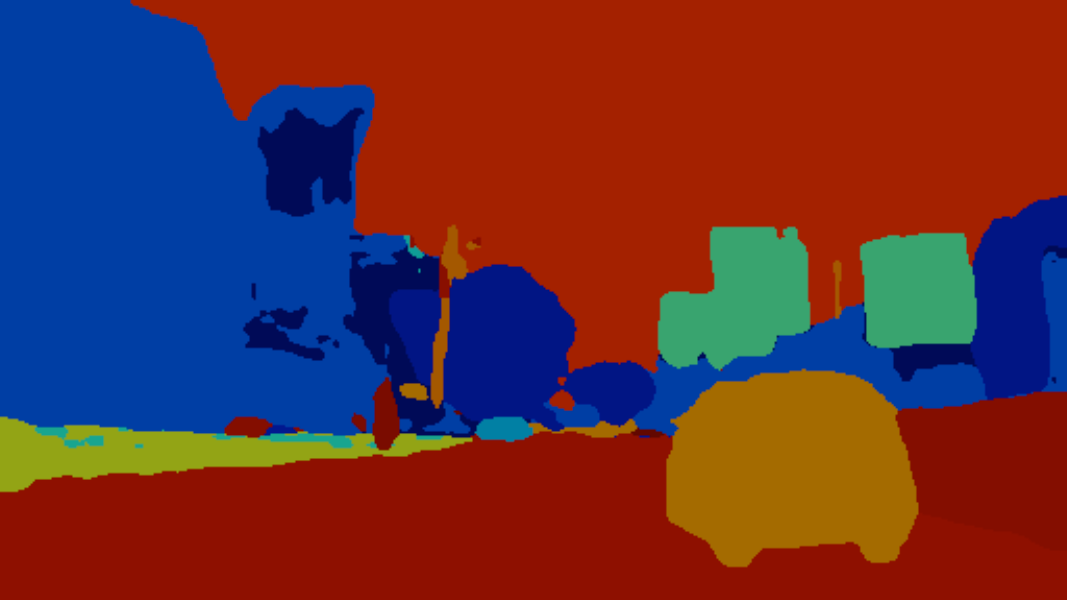}
        \includegraphics[width=\resultimagewidth\linewidth]{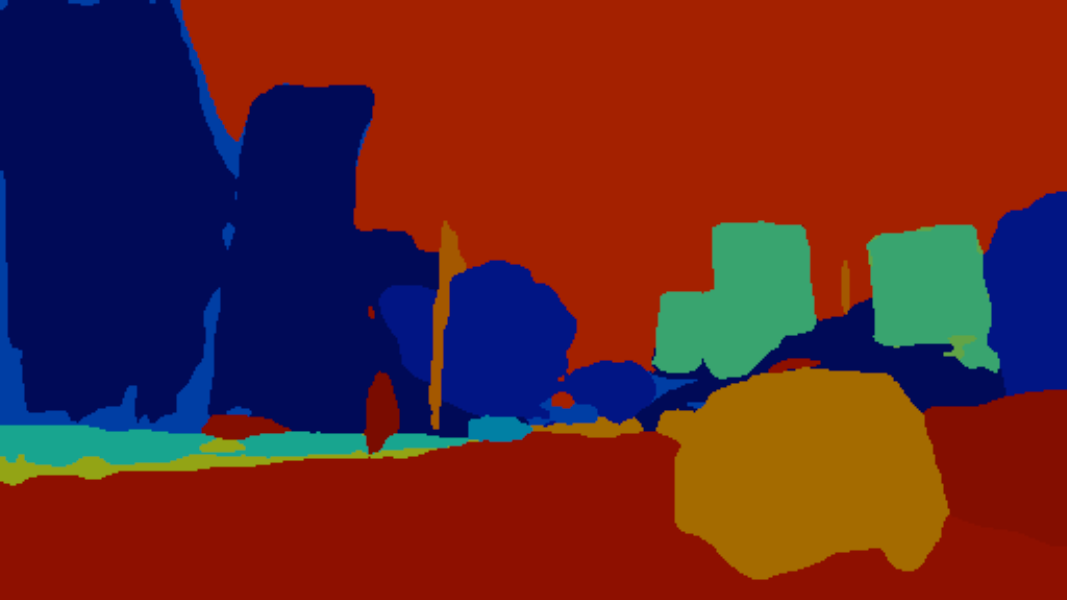}
        \includegraphics[width=\resultimagewidth\linewidth]{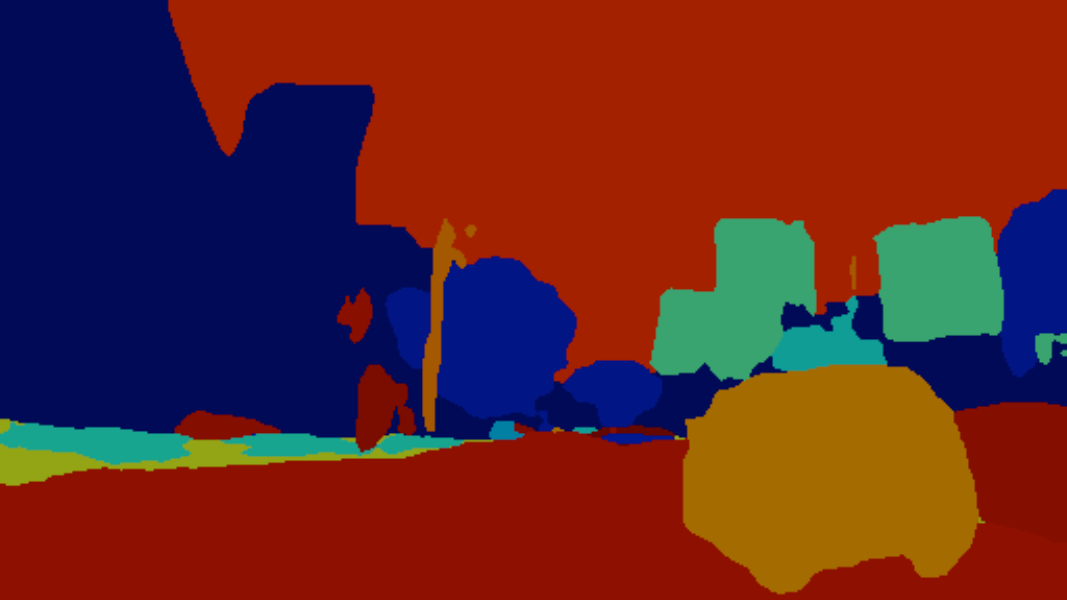}
        \includegraphics[width=\resultimagewidth\linewidth]{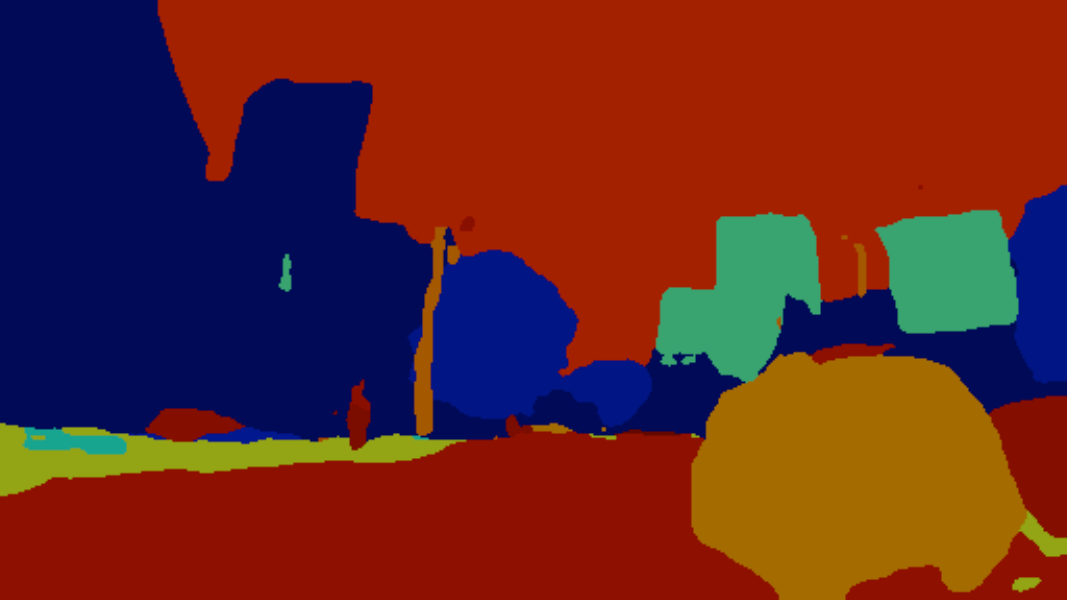}
        \includegraphics[width=\resultimagewidth\linewidth]{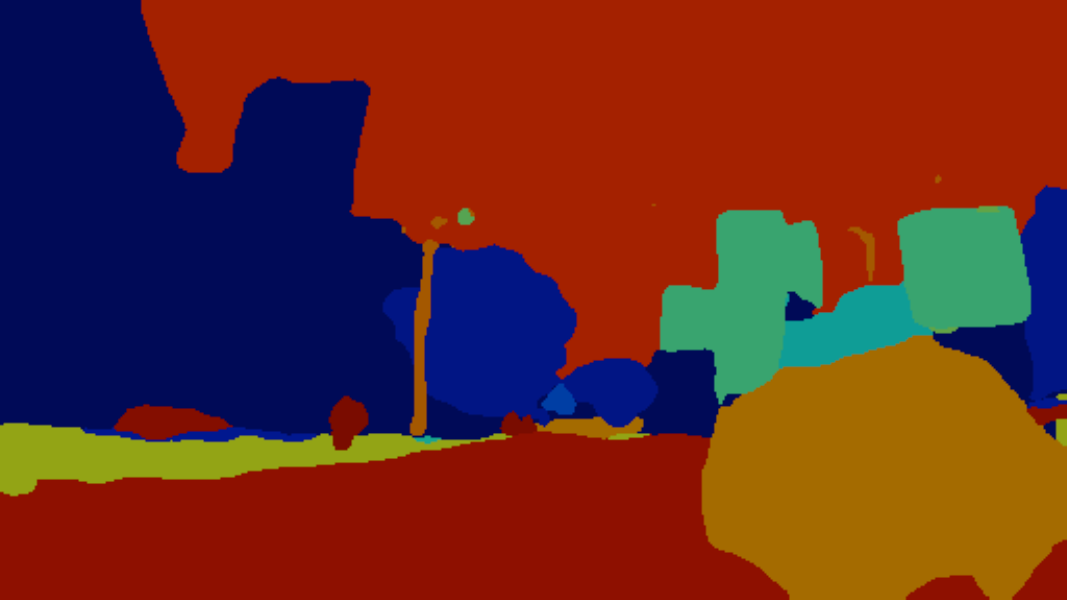}
        \includegraphics[width=\resultimagewidth\linewidth]{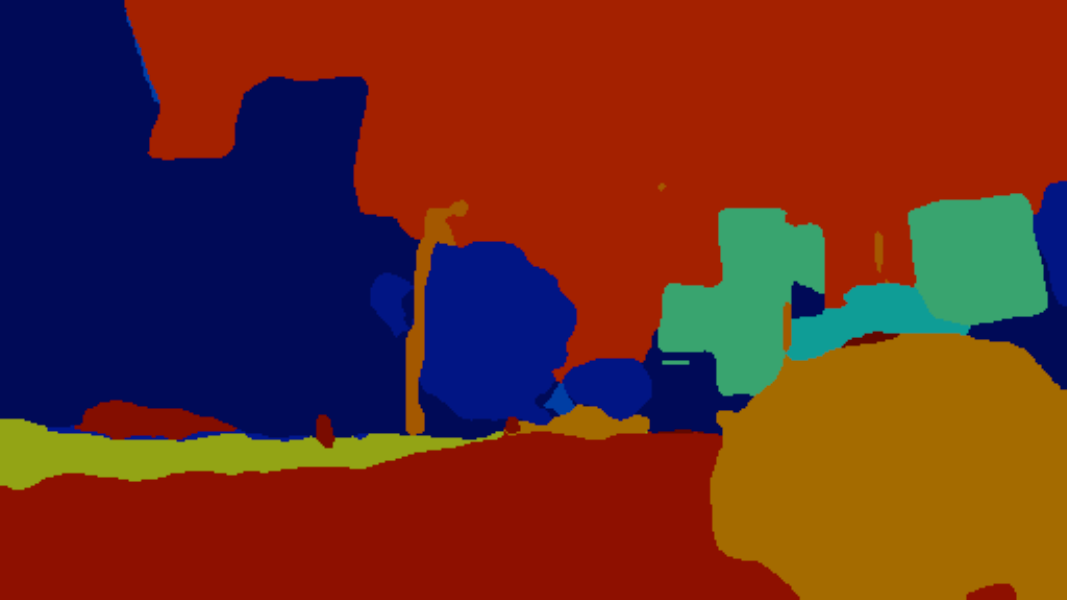}
    }
    
    \caption{
        Segmentation results for VSPW.
        First two rows show
        (a) original frames, and 
        (b) ground truth segmentation labels.
        The following rows show results of
        (c) the baseline (no shifts, no query matching),
        (d) 1/16 shift with and (e) without matching,
        (f) 1/8  shift with and (g) without matching.
    }
    \label{segmentation_result_2}
\end{figure*}

Figure~\ref{segmentation_result_2} shows examples of segmentation results for VSPW with and without query matching for a shift of 1/16 and 1/8.
In frames 4 to 6, an incorrect segmentation of ``road'' (dark red) was observed as ``path'' (dark yellow) without shift or with a small shift. This was corrected by increasing the shift fraction to 1/16. Even with a shift of 1/16, performance was still poor without matching (for example, ``building'' was confused with ``wall''). However, improvement was observed with the proposed query matching, which is also seen in the category ``billboard\_or\_Bulletin\_Board,'' suggesting that appropriate matching of queries leads to better results.

\section{Conclusion}

In this paper, we propose a video segmentation method that leverages pre-trained image segmentation models and incorporates feature shift to model temporal information. The feature shift, commonly used in action recognition, is applied to segmentation for the first time. The proposed method uses a query-based architecture, where we find correspondences between decoded queries representing segmentation masks in different frames. Using the proposed query matching, we perform feature shift to maintain coherency across frames. Our results demonstrate significant improvements in performance, especially on dense video datasets like VSPW, highlighting the effectiveness of the proposed method in enhancing segmentation quality while efficiently reusing pre-trained weights.

Future work includes improving the performance of the proposed method and comparing it with other video segmentation methods. Instead of developing an entirely new video model, we will explore approaches that retain the core architecture, applying a segmentation model to images frame by frame, with a detailed yet straightforward extension.

\section*{Acknowledgements}
The authors thank Shimon Hori for his help in implementing the shift to a query-based architecture.

This work was supported in part by JSPS KAKENHI Grant Number JP22K12090.

\bibliographystyle{splncs04}
\bibliography{all,mybib}

\end{document}